\documentclass[10pt,twocolumn,letterpaper]{article}

\usepackage{cvpr}
\usepackage{times}
\usepackage{epsfig}
\usepackage{graphicx}
\usepackage{amsmath}
\usepackage{amssymb}
\usepackage{color}
\usepackage{stackengine}
\usepackage{wrapfig,booktabs}
\usepackage{float}
\usepackage{url}
\usepackage{multirow}
\usepackage{lipsum}  

\usepackage[pagebackref=true,breaklinks=true,letterpaper=true,colorlinks,bookmarks=false]{hyperref}

\cvprfinalcopy 

\def\cvprPaperID{4} 

\ifcvprfinal\fi
\begin{document}

\title{Towards Automated Melanoma Detection with Deep Learning:\\ Data Purification and Augmentation}
\author{Devansh Bisla, Anna Choromanska\\
Dept of Electrical Engineering, \\
Tandon School of Engineering, New York University\\
{\tt\small bisla@nyu.edu, ac5455@nyu.edu}
\and 
 Russell S. Berman \\
Division of Surgical Oncology, Department of Surgery, \\
New York University School of Medicine \\
\and
Jennifer A. Stein, David Polsky \\
Ronald O. Perelman Department of Dermatology,\\
New York University School of Medicine
}

\maketitle
\begin{abstract}
Melanoma is one of ten most common cancers in the US. Early detection is crucial for survival, but often the cancer is diagnosed in the fatal stage. Deep learning has the potential to improve cancer detection rates, but its applicability to melanoma detection is compromised by the limitations of the available skin lesion data bases, which are small, heavily imbalanced, and contain images with occlusions. We build deep-learning-based tools for data purification and augmentation to counter-act these limitations. The developed tools can be utilized in a deep learning system for lesion classification and we show how to build such system. The system heavily relies on the processing unit for removing image occlusions and the data generation unit, based on generative adversarial networks, for populating scarce lesion classes, or equivalently creating virtual patients with pre-defined types of lesions. We empirically verify our approach and show that incorporating these two units into melanoma detection system results in the superior performance over common baselines.
\end{abstract}

\section{Introduction}
\begin{figure}[h]
    \centering
    \includegraphics[width=0.45\textwidth]{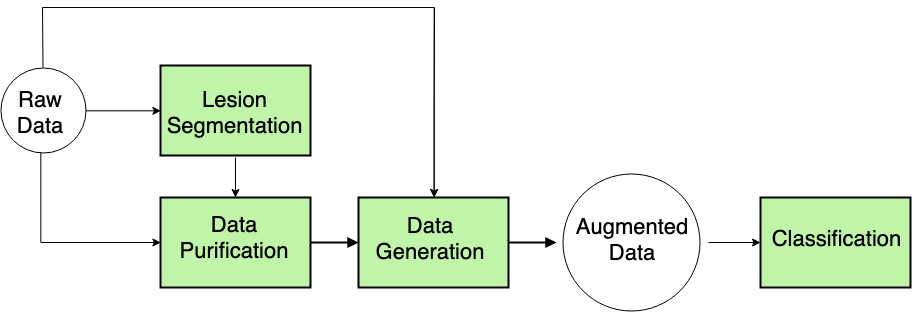}
    \caption{The block diagram of the complete system for lesion classification. Lesion segmentation, data purification, and data generation models are utilized to purify and augment the training data set.}
    \label{fig:block}
\end{figure}

According to the American Cancer Society, melanoma takes one life in every $54$ minutes \cite{CancerFa15:online} in US. In Australia, where the incidence of melanoma are the highest in the world along with New Zealand, one person dies every five hours from melanoma \cite{CancerAu15:online}. There were approximately $87,110$ cases of melanoma in the US and $13,941$ in Australia \cite{AustraliaGov} alone in the year $2017$. Even though it accounts for less than $1\%$ of the total skin diseases, it is the major cause of deaths related to these diseases. The $5$-year survival rate in the US is $98\%$ and reduces down to $18\%$ once it spreads to distant organs. Therefore, early detection of melanoma is of fundamental importance to increase the survival rates. The techniques and technologies that aim at automating the visual examination of skin lesions, traditionally done by dermatologists, are targeted to i) assist clinicians with navigating patients lesions and detecting early signs of cancer, ii) enable every patient to assess their lesions, iii) promote melanoma prevention and increase the awareness of this disease, and iv) provide the platform for massive data collection stimulating further research on skin cancer. The fundamental obstacle in advancing automated methods is the lack of large and balanced data sets that can be used to train computational models, i.e. many publicly available skin lesion data sets are small, imbalanced (contain significant disproportions in the number of data points between different classes of lesions and are heavily dominated by the images of benign lesions), and furthermore contain occlusions such as hairs. Publicly available data-sets are obtained from multiple different imaging centers, hospitals, and research institutes, each with different data collection and management standards. Furthermore, some imaging centers mark lesions for example by placing the ruler next to the lesion to measure the diameter of the lesion. Such practices skew the data to comply with the requirements of a particular organization but also introduce bias to the data. We contribute to the computer-aided dermatological techniques with a new set of tools for careful preparation of the training data that mitigate the above mentioned negative data aspects compromising deep network performance in practice. The data preparation consists of the purification and augmentation stages. The process of data purification, \eg removal of occlusions such as hairs and rulers from the images, relies on the efficient lesion segmentation method combined with traditional data processing techniques. The process of data augmentation is two-folded and comprises of balancing the data through the generation of artificial dermoscopic images and performing additional data augmentation. Our tools for data purification and augmentation eventually enhance the performance of deep-learning-based systems for lesion classification. The block diagram of the complete system is captured in Figure~\ref{fig:block}. 

We tested our approach\footnote{Open-sourced under the link \url{https://github.com/devansh20la/Beating-Melanoma} } on popular ISIC-2017 data set~\cite{ISIC} containing lesion images from three classes: melanoma, nevus, and seborrheic keratosis. Additional results on the ISIC 2018~\cite{Tschandl2018_HAM10000} data set containing lesion images from seven classes: acitinic keratosis, basal cell carcinoma, benign keratosis, dermatofibroma, melanoma, nevus, and vascular lesion are provided in the Supplementary section. The paper is organized as follows: Section~\ref{sec:R} describes related work, Section~\ref{sec:data} discusses the data, Section~\ref{sec:DP} explains our approach for data purification that relies on the segmentation step, Section~\ref{sec:DG} addresses the problem of data imbalancedness with a tool based on generative adversarial network for synthetic data generation, and finally Section~\ref{sec:E} incorporates the proposed solutions into a classification network and reports experimental results. Section~\ref{sec:CC} concludes the paper. Supplement contains additional results.

\section{Related Work}
\label{sec:R}
There is a substantial body of work on developing algorithms for computer-assisted dermatology. A number of approaches rely on hand-crafted features~\cite{Barata,DBLP:conf/dicta/YaoWXGF16,DBLP:conf/isbi/BiKAFF16,Yan} and are not scalable to massive data sets.  Recent development of the field of deep learning has triggered a sudden shift from vision algorithms largely relying on expensive manual feature extraction. The development of convolutional neural networks (CNNs)~\cite{lecun}, and in particular AlexNet~\cite{alexnet}, is widely regarded as the turning point in deep learning research. CNN-based approaches later outsmarted humans in several tasks ranging from image classification to playing GO~\cite{go}. Consequently, the techniques for automatic analysis of skin lesions recently became dominated with deep learning algorithms. A striking majority of these approaches are ensemble techniques. They include i) ensembles of different architectures used for extracting data features, which are further inputted to the SVM classifier~\cite{Codella2016DeepLE,DBLP:journals/corr/MenegolaTFLAV17,DBLP:conf/isbi/MenegolaFPBAV17} (deep learning combined with SVM is also used in other approaches~\cite{conf/miccai/CodellaCAGHS15}), ii) segmentation networks followed by bagging-type ensemble strategies~\cite{DBLP:journals/corr/YuanCL17}, iii) segmentation networks that use smooth F1-score loss function to factor-in data imbalancedness~\cite{Kawahara2017FullyCN}, iv) methods that interpolate between the outputs obtained from two fully-convolutional residual networks and refine the results with lesion index calculation unit~\cite{DBLP:journals/corr/LiS17a}, v) highly-complicated ensembles with multi-view image generator~\cite{DBLP:journals/corr/MatsunagaHMK17}, vi) systems based on multiple segmentation networks that provide segmentations conditioned on the pre-defined sets of local patterns that are of interest to clinicians in their diagnosis~\cite{DBLP:journals/corr/Gonzalez-Diaz17}, and finally vii) ensemble techniques in which multiple fully convolutional networks learn complementary visual characteristics of different skin lesions as well as lesion boundaries and an integration method forms a final segmentation from individual segmentations~\cite{Bi2017DermoscopicIS}. Non-ensemble strategies propose using i) U-Net model that relies on CNN with skipped connections~\cite{doi:10.1080/24699322.2017.1389405} for lesion segmentation and also classification and ii) residual network ResNet-50 pre-trained on ImageNet~\cite{Cicero} for lesion classification, or iii) GoogleNet Inception v3 CNN architecture pre-trained on ImageNet for lesion classification~\cite{Esteva2017}, where in the latter paper the experiments are done on a data base that is not publicly available. Only a subset of these deep learning methods perform data augmentation to increase the training data size~\cite{Increasing} or balance the data, but they use fairly standard tools, i.e. image cropping, scaling, and flipping, and none of them addresses the problem of the removal of image occlusions (note that throughout the paper we refer to cropping, scaling and flipping as the standard/traditional augmentation techniques).  

\section{Proposed Methods}

\subsection{Data sets}
\label{sec:data}
We focus on dermoscopic images which allow higher diagnostic accuracy than traditional eye examination~\cite{dermo} as they show the micro-structures in the lesion and provide a view unobstructed from skin reflections. We took $803$ cases of melanoma, $2107$ cases of nevus, and $288$ cases of seborrheic keratosis from the ISIC 2017 challenge dataset, $40$ cases of melanoma and $80$ cases of nevus from the PH$^2$ data set\cite{PH2}, and $76$ cases of melanoma, $331$ cases of nevus and $257$ cases of seborrheic keratosis from the Edinburgh data set \cite{Dermofit} (see Figure~\ref{fig:il}). Diagnoses of these lesions were established clinically (ISIC, Dermofit) and/or via histopathology (ISIC, PH$^{2}$) , which will be discussed more in details later in the paper, regarding the training data size). Additionally, each data set contains lesion segmentation masks. We tested our models on ISIC 2017 and ISIC 2018 (Supplementary section) test data sets obtained from the ISIC Archives. 

\subsection{Data Purification}
\label{sec:DP}

\subsubsection{Data Purification Problem}

\begin{figure}[t]
    \includegraphics[width=0.09\textwidth]{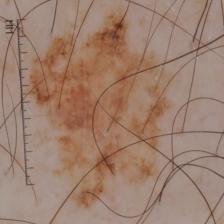}
    \includegraphics[width=0.09\textwidth]{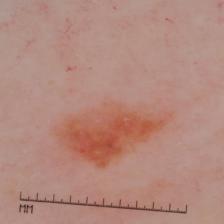}
    \includegraphics[width=0.09\textwidth]{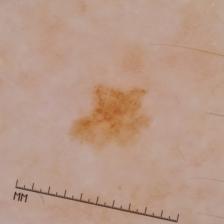}
    \includegraphics[width=0.09\textwidth]{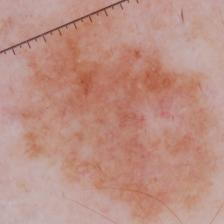}
    \includegraphics[width=0.09\textwidth]{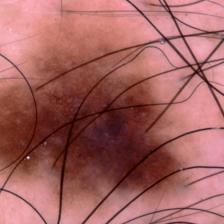}

    \includegraphics[width=0.09\textwidth]{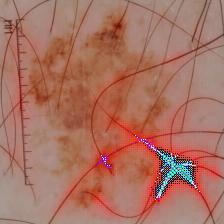}
    \includegraphics[width=0.09\textwidth]{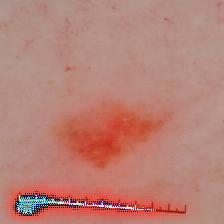}
    \includegraphics[width=0.09\textwidth]{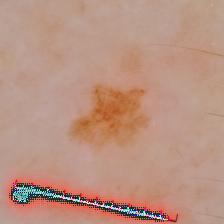}
    \includegraphics[width=0.09\textwidth]{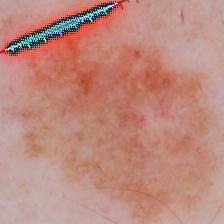}
    \includegraphics[width=0.09\textwidth]{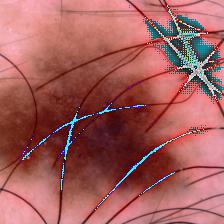}
    \caption{Visualization results for the conventionally-trained model on the ISIC 2017 data set. \textbf{(Top)}: Original image. \textbf{(Bottom)}: Visualization mask overlaid on the original image. The model overfits to image occlusions such as hairs and rulers. }
    \label{fig:visualization}
\end{figure}
Dermoscopic images often contain occlusions such as hairs and/or rulers. Deep learning approaches in general can handle such image occlusions and learn to avoid these objects while making predictions, though their learning ability is conditioned upon the availability of large training data. In case of dermoscopic data bases, which typically have small or medium size, the deep learning models are easily prone to overfitting, i.e. they use visual cues such as hairs and rulers as indicators of the lesion category. To demonstrate this problem we employ the visualization technique, called VisualBackProp \cite{visualization}, that highlights the part of the image that the network focuses on when forming its prediction. Figure~\ref{fig:visualization} shows the results obtained for the traditionally-trained deep model (without performing data purification or augmentation) on the raw data.

\subsubsection{Data Purification Method}

\begin{figure}[t]
    \centering
    \includegraphics[width = 0.09\textwidth]{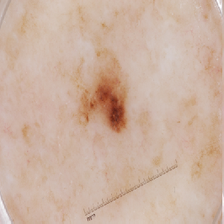}
    \includegraphics[width = 0.09\textwidth]{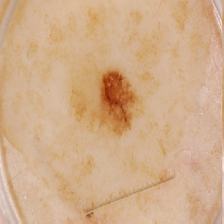}
    \includegraphics[width = 0.09\textwidth]{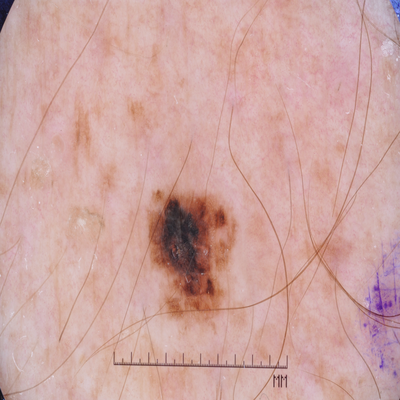}
    \includegraphics[trim={0 0 0 20pt},clip, width = 0.09\textwidth,height=0.09\textwidth]{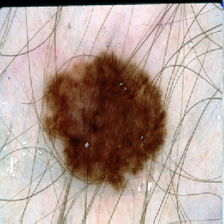}
    \includegraphics[width = 0.09\textwidth]{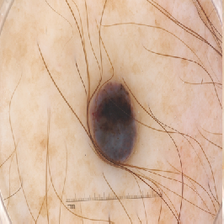}
    
    \stackunder[5pt]{\includegraphics[width = 0.09\textwidth]{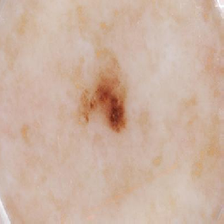}}{(a)}
    \stackunder[5pt]{\includegraphics[width = 0.09\textwidth]{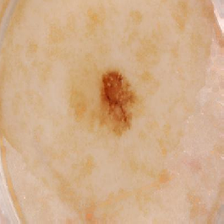}}{(b)}
    \stackunder[5pt]{\includegraphics[width = 0.09\textwidth]{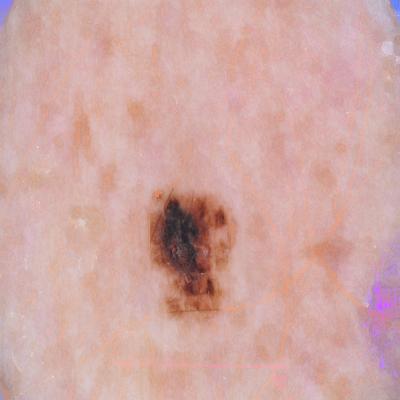}}{(c)}
    \stackunder[5pt]{\includegraphics[trim={0 0 0 20pt},clip, width = 0.09\textwidth,height=0.09\textwidth]{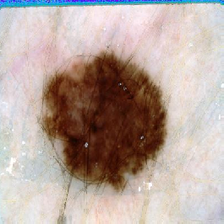}}{(d)}
    \stackunder[5pt]{\includegraphics[width = 0.09\textwidth]{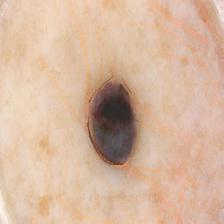}}{(e)}
    
    \caption{\textbf{Top}: Original images. \textbf{Bottom}: Images obtained after a,b) scales, c) hairs and scales, and d,e) hairs removal.}
    \label{hair_removal}
\end{figure}

We utilize traditional data processing approaches to find and remove hairs and rulers on the images. We extended the hair-removal algorithm~\cite{hair_removal}. One of its steps involves thresholding the luminance channel of the image in the LUV color space, which may also remove dark regions belonging to the lesion itself. To correct that, we overlay the processed image with the segmented lesion obtained from our segmentation network that will be described later. The holes in the background were eliminated by performing the closing operation. We compare two different input images before and after pre-processing to judge the quality of the purification algorithm. This is captured in Figure \ref{hair_removal}. We preprocessed $470$ cases of nevus, $147$ cases of melanoma, and $182$ cases of seborrheic keratosis, which contained hairs and rulers and removed those from the images. The pre-processed images were added to the training data set of the lesion classification model to make it more robust to the presence of occlusions and prevent over-fitting. 

\color{black}

\begin{table*}[t]
    \centering
    \renewcommand{\arraystretch}{1.0}
    \begin{tabular}{|c|p{6em}|p{7em}|l|l|l|l|}
        \hline
        & Layer & Output size & Kernel & Stride & Padding & Dilation \\
        \hline
        \multirow{13}{7em}{Downsampling}
        & Conv 		& 64$\times$378$\times$378 	& 3$\times$3 	& 1 & 0 & 1 \\
        & Conv 		& 64$\times$376$\times$376 	& 3$\times$3 	& 1 & 0 & 1 \\
        & Maxpool 	& 64$\times$188$\times$188 	& 2$\times$2	& 2 & 0 & 1 \\
        & Conv 		& 128$\times$186$\times$186 & 3$\times$3 	& 1 & 0 & 1 \\
        & Conv 		& 128$\times$184$\times$184 & 3$\times$3 	& 1 & 0 & 1 \\
        & Maxpool 	& 128$\times$92$\times$92 	& 2$\times$2 	& 2 & 0 & 1 \\
        & Conv 		& 256$\times$90$\times$90 	& 3$\times$3 	& 1 & 0 & 1 \\
        & Conv 		& 256$\times$88$\times$88 	& 3$\times$3 	& 1 & 0 & 1 \\
        & Maxpool 	& 256$\times$44$\times$44	& 2$\times$2 	& 2 & 0 & 1 \\
        & Conv 		& 512$\times$40$\times$40 	& 3$\times$3 	& 1 & 0 & 2 \\
        & Conv 		& 512$\times$36$\times$36 	& 3$\times$3 	& 1 & 0 & 2 \\
        & Conv 		& 1024$\times$28$\times$28 	& 3$\times$3	& 1 & 0 & 4 \\
        & Conv 		& 1024$\times$20$\times$20 	& 3$\times$3	& 1 & 0 & 4 \\
        \hline
        \multirow{14}{7em}{Up-sampling}
        & Trans-Conv 	& 1024$\times$28$\times$28 	& 3$\times$3	& 1 & 0& 4 \\
        & Trans-Conv 	& 512$\times$36$\times$36 	& 3$\times$3	& 1 & 0& 4 \\
        & Trans-Conv 	& 512$\times$40$\times$40 	& 3$\times$3 	& 1 & 0& 2 \\
        & Trans-Conv 	& 512$\times$44$\times$44 	& 3$\times$3 	& 1 & 0& 2 \\
        & Up-Conv 		& 256$\times$88$\times$88 	& 3$\times$3 	& 1 & 1& 1 \\
        & Trans-Conv 	& 256$\times$90$\times$90 	& 3$\times$3 	& 1 & 0& 1 \\
        & Trans-Conv 	& 256$\times$92$\times$92 	& 3$\times$3 	& 1 & 0& 1 \\
        & Up-Conv 		& 128$\times$184$\times$184 & 3$\times$3 	& 1 & 1& 1 \\
        & Trans-Conv 	& 128$\times$186$\times$186 & 3$\times$3 	& 1 & 0& 1 \\
        & Trans-Conv 	& 128$\times$188$\times$188 & 3$\times$3 	& 1 & 0& 1 \\
        & Up-Conv 		& 64$\times$376$\times$376 	& 3$\times$3	& 1 & 1& 1 \\
        & Trans-Conv 	& 64$\times$378$\times$378 	& 3$\times$3 	& 1 & 0& 1 \\
        & Trans-Conv 	& 64$\times$380$\times$380 	& 3$\times$3 	& 1 & 0& 1 \\
        & Conv 	        & 1$\times$380$\times$380 	& 1$\times$1 	& 1 & 0& 1 \\
        \hline
    \end{tabular}
    \caption{Details of the architecture of the proposed segmentation network. Let $n$ be the number of features maps, $h$ be the height and $w$ be the width. Size of the output feature maps is represented as $n\times h \times w$. Each convolution layer, except for the last one, is followed by a ReLU nonlinearity. Up-Conv performs upsampling by a factor of $2$ followed by a convolution operation. The last convolution layer is followed by a sigmoid.}
    \label{tab:U-Net}
    \vspace{-0.15in}
\end{table*}

\subsubsection{Segmentation Network}
\label{sec:SN}
\begin{figure}[htp!]
    (A)\begin{minipage}{0.44\textwidth}
    \centering
    \includegraphics[width=\textwidth]{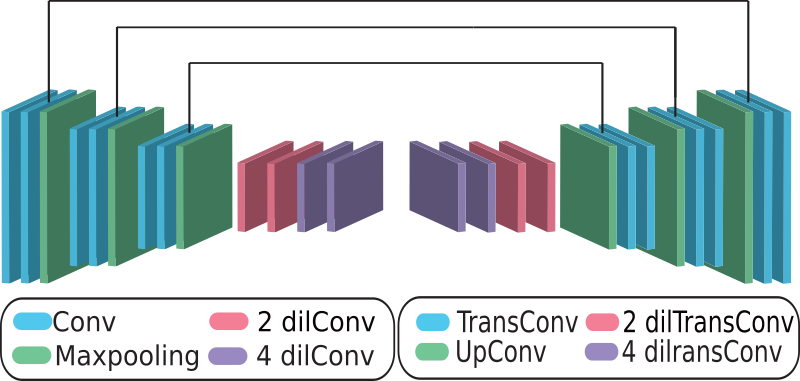}
    \end{minipage}\\
    (B)\begin{minipage}{0.44\textwidth}
    \centering
    \includegraphics[trim={4pt 4pt 4pt 4pt},clip,height=0.18\textwidth,width=0.18\textwidth]{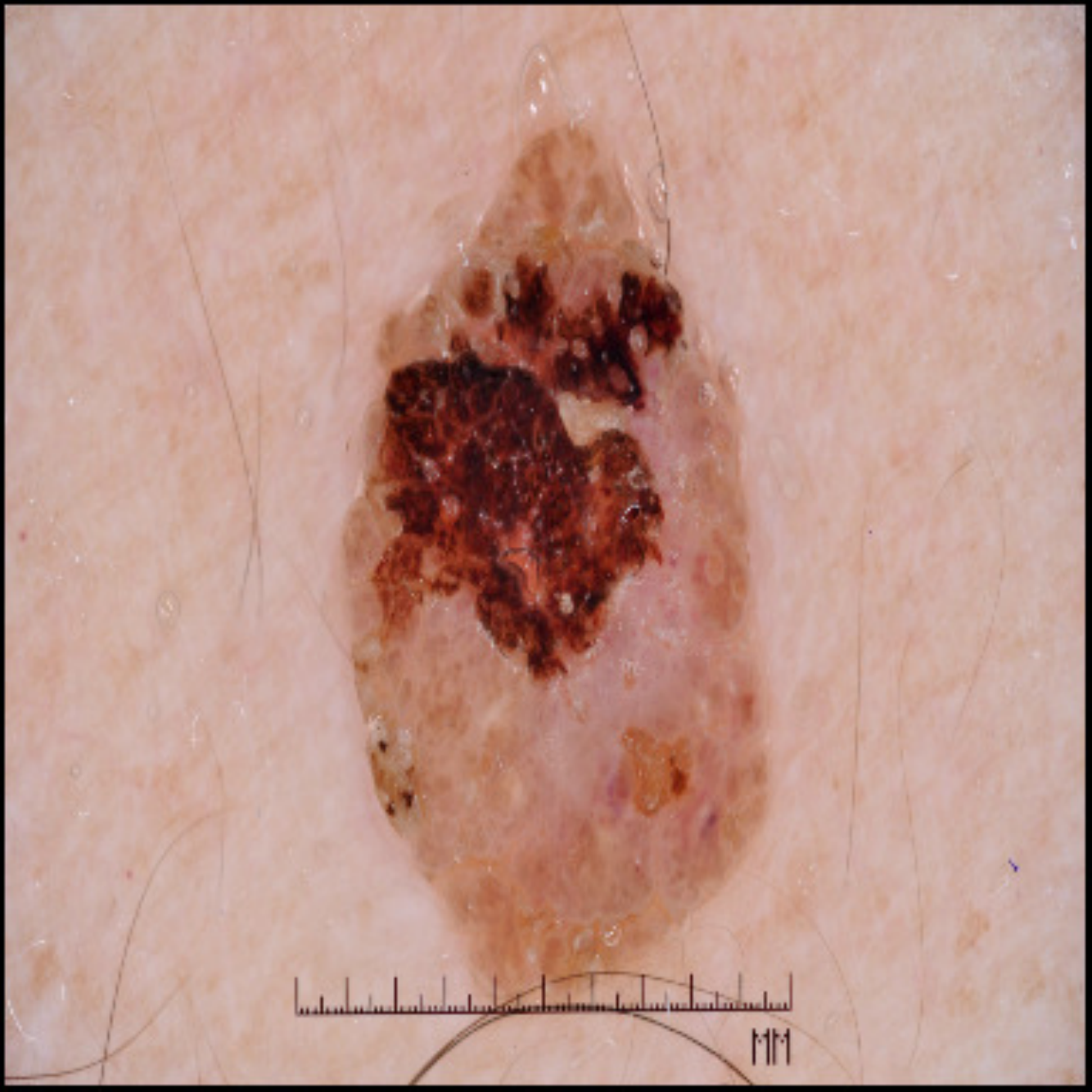}
    \includegraphics[trim={4pt 4pt 4pt 4pt},clip,height=0.18\textwidth,width=0.18\textwidth]{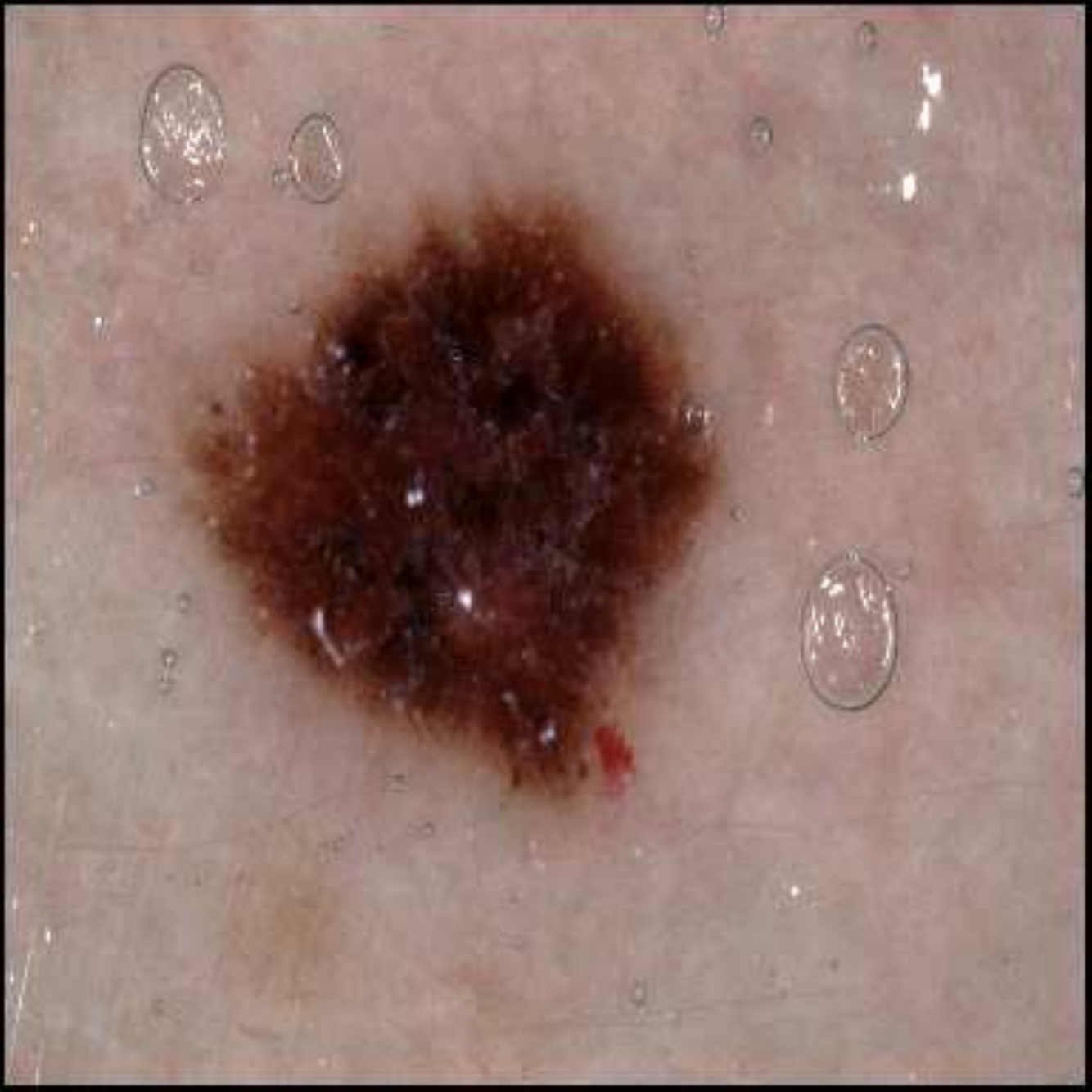}
    \includegraphics[trim={4pt 4pt 4pt 4pt},clip,height=0.18\textwidth,width=0.18\textwidth]{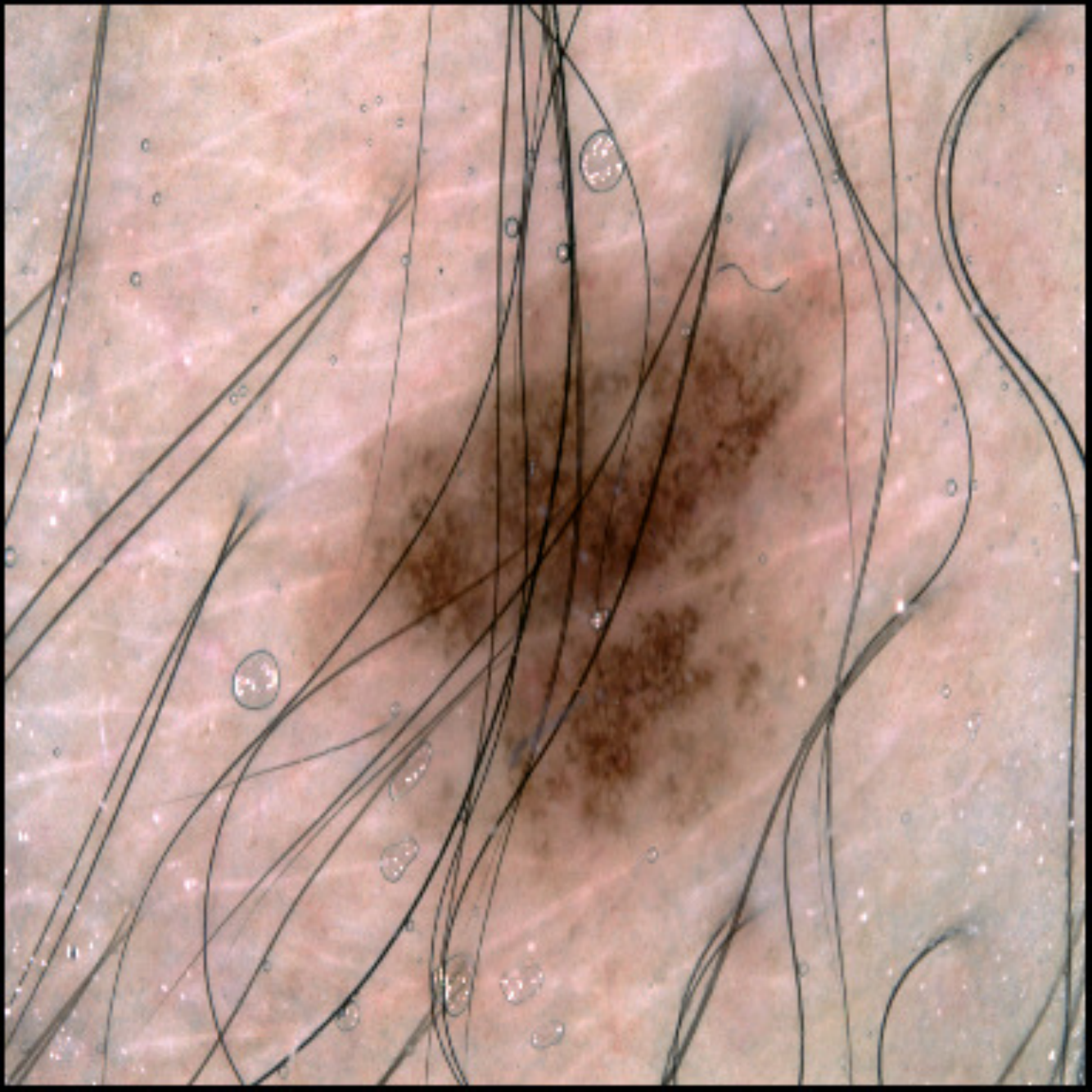}
    \includegraphics[trim={4pt 4pt 4pt 4pt},clip,height=0.18\textwidth,width=0.18\textwidth]{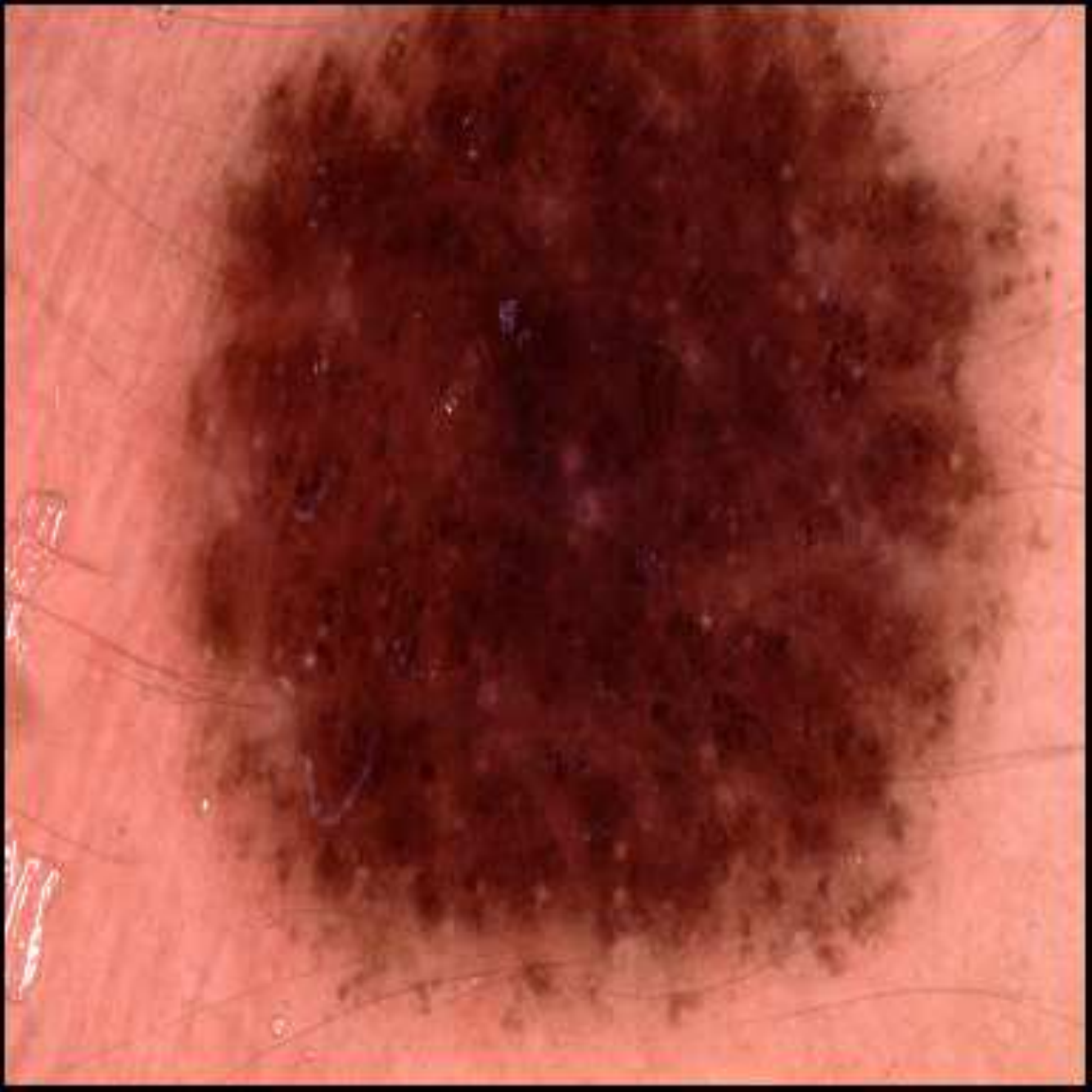}
    \includegraphics[trim={4pt 4pt 4pt 4pt},clip,height=0.18\textwidth,width=0.18\textwidth]{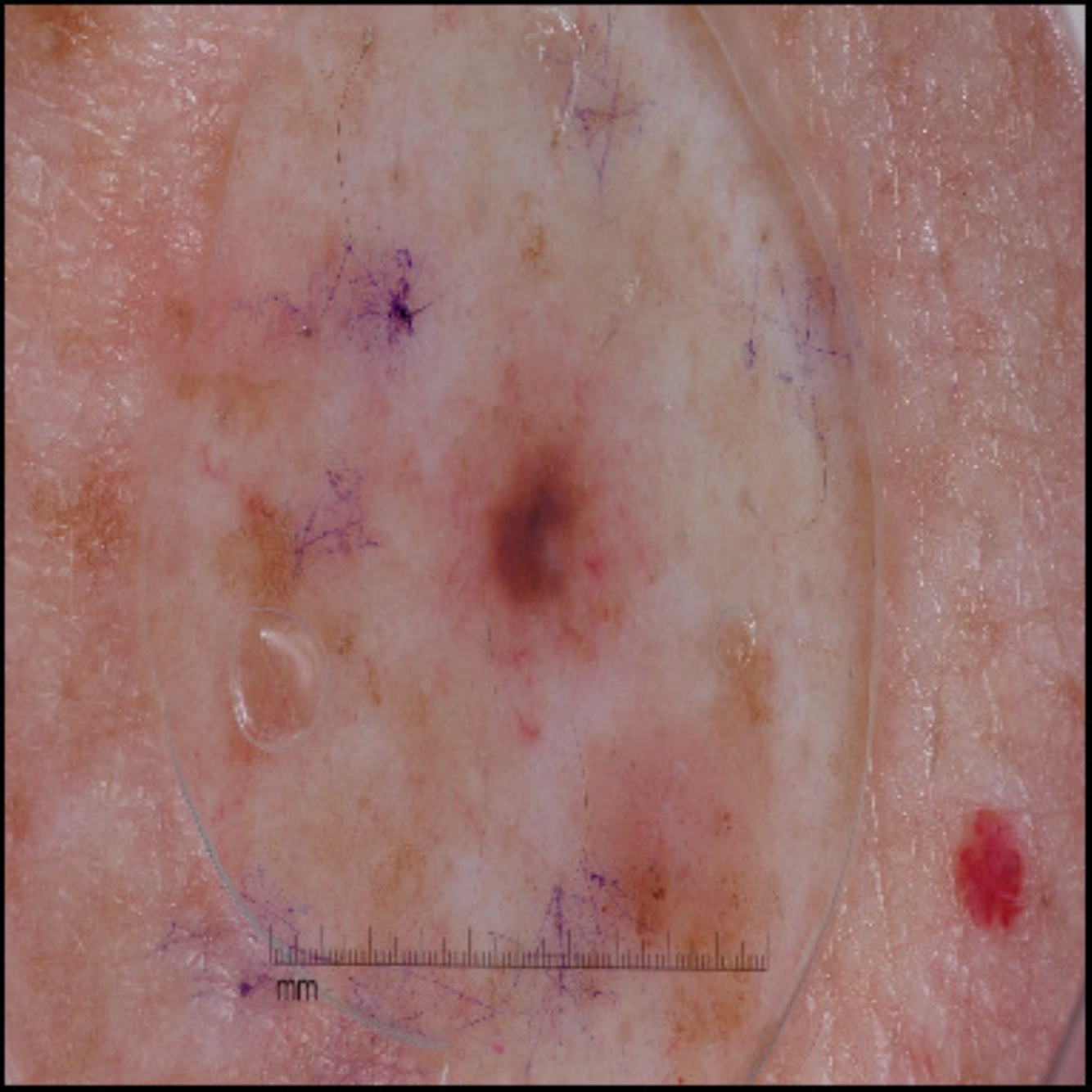}
    
    \includegraphics[trim={4pt 4pt 4pt 4pt},clip,height=0.18\textwidth,width=0.18\textwidth]{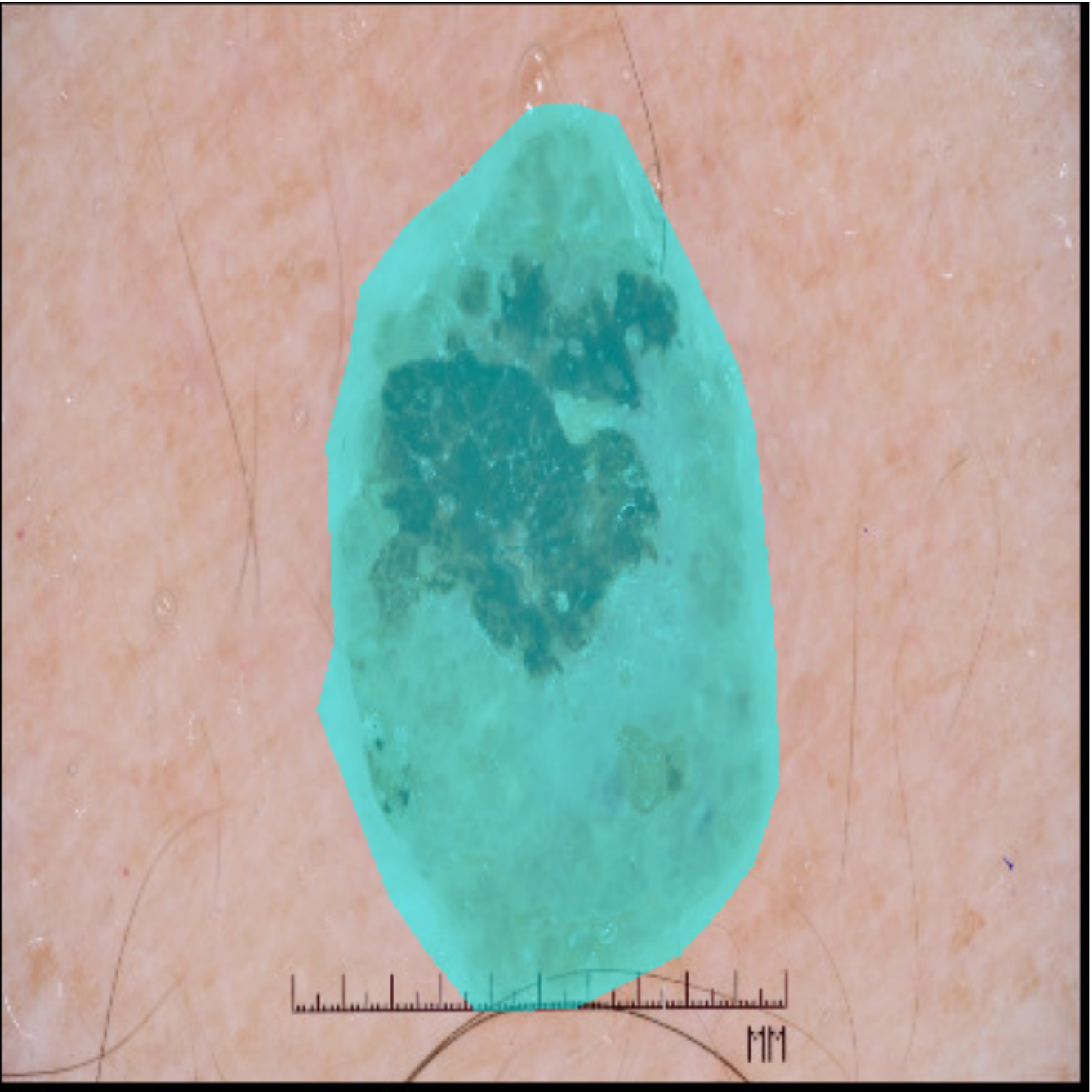}
    \includegraphics[trim={4pt 4pt 4pt 4pt},clip,height=0.18\textwidth,width=0.18\textwidth]{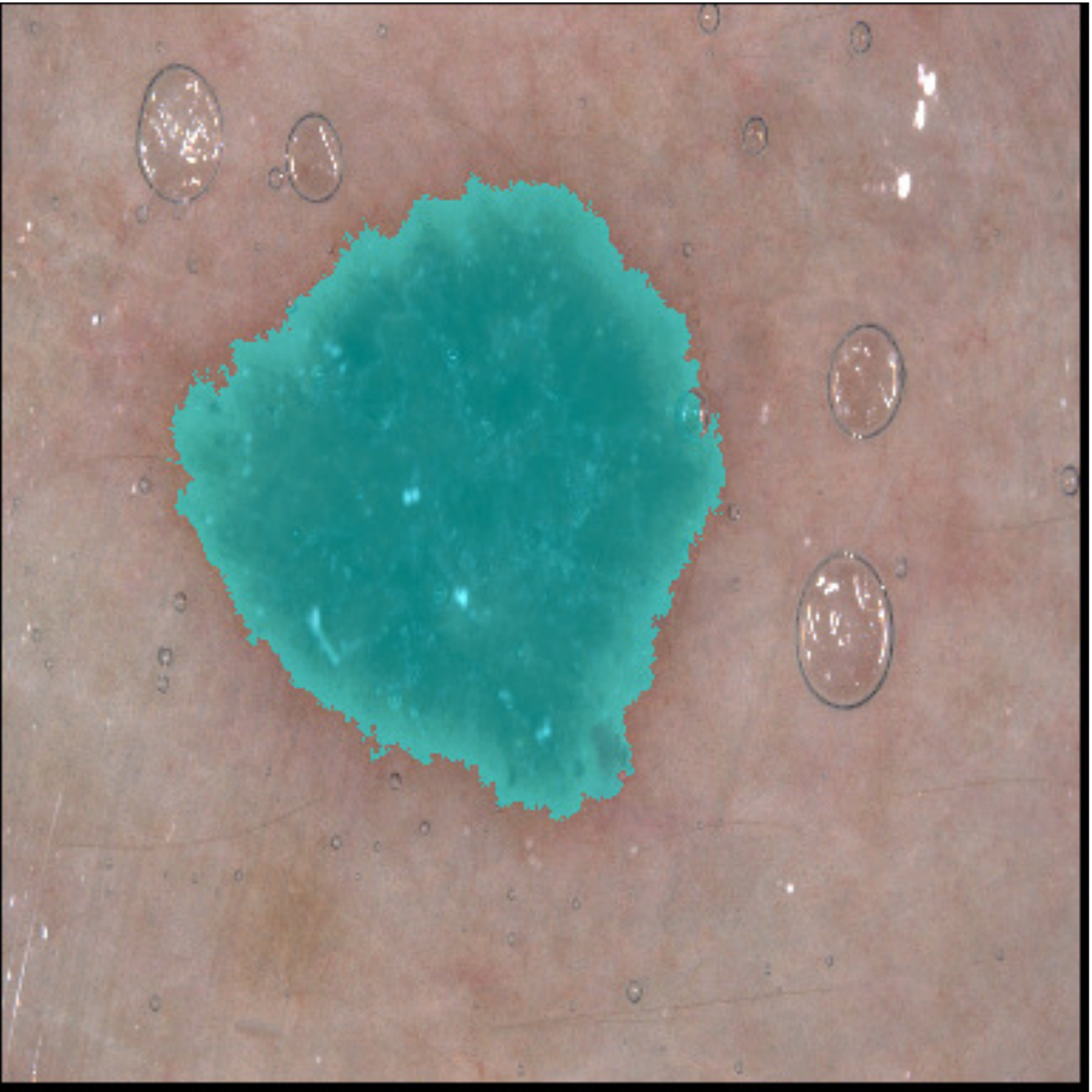}
    \includegraphics[trim={4pt 4pt 4pt 4pt},clip,height=0.18\textwidth,width=0.18\textwidth]{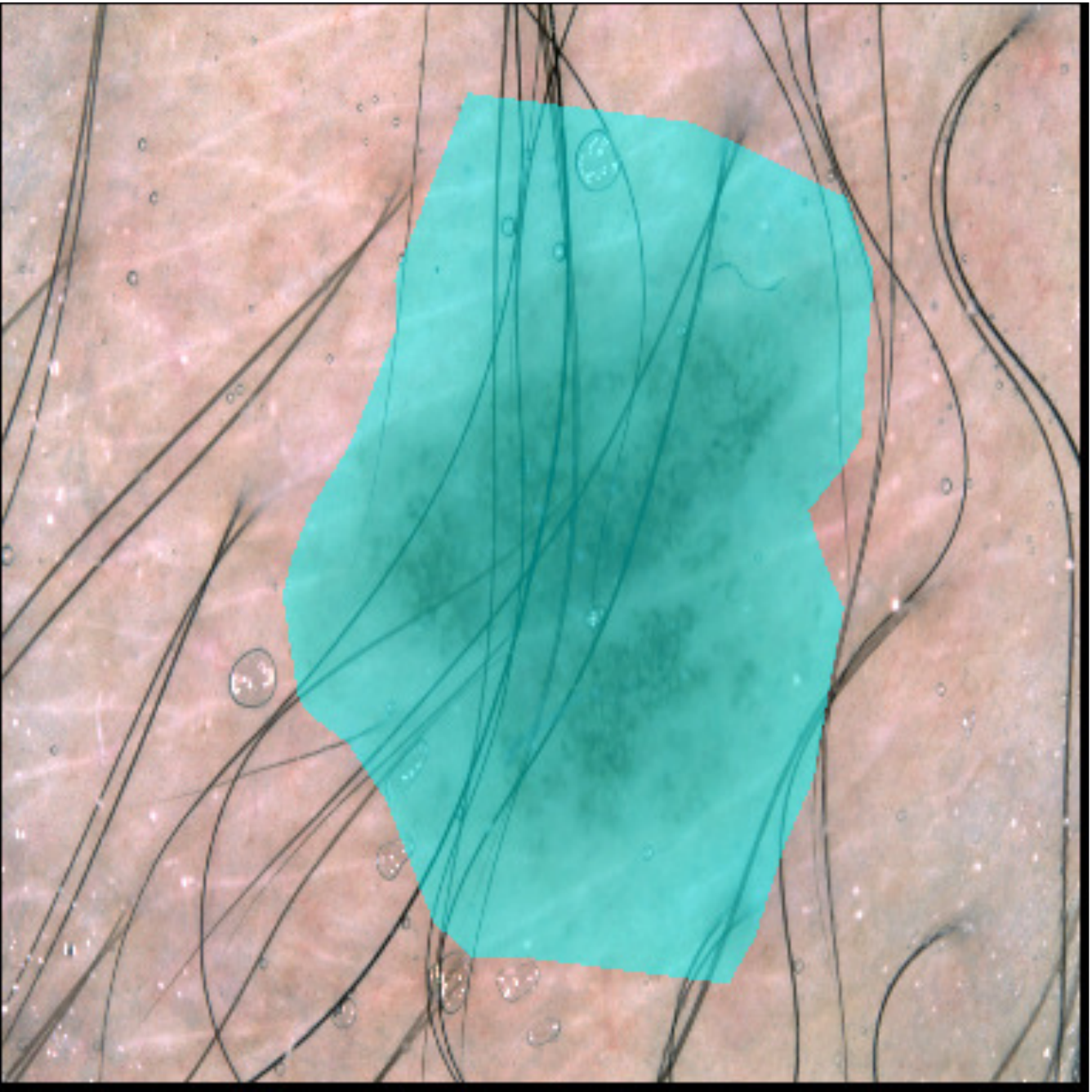}
    \includegraphics[trim={4pt 4pt 4pt 4pt},clip,height=0.18\textwidth,width=0.18\textwidth]{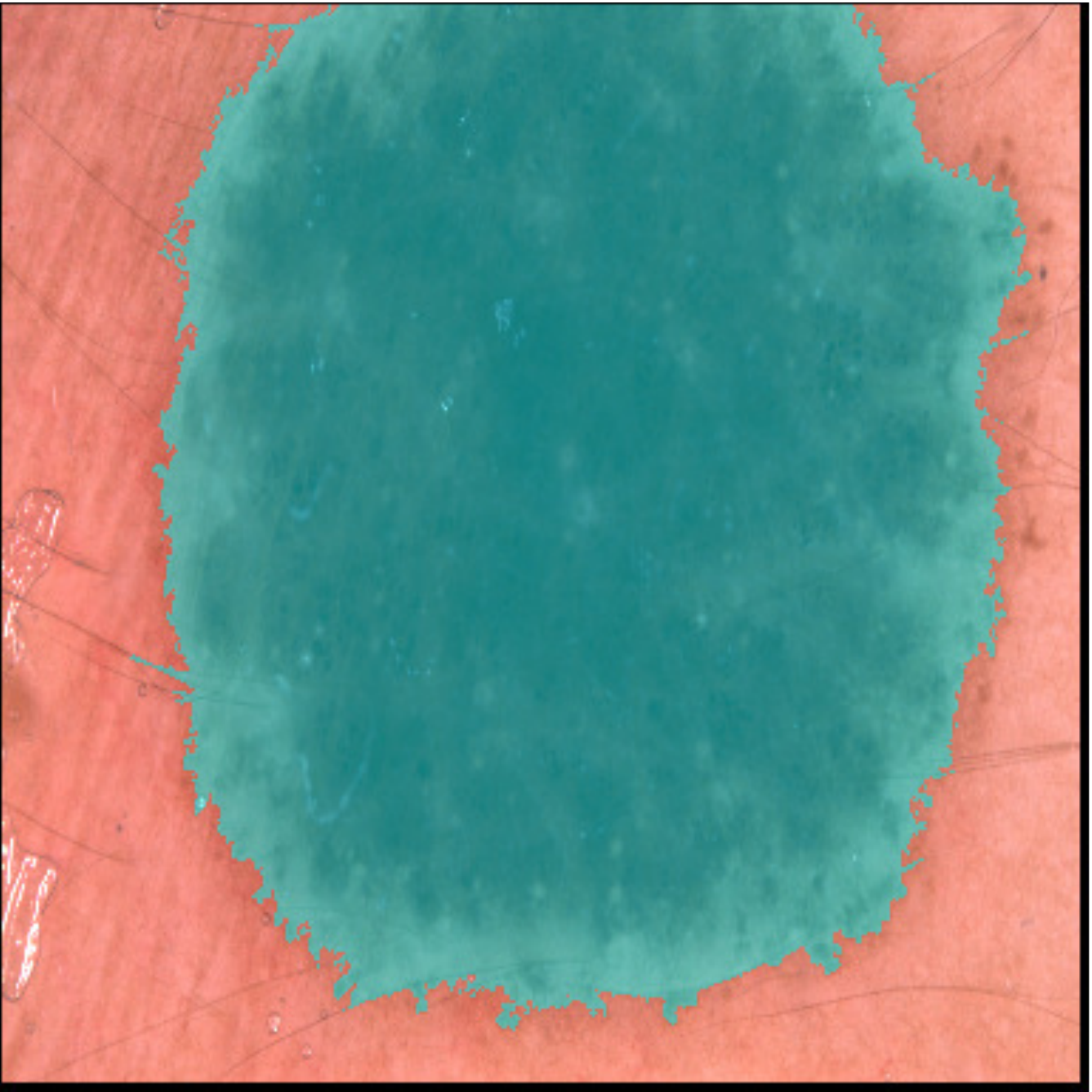}
    \includegraphics[trim={4pt 4pt 4pt 4pt},clip,height=0.18\textwidth,width=0.18\textwidth]{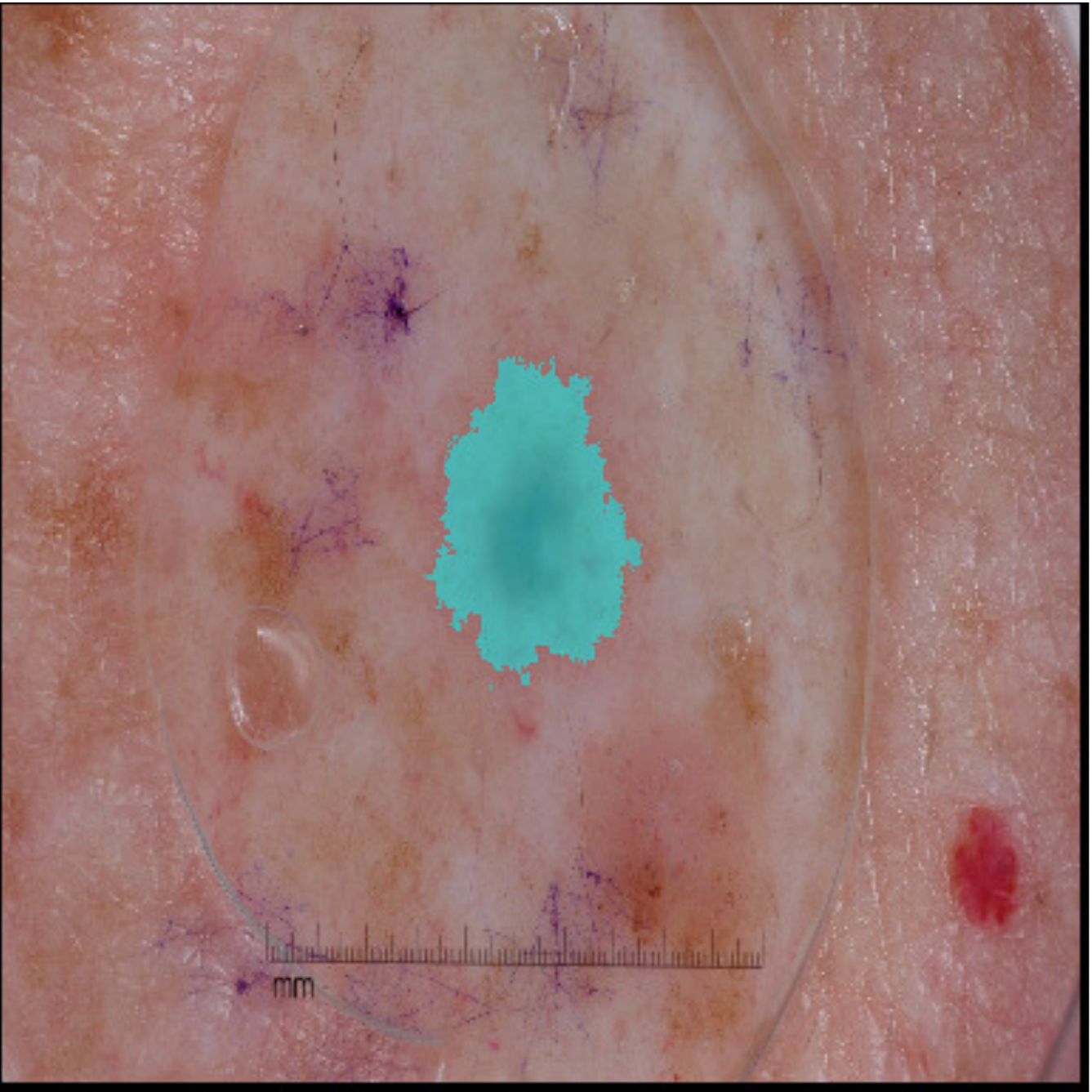}
    
    \includegraphics[trim={4pt 4pt 4pt 4pt},clip,height=0.18\textwidth,width=0.18\textwidth]{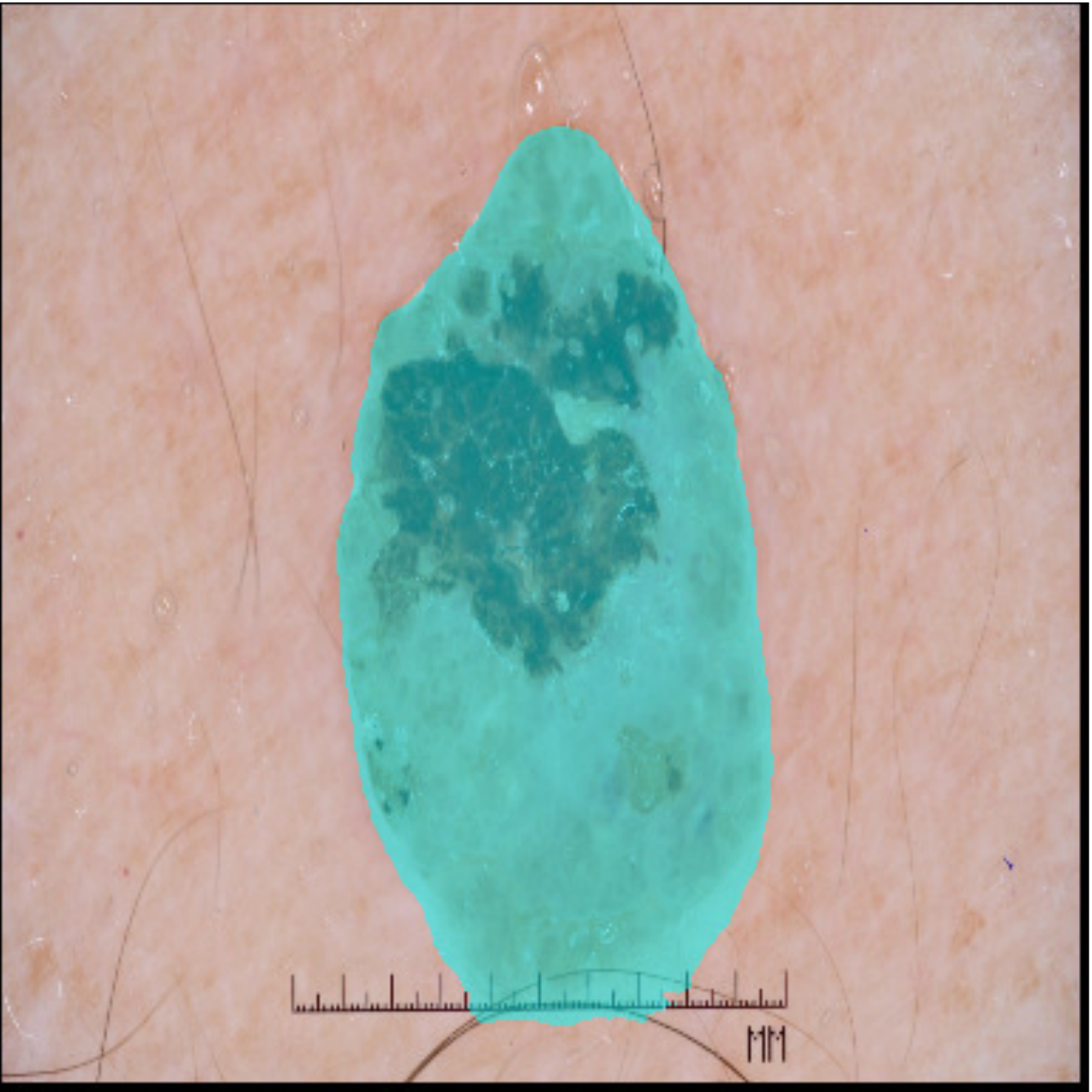}
    \includegraphics[trim={4pt 4pt 4pt 4pt},clip,height=0.18\textwidth,width=0.18\textwidth]{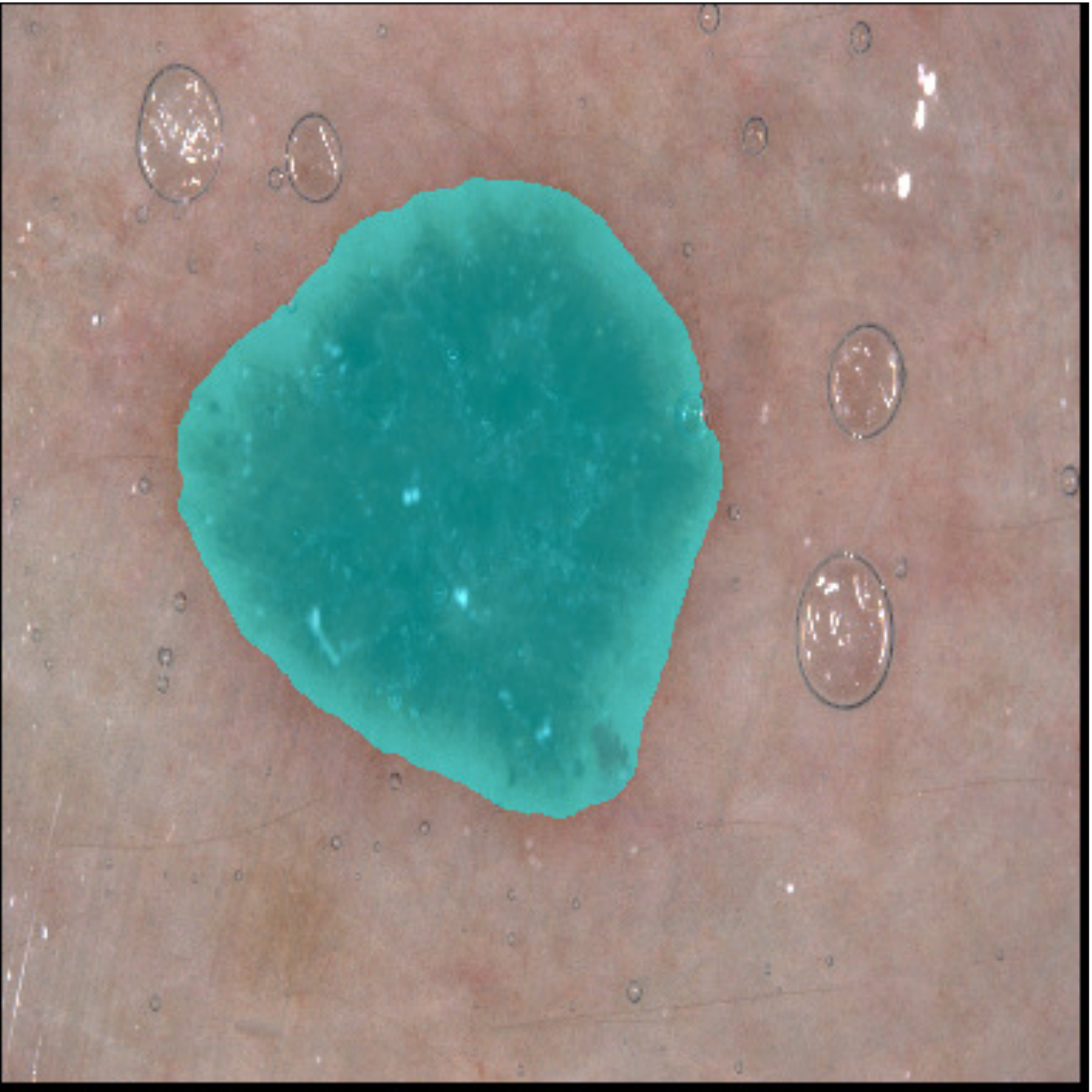}
    \includegraphics[trim={4pt 4pt 4pt 4pt},clip,height=0.18\textwidth,width=0.18\textwidth]{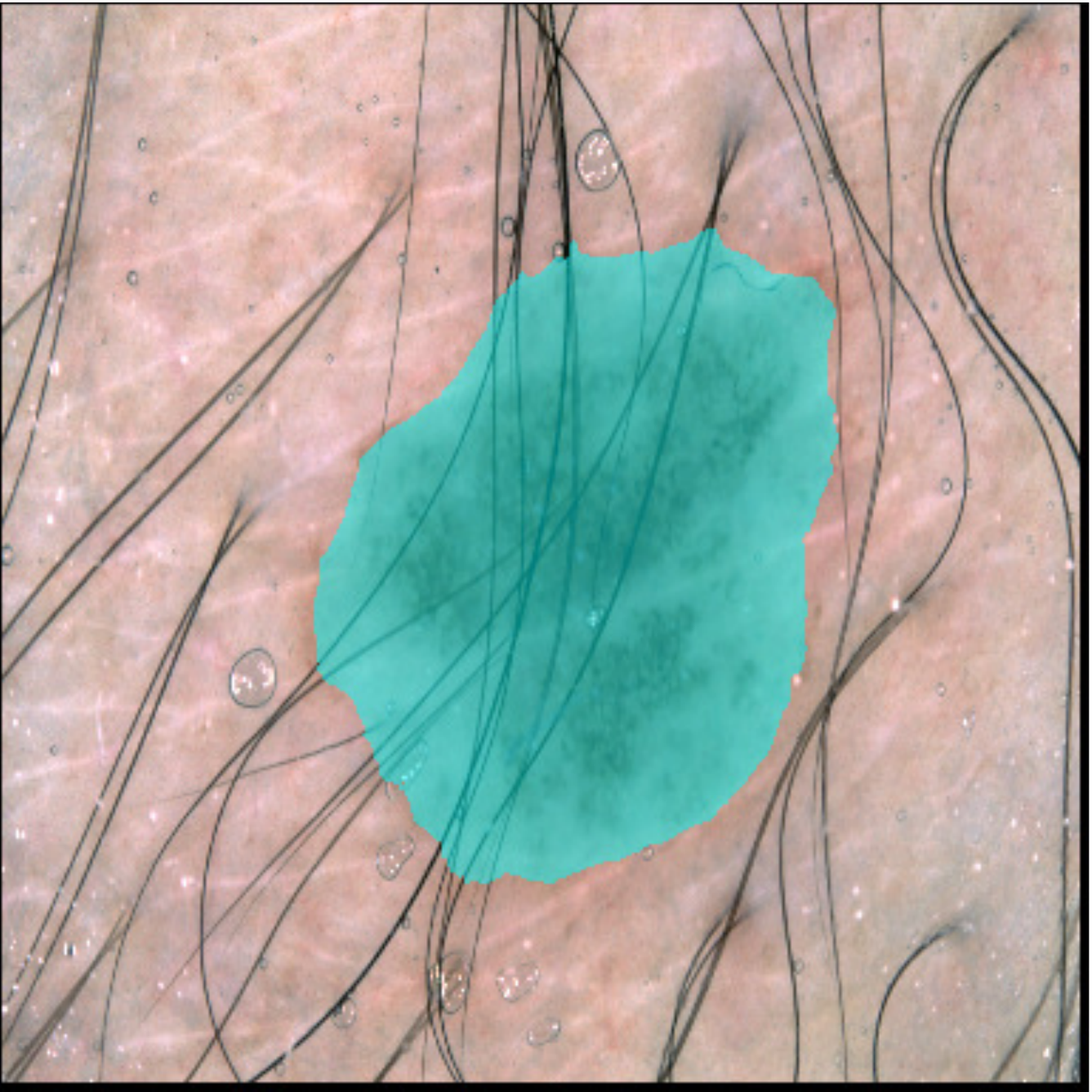}
    \includegraphics[trim={4pt 4pt 4pt 4pt},clip,height=0.18\textwidth,width=0.18\textwidth]{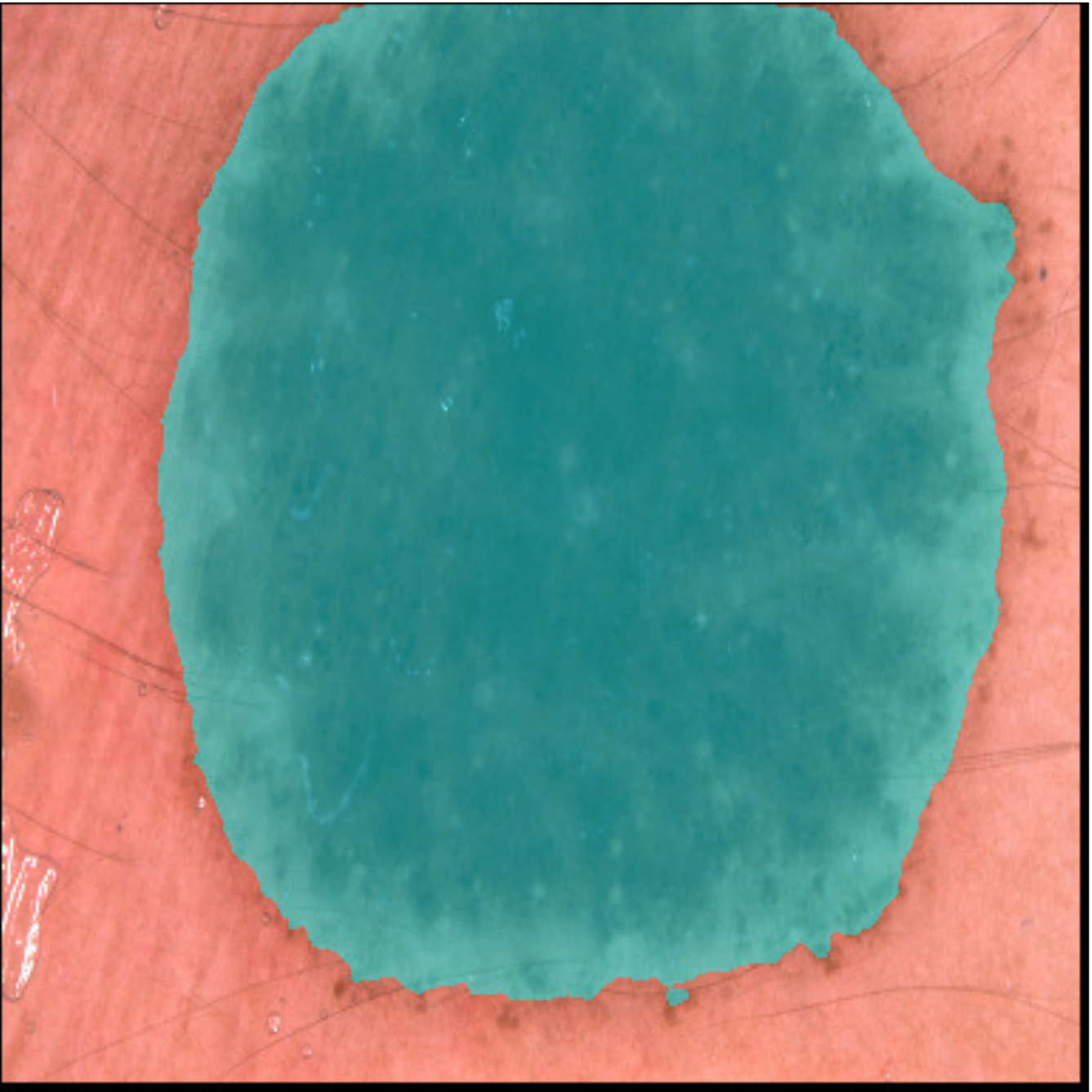}
    \includegraphics[trim={4pt 4pt 4pt 4pt},clip,height=0.18\textwidth,width=0.18\textwidth]{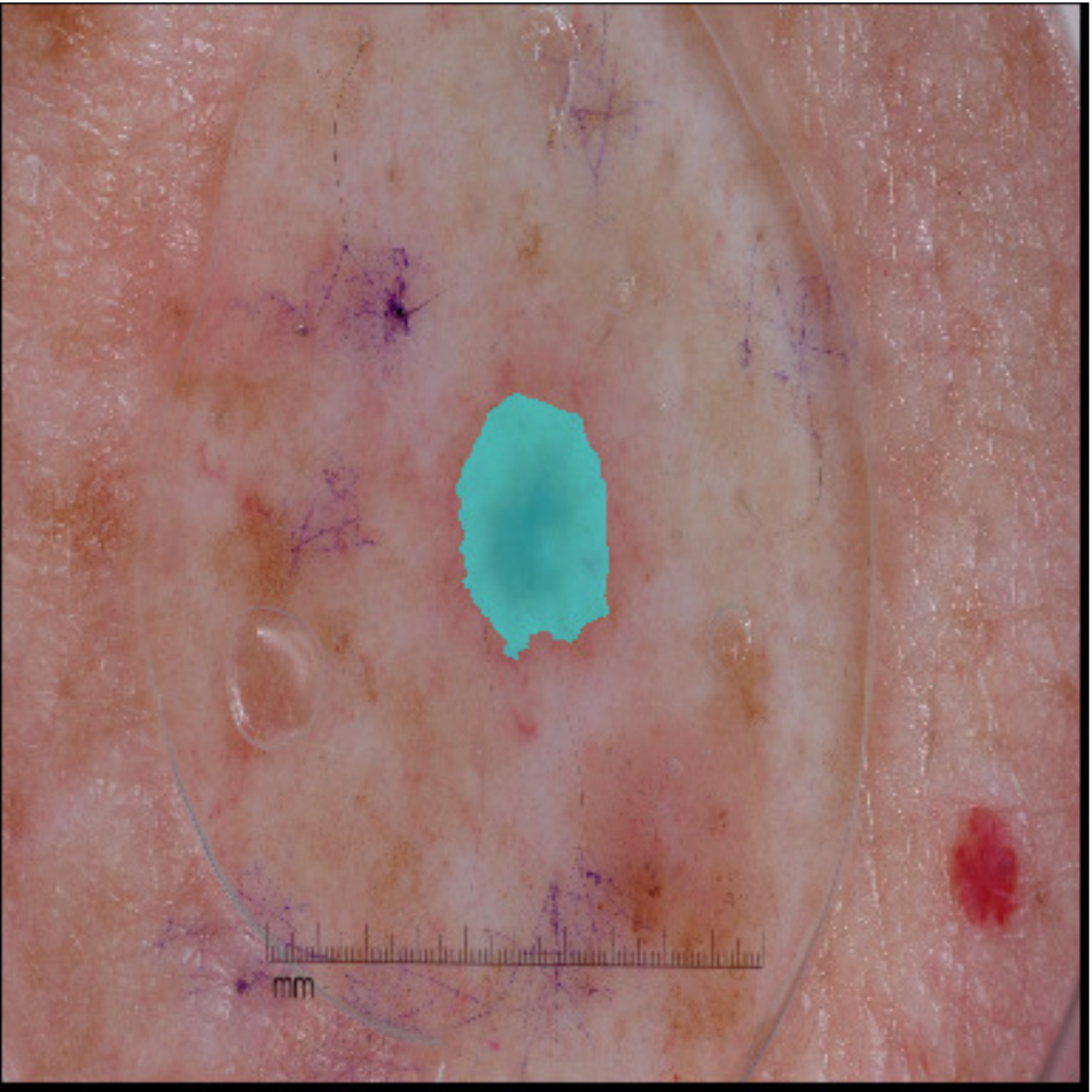}
    \end{minipage}
    \caption{\textbf{(A)} Architecture of the  Segmentation Network. Black lines represent skipped connections. Each convolution operation is followed by ReLU activation; 2 dil and 4 dil refers to resp. 2 and 4 dilated convolution operations.
    \textbf{(B)} \textbf{Top}: Original images from the test data. \textbf{Middle}: Segmentation masks overlaid on the original images. \textbf{Bottom}: Corresponding segmentation masks obtained from Segmentation Network overlaid on the original images.}
    \label{fig:seg}
    \vspace{-0.15in}
\end{figure}

For the lesion segmentation, we utilized a U-Net architecture \cite{unet}, which was previously successfully applied to the problem of medical image segmentation and won the ISBI Cell Tracking Challenge \cite{ISBI} in 2015. We modified the U-Net architecture, since it contains max pooling layers, which cause the increase of the field of view of convolutional filters as the input propagates through the model but at the same time also reduces the resolution of the input images. We however require output and input of the segmentation network to have the same dimension. To overcome this problem, we added dilated \cite{dilated} convolutions in the final layers of the down-sampling block of our architecture. We furthermore removed the up-sampling blocks in the traditional U-Net architecture and instead added transposed convolutional layers. The architecture of the model is presented in Figure~\ref{fig:seg} and Table~\ref{tab:U-Net}. The input to the segmentation architecture consists of seven channels (RGB and HSV channels and luminance). Thus the network input image size is $7\times 380\times 380$. We normalized each channel of the input to mean value $0.5$ and standard deviation $0.5$. We used a binary cross-entropy loss to train the network and an Adam \cite{adam} optimizer with a global learning rate of $1e^{-4}$ and beta values of $(0.9,0.999)$. These settings of optimization hyper-parameters led to the best performance. Furthermore, the output of the segmentation model was passed through a binary hole filling procedure to fill empty holes in the segmentation mask. The segmentation masks obtained for test lesions are demonstrated in Figure~\ref{fig:seg}. To evaluate the performance of the segmentation model the ISIC challenge proposes to use the Jaccard index. The Jaccard index, also known as the intersection over union is used to measure the similarity between two sample sets and is defined as the size of the intersection over the size of the union of the two sets. To calculate the index for segmentation masks we evaluate the area of the overlap between the true label and model output and divide it by the area of the union between them. We obtained an average index of $0.77$ while the best performer of the ISIC 2017 challenge obtained $0.765$.

\subsection{Data Augmentation}
\label{sec:DG}
\subsubsection{Data Imbalancedness Problem}
The data imbalancedness, illustrated in Figure~\ref{fig:il}, is yet another significant factor that deteriorates the performance of deep learning systems analyzing dermoscopic images. The classifiers tend to be biased towards majority classes that correspond to benign lesions. This problem can be partially mitigated by introducing higher penalty for misclassifying rare classes, though data augmentation techniques replace this approach nowadays as they have the advantage of increasing data variability while balancing the data. We propose to use data augmentation technique that relies on generating the images from scarce classes that obey the data distribution of these classes. This is equivalent to creating virtual patients with lesions from scarce classes in order to even their size with the large-size classes.

\subsubsection{Data Generation Network with de-coupled DCGANs}
\begin{figure}[h]
    \centering
    (A)\begin{minipage}{0.45\textwidth}
    \includegraphics[width=\textwidth]{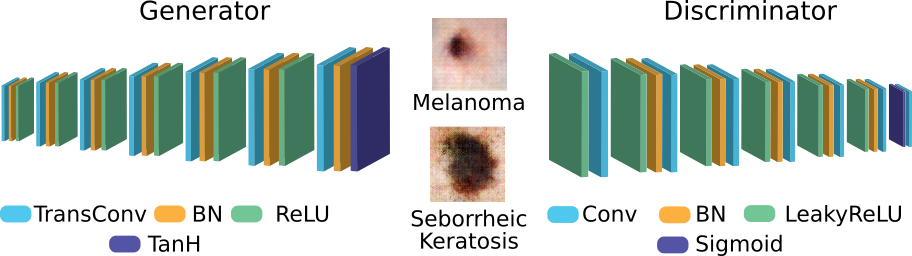}
    \end{minipage}\\
    (B)\begin{minipage}{0.45\textwidth}
    \centering
    \includegraphics[width=0.49\textwidth]{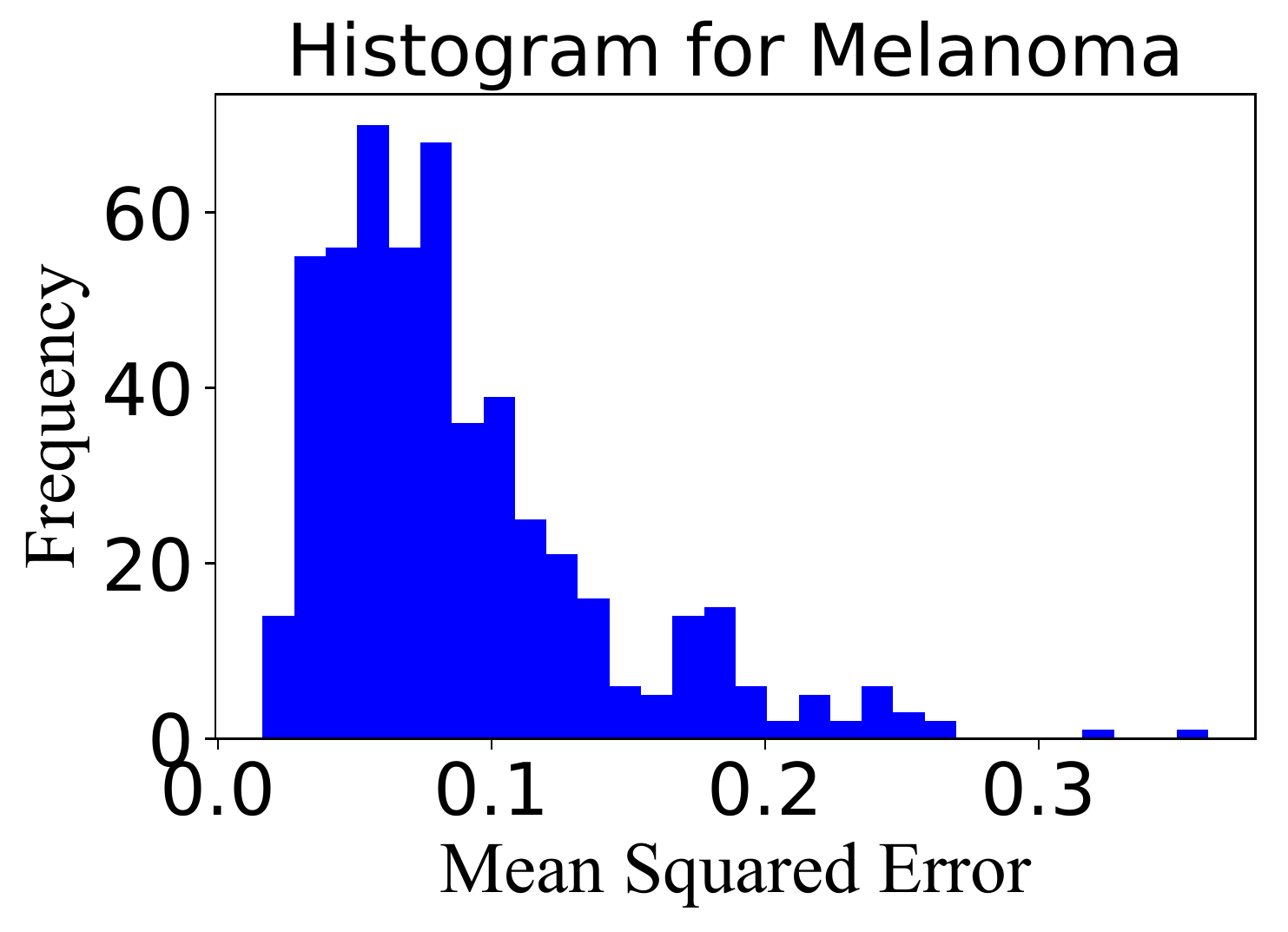}
    \includegraphics[width=0.49\textwidth]{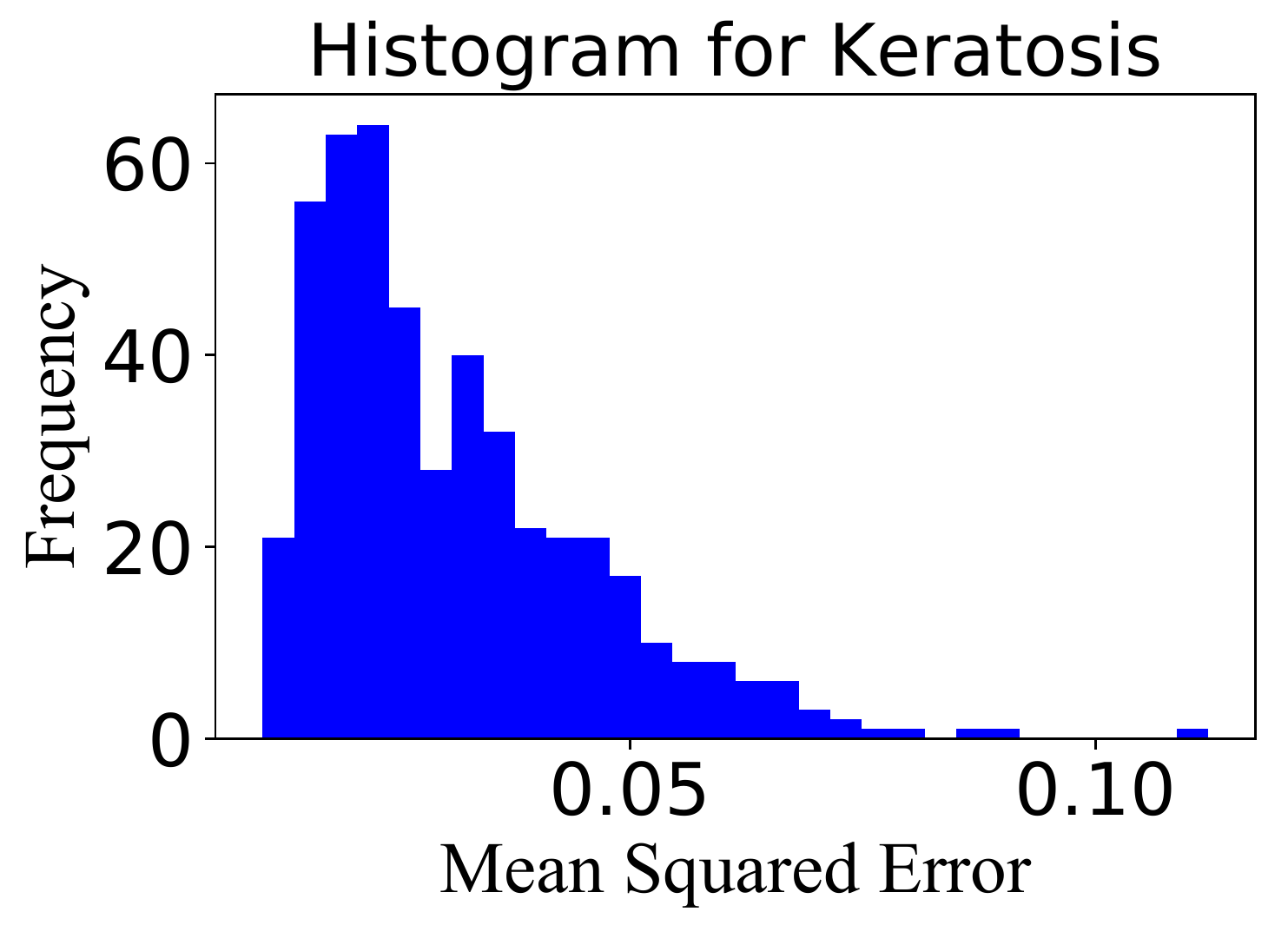}
    \end{minipage}\\
    \caption{\textbf{(A)} Architecture of the DCGAN model. \textbf{(B)} Histograms of the MSE values for \textbf{(left)} melanoma (the mean and std of the MSE are  $0.088 \pm  0.052$) \textbf{(right)} seborrheic keratosis (the mean and std of the MSE are  $0.030 \pm 0.015$).}
    \label{fig:histo}
\end{figure}
\begin{figure}
    \centering
    (A)\begin{minipage}{0.45\textwidth}
    \begin{figure}[H]
    \includegraphics[width=0.19\textwidth]{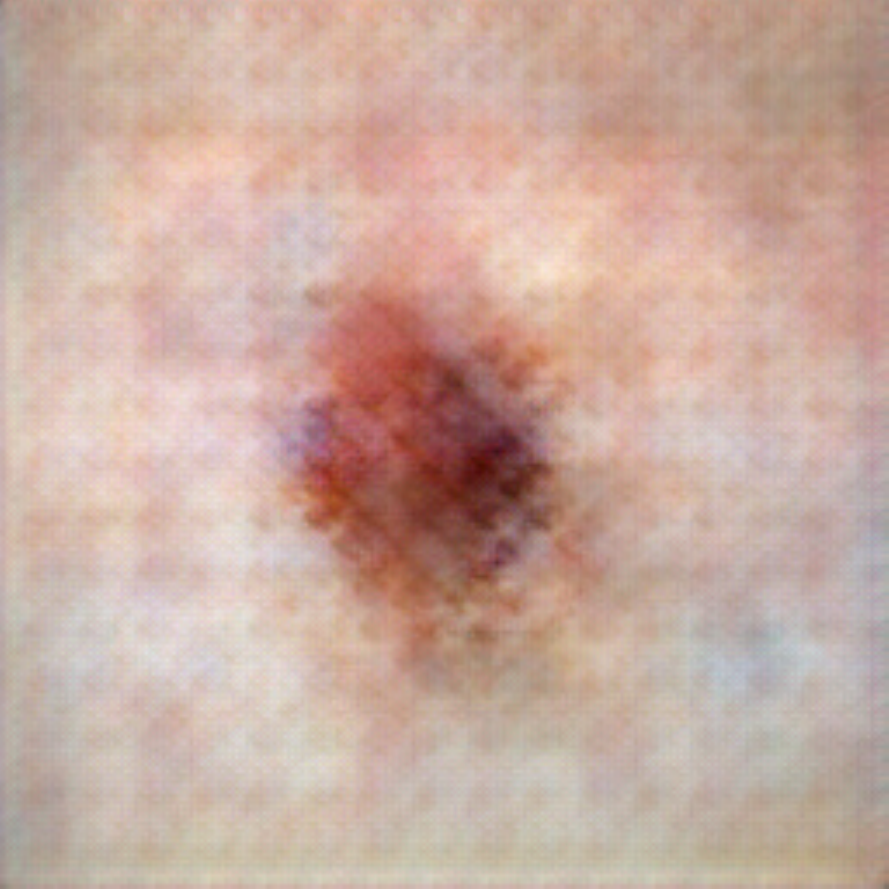}
    \includegraphics[width=0.19\textwidth]{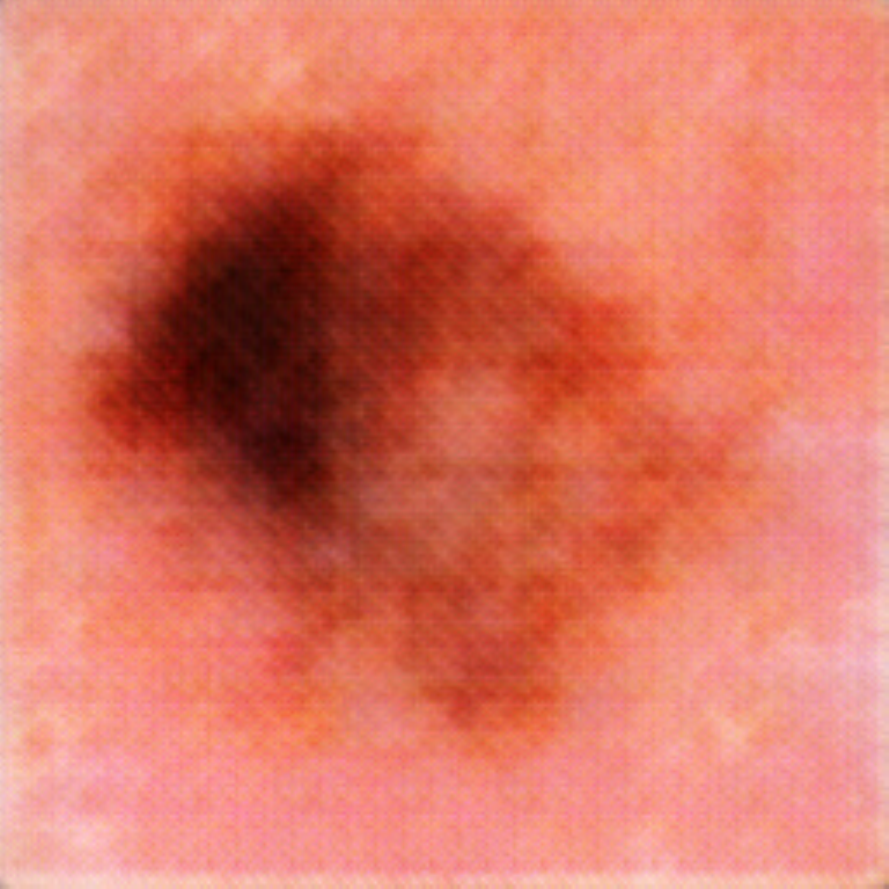}
    \includegraphics[width=0.19\textwidth]{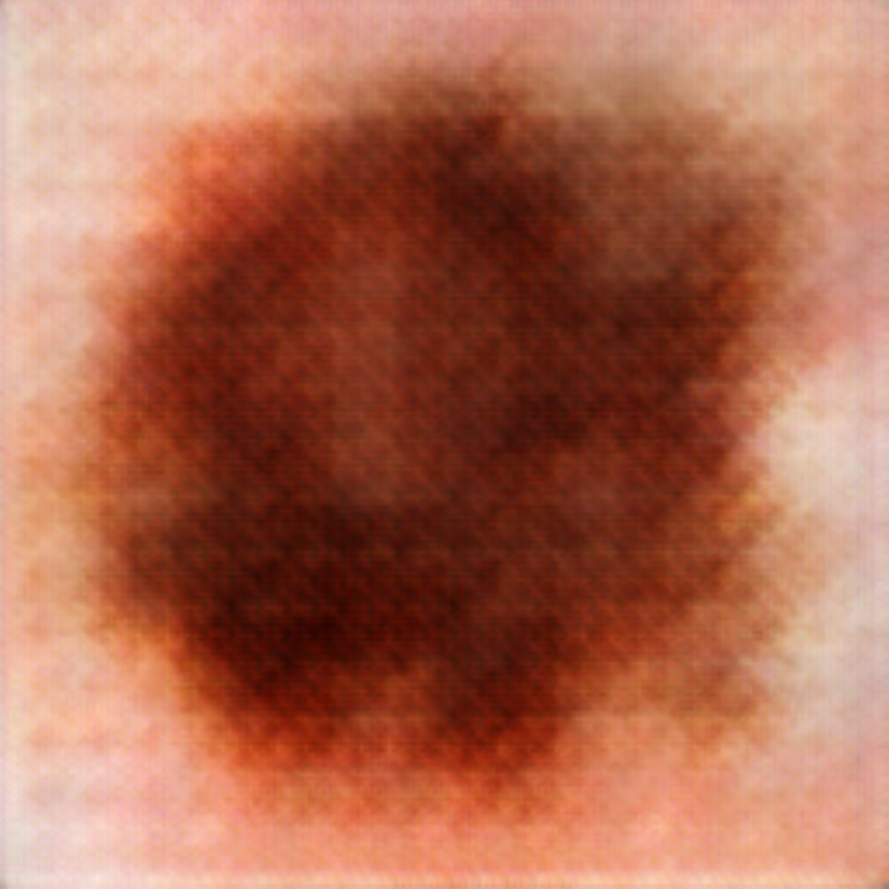}
    \includegraphics[width=0.19\textwidth]{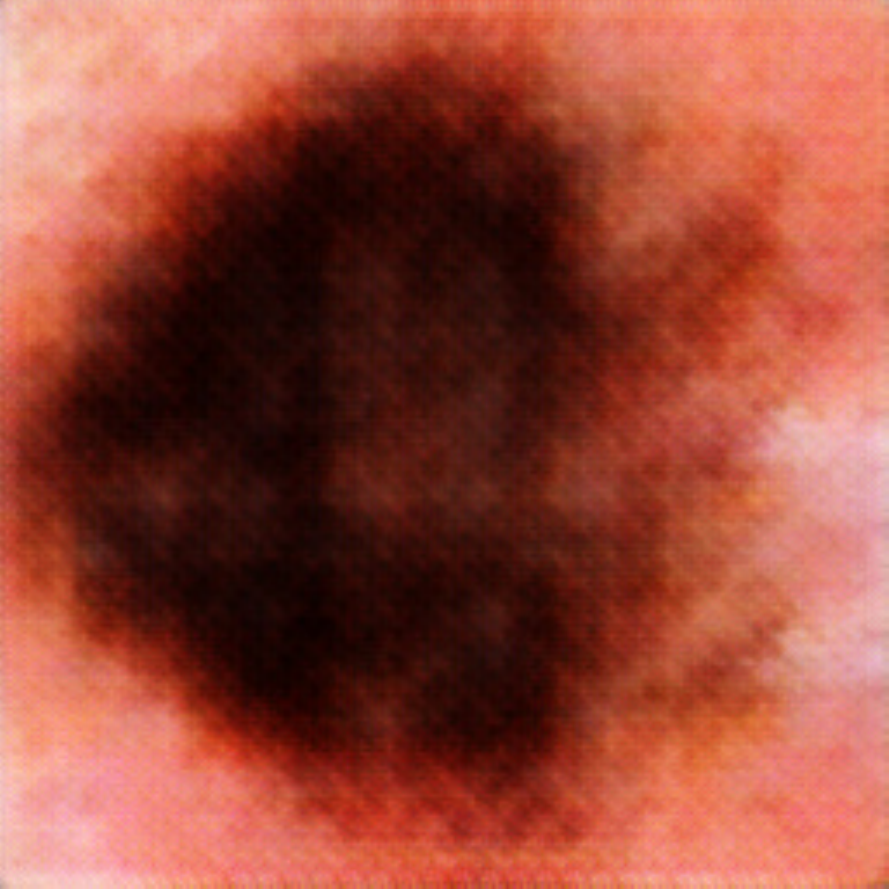}
    \includegraphics[width=0.19\textwidth]{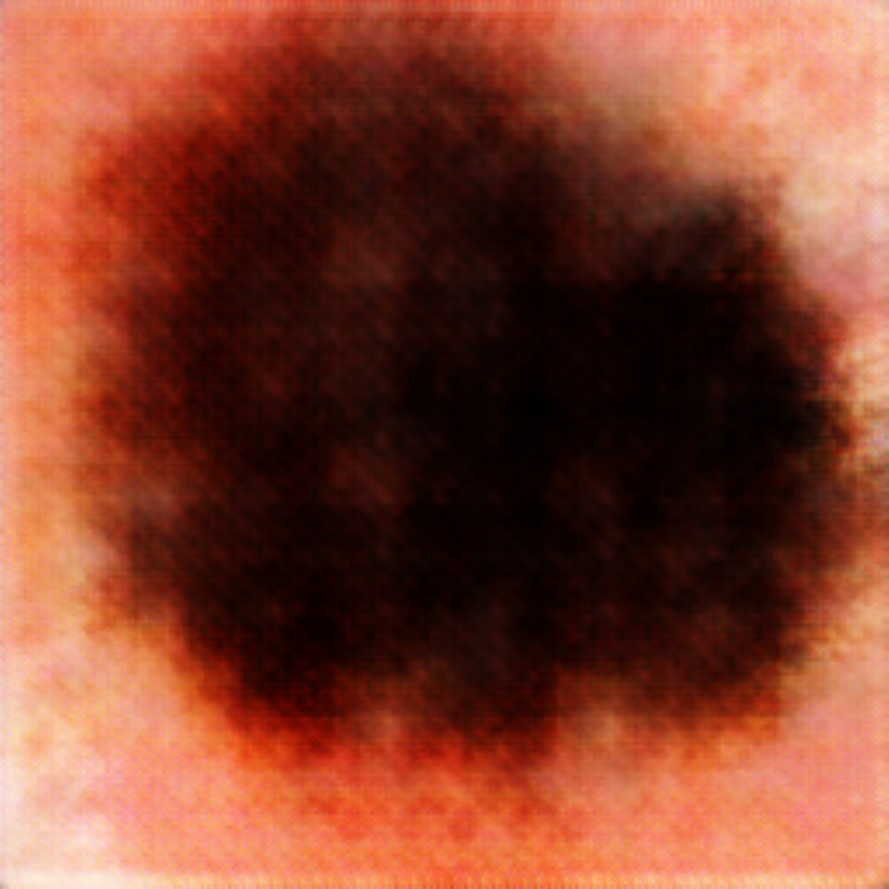}
    
    \stackunder[5pt]{\includegraphics[width=0.19\textwidth]{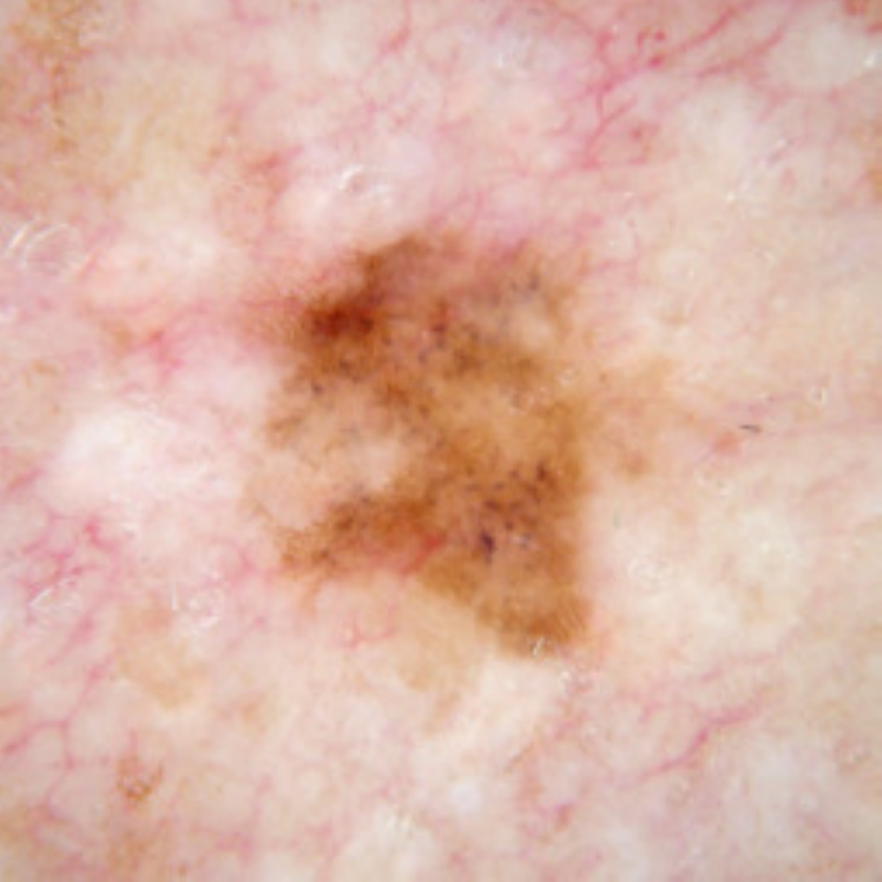}}{0.02}
    \stackunder[5pt]{\includegraphics[width=0.19\textwidth]{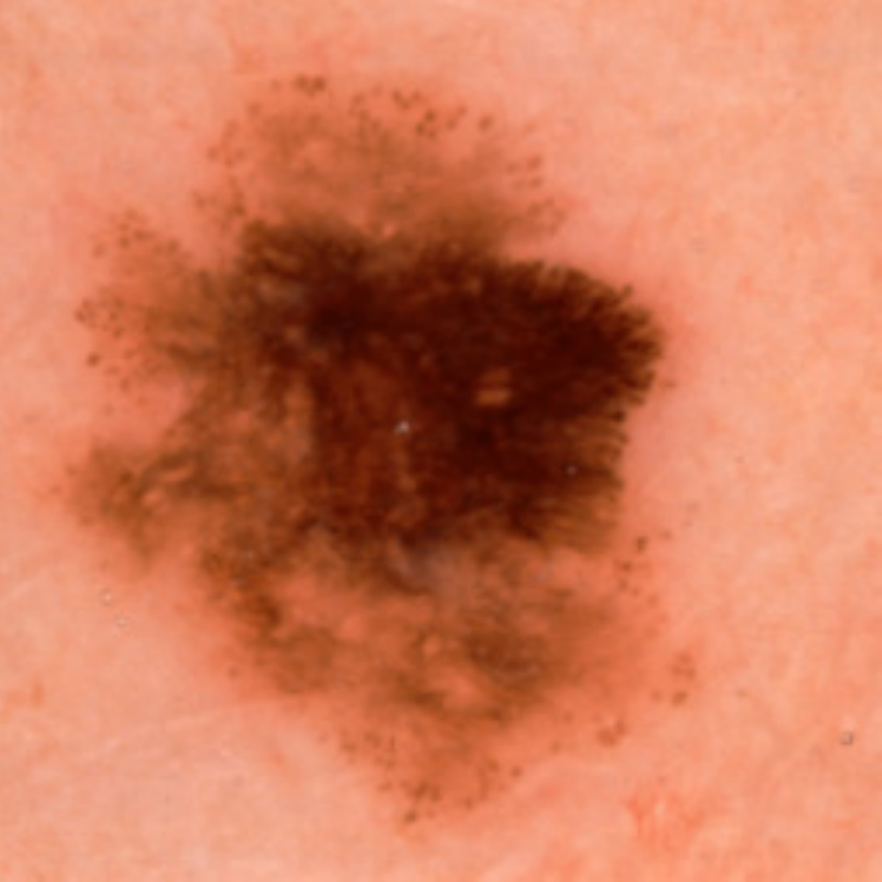}}{0.09}
    \stackunder[5pt]{\includegraphics[width=0.19\textwidth]{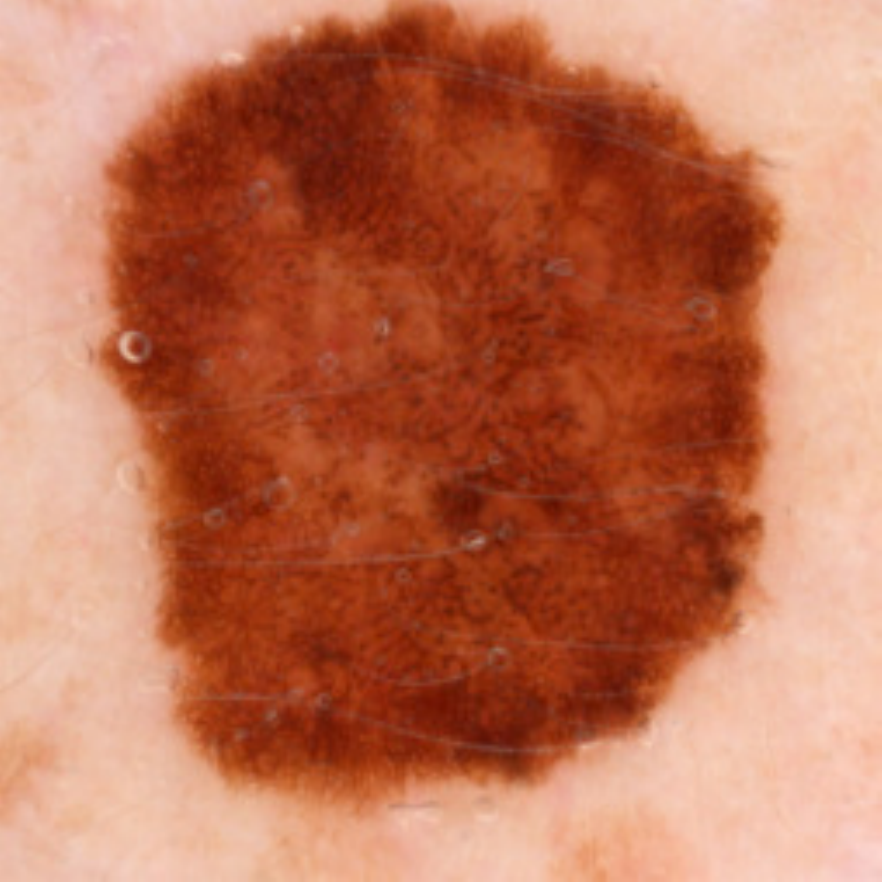}}{0.18}
    \stackunder[5pt]{\includegraphics[width=0.19\textwidth]{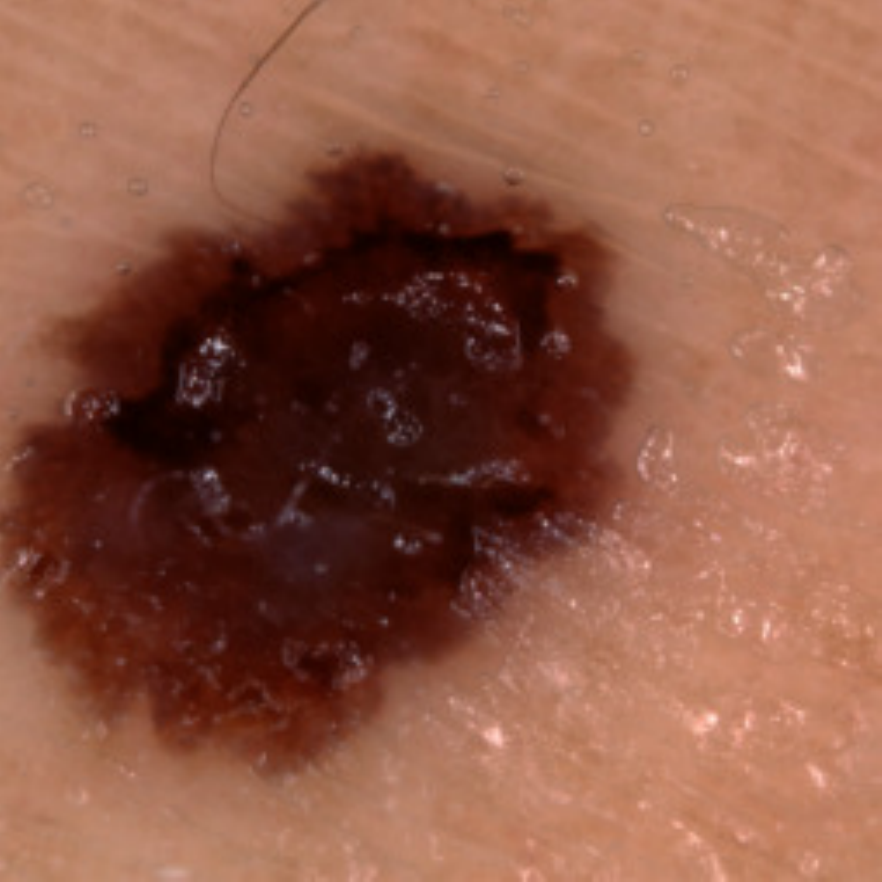}}{0.25}
    \stackunder[5pt]{\includegraphics[width=0.19\textwidth]{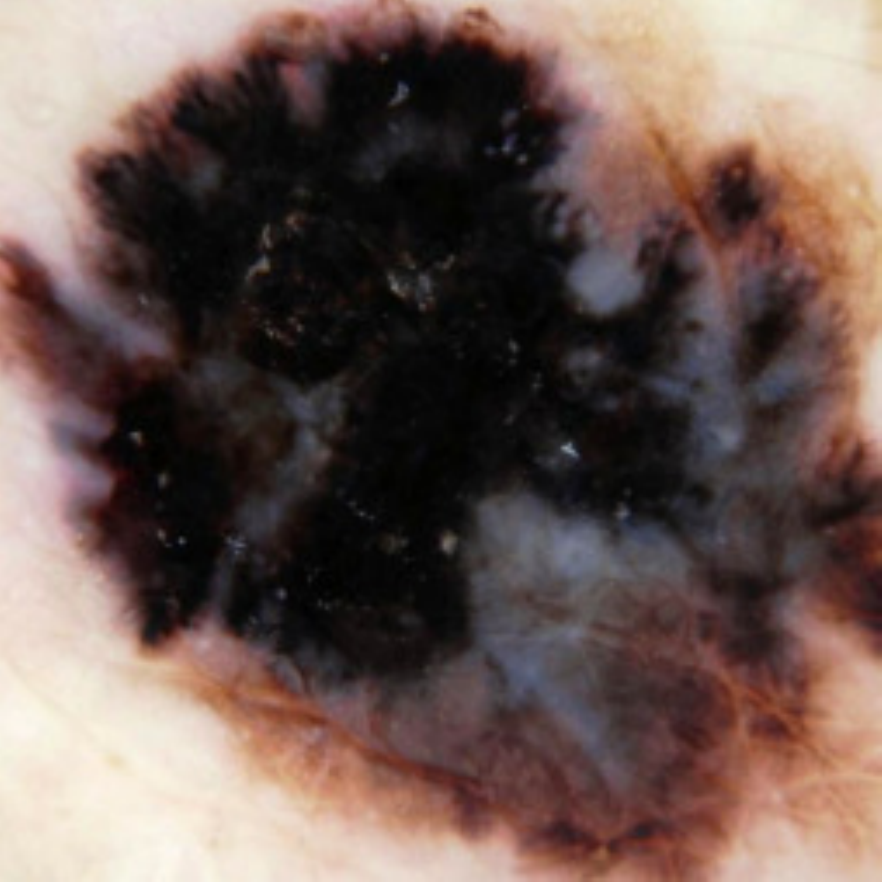}}{0.31}
    \end{figure}
    \end{minipage}\\
    (B)\begin{minipage}{0.45\textwidth}
    \begin{figure}[H]
    \centering
    \includegraphics[width=0.19\textwidth]{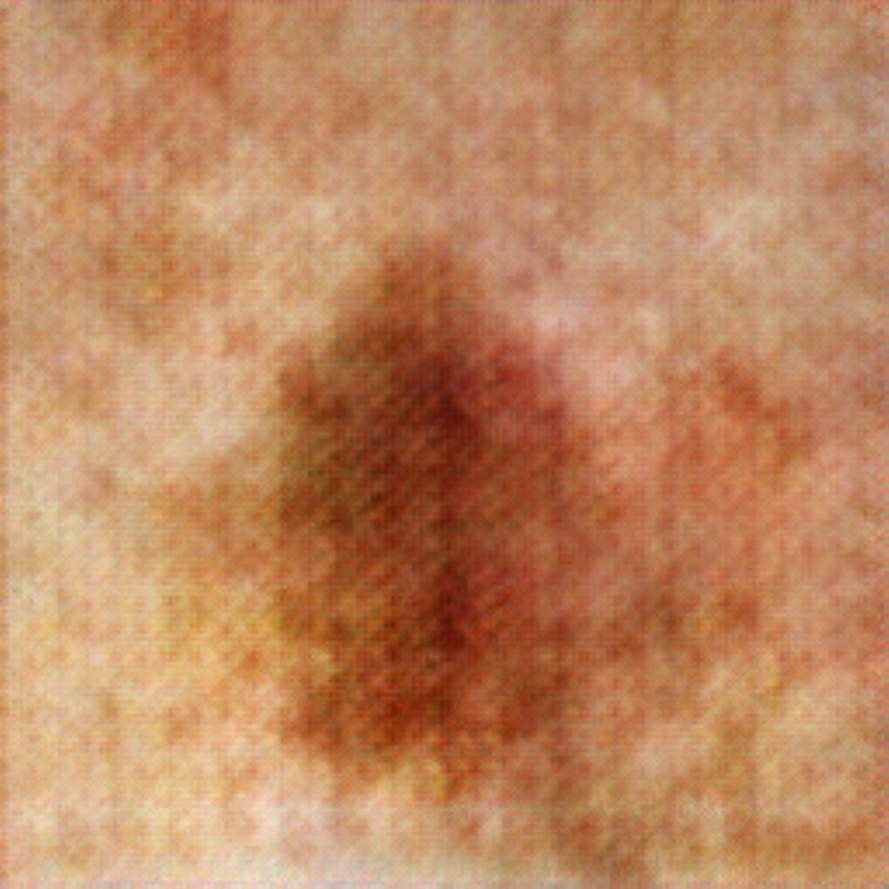}
    \includegraphics[width=0.19\textwidth]{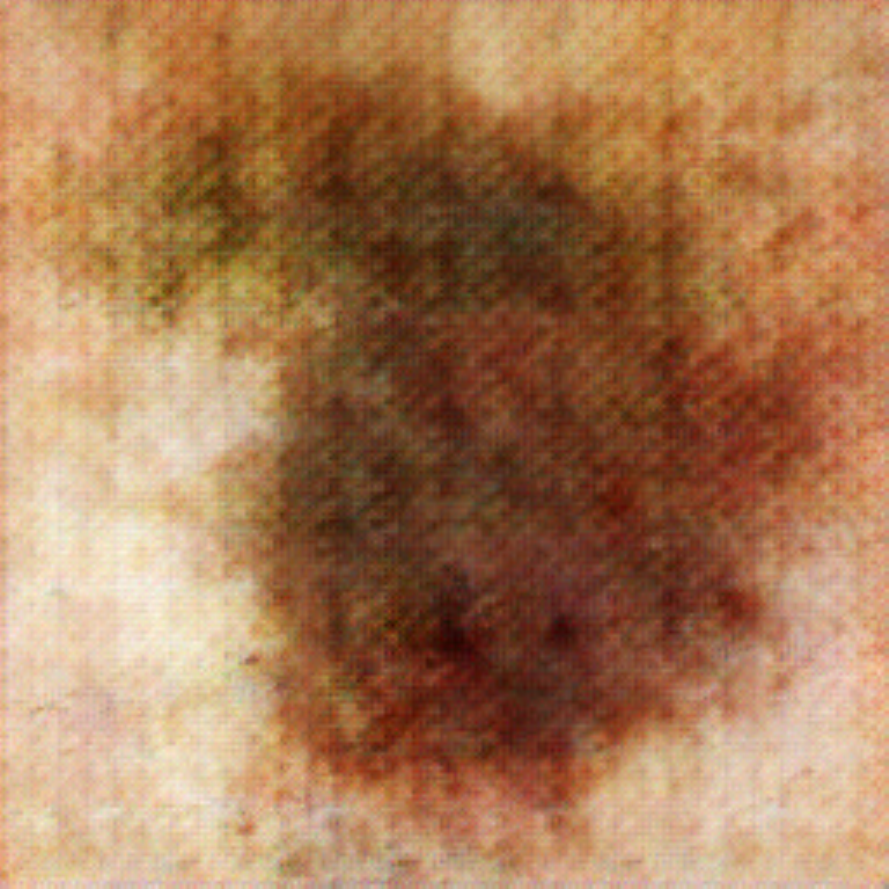}
    \includegraphics[width=0.19\textwidth]{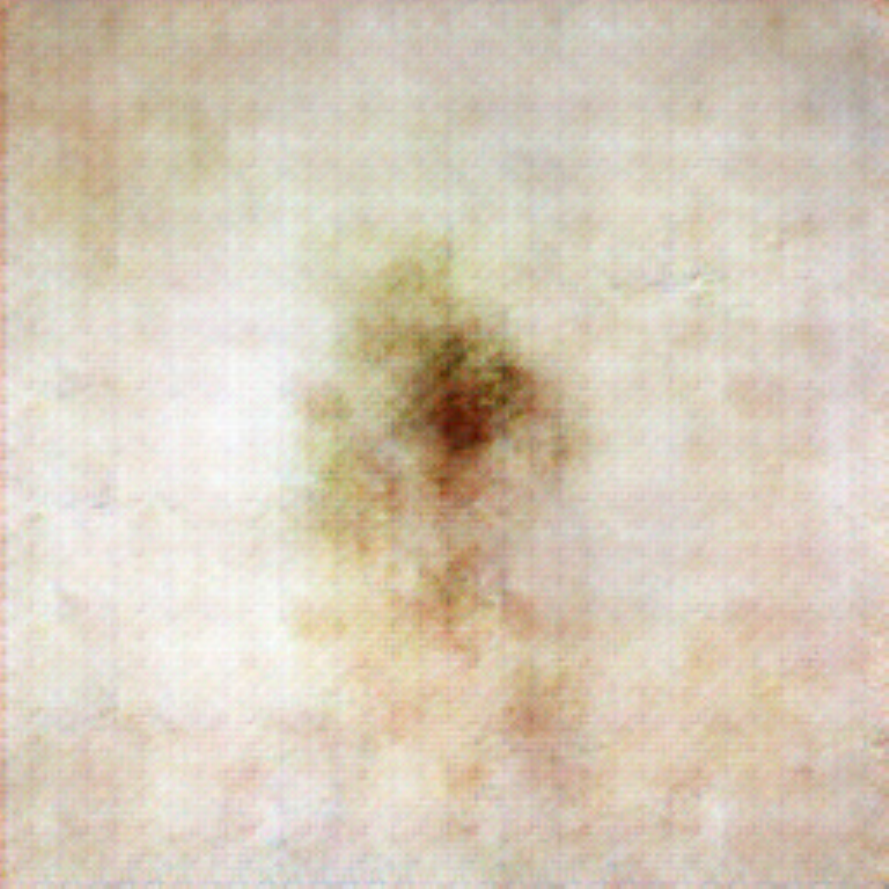}
    \includegraphics[width=0.19\textwidth]{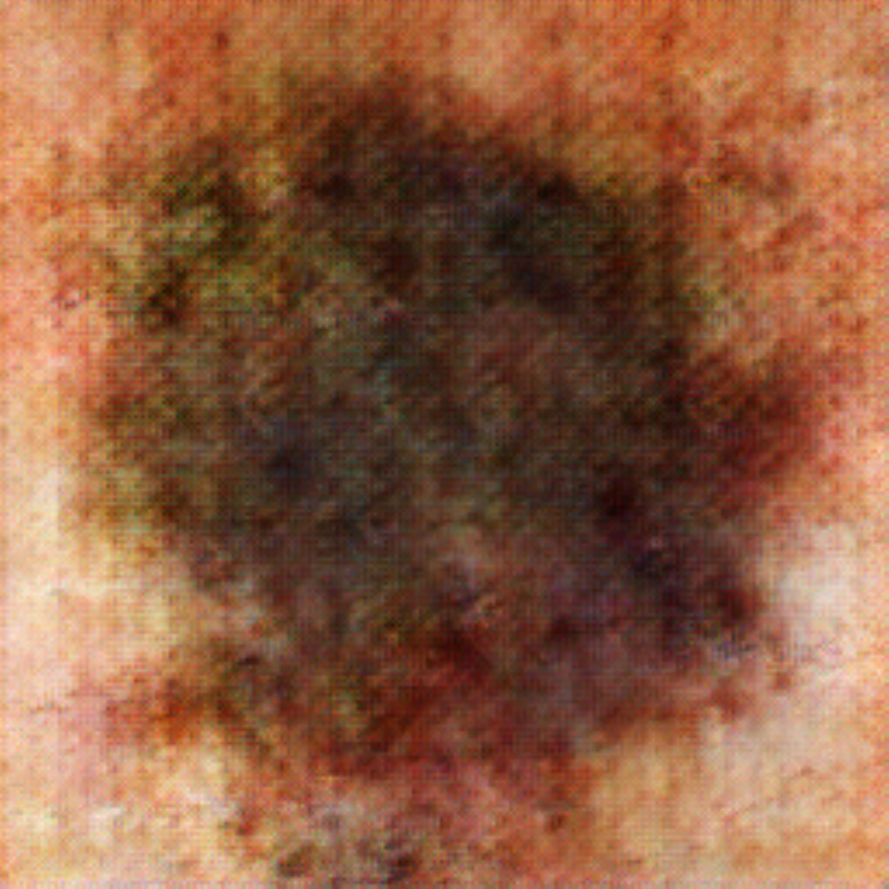}
    \includegraphics[width=0.19\textwidth]{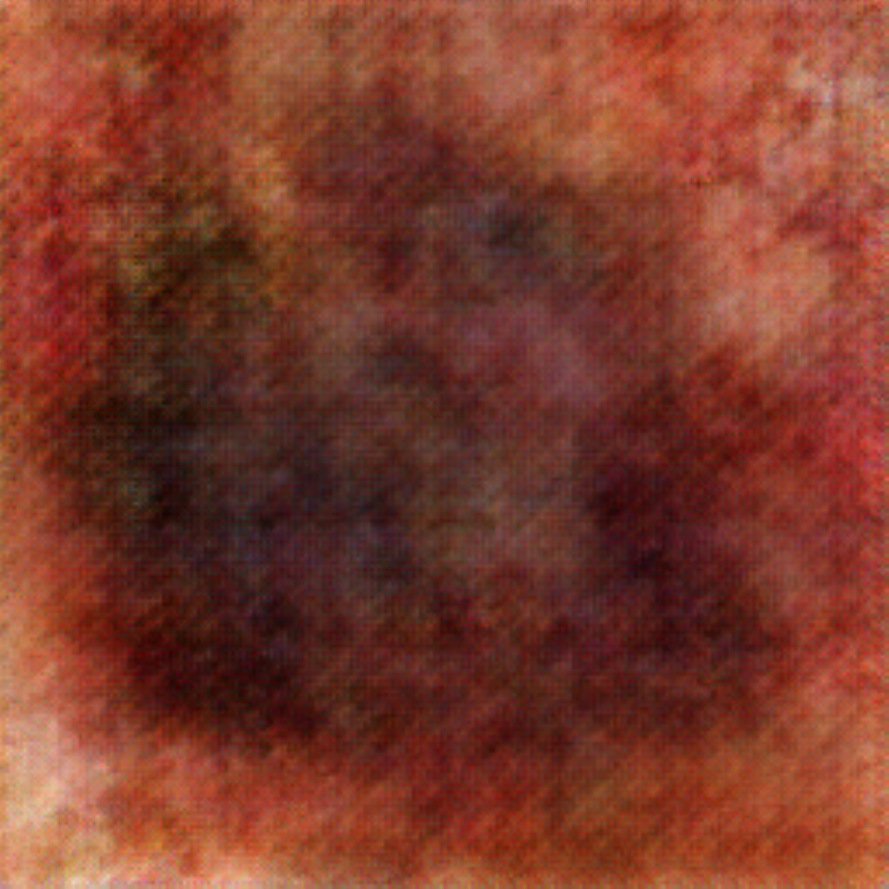}
    \stackunder[5pt]{\includegraphics[width=0.19\textwidth]{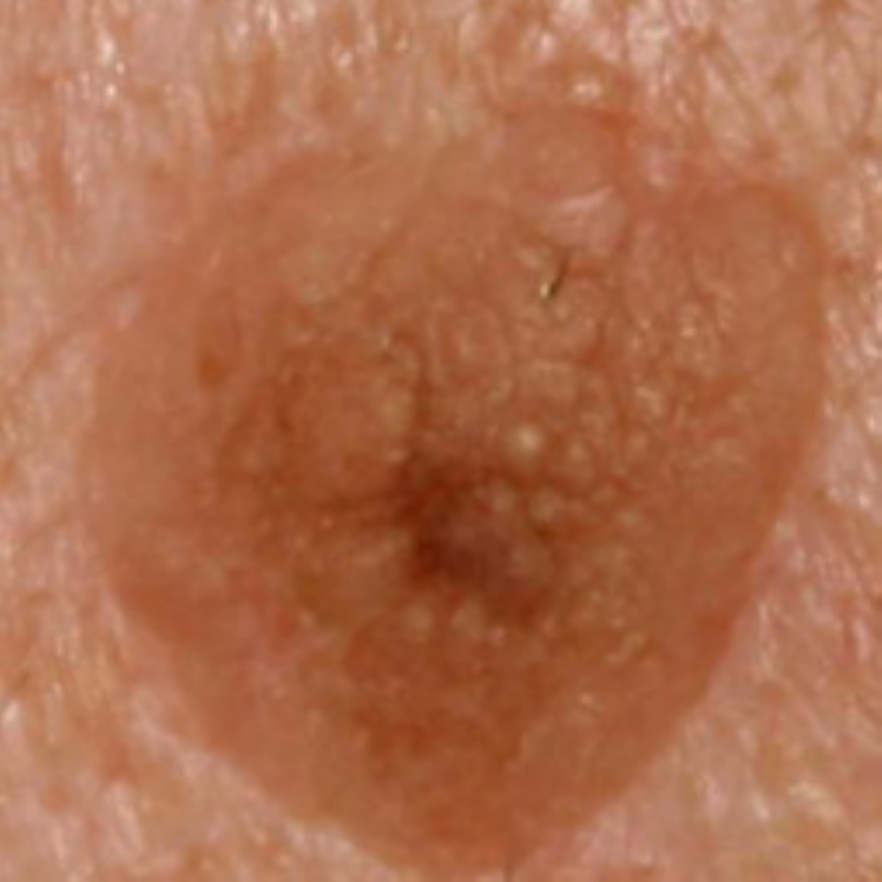}}{0.02}
    \stackunder[5pt]{\includegraphics[width=0.19\textwidth]{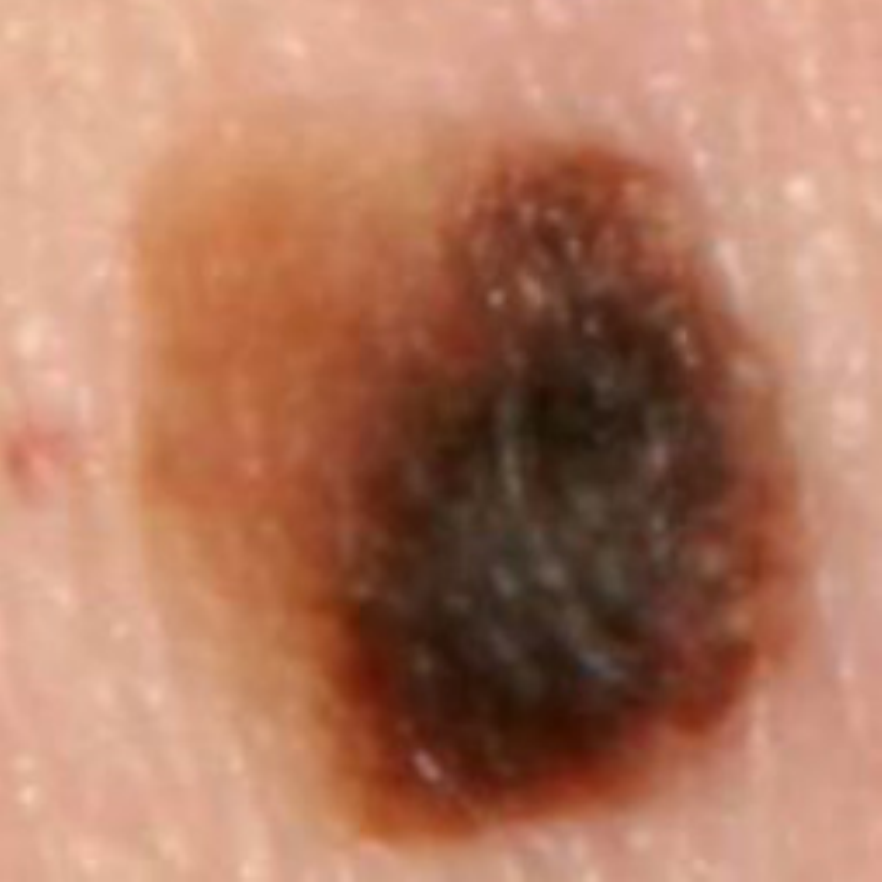}}{0.04}
    \stackunder[5pt]{\includegraphics[width=0.19\textwidth]{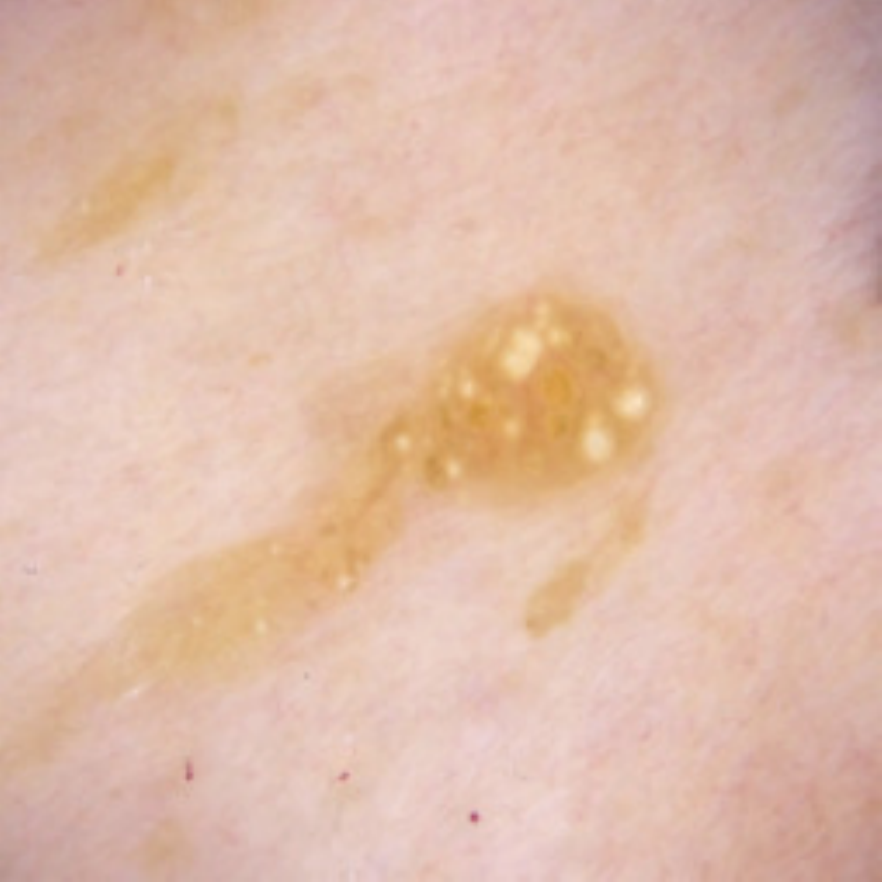}}{0.059}
    \stackunder[5pt]{\includegraphics[width=0.19\textwidth]{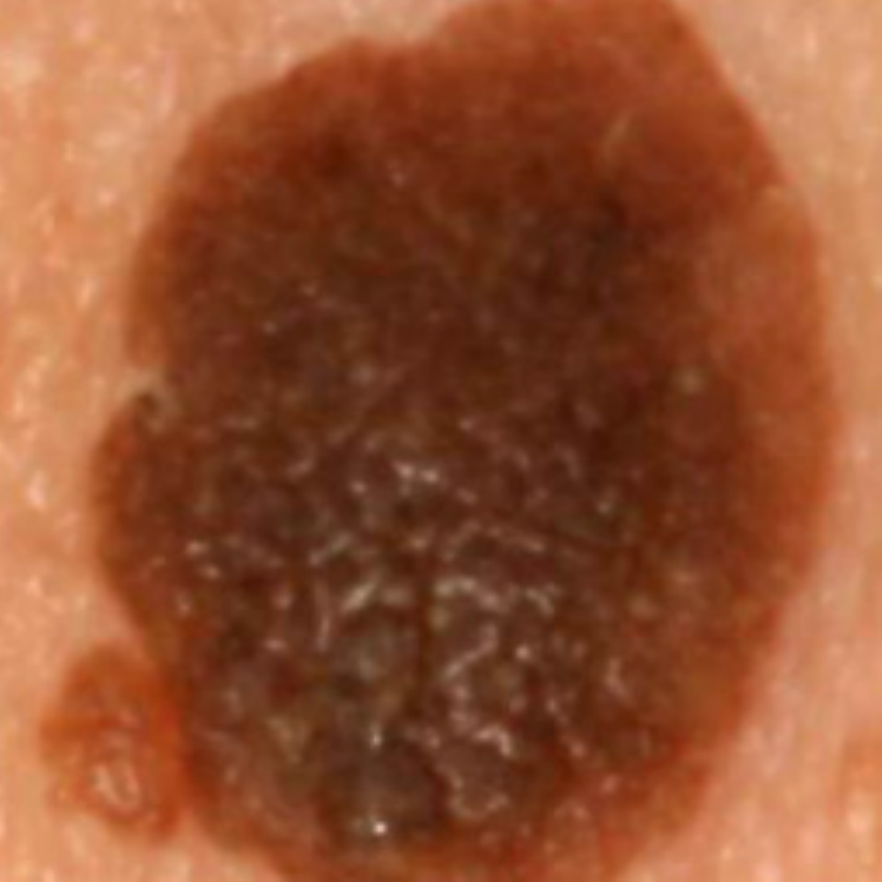}}{0.07}
    \stackunder[5pt]{\includegraphics[width=0.19\textwidth]{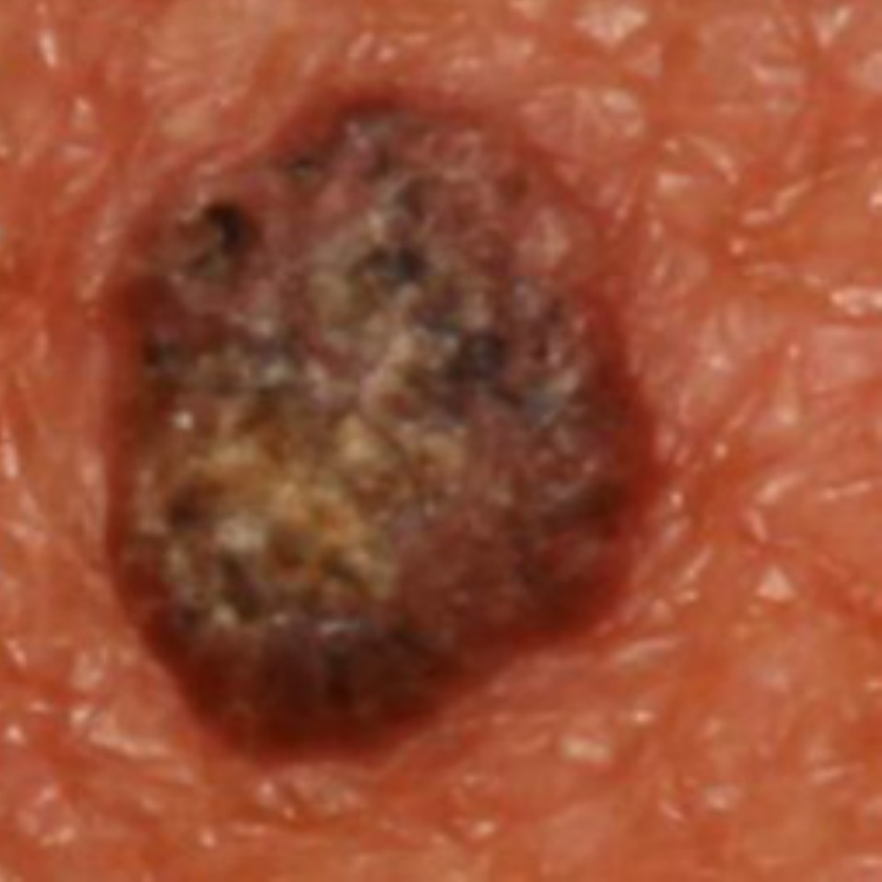}}{0.11}
    \end{figure}
    \end{minipage}\\
    (C)\begin{minipage}{0.45\textwidth} 
    \begin{figure}[H]
        \includegraphics[width=0.19\textwidth]{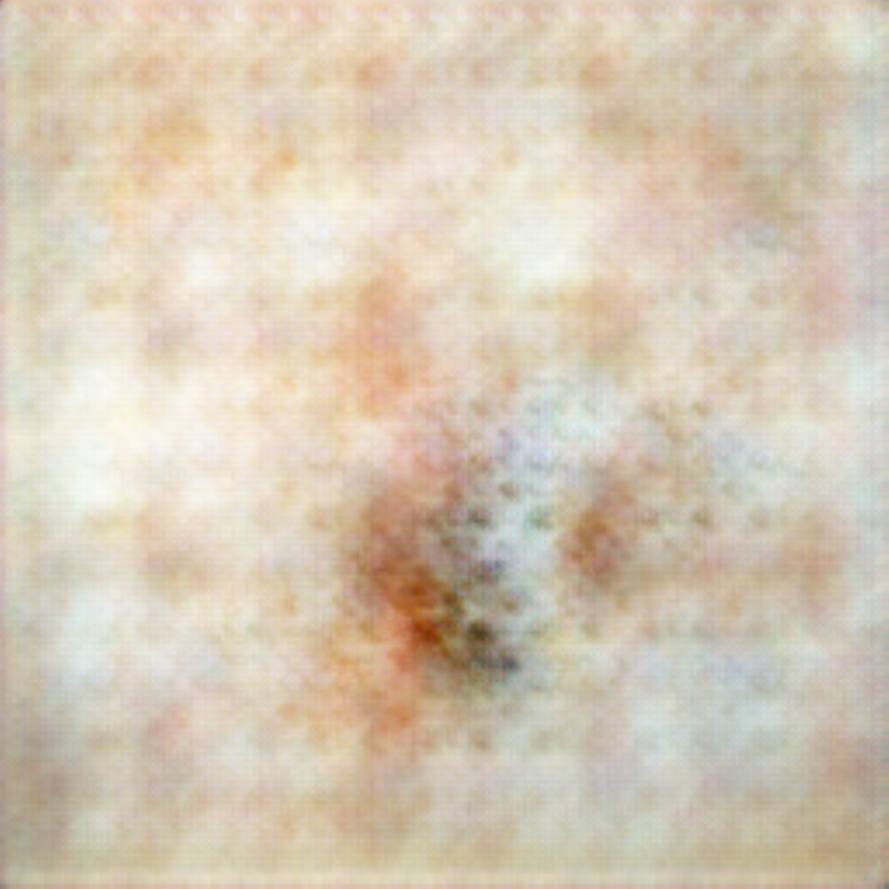}
        \includegraphics[width=0.19\textwidth]{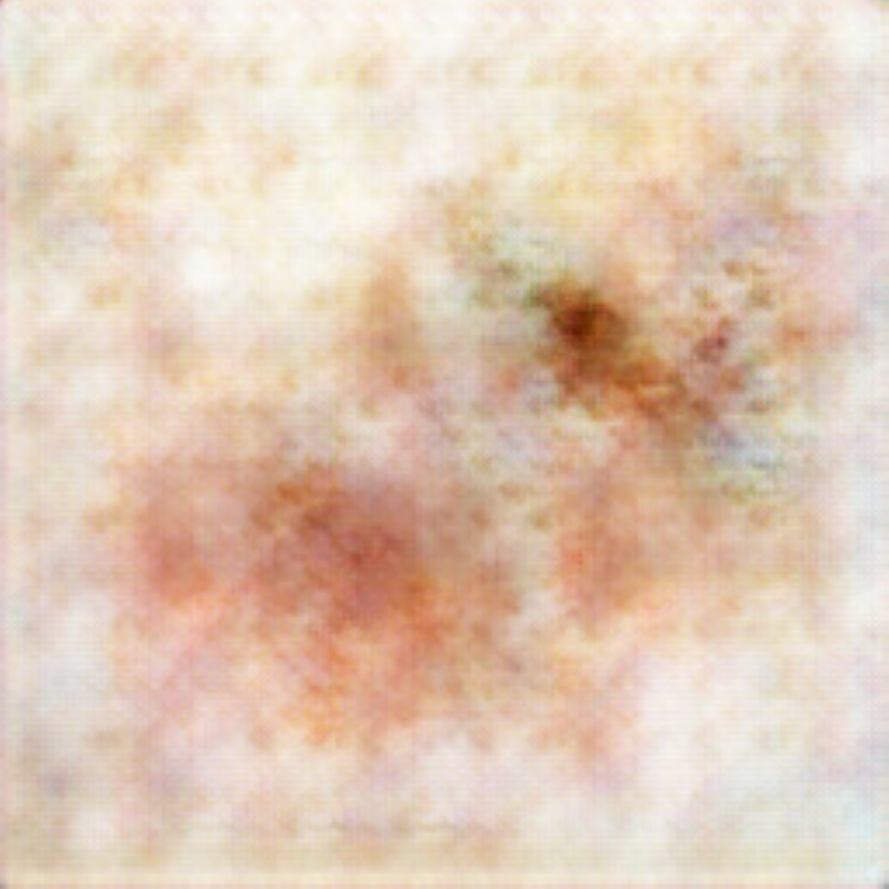}
        \includegraphics[width=0.19\textwidth]{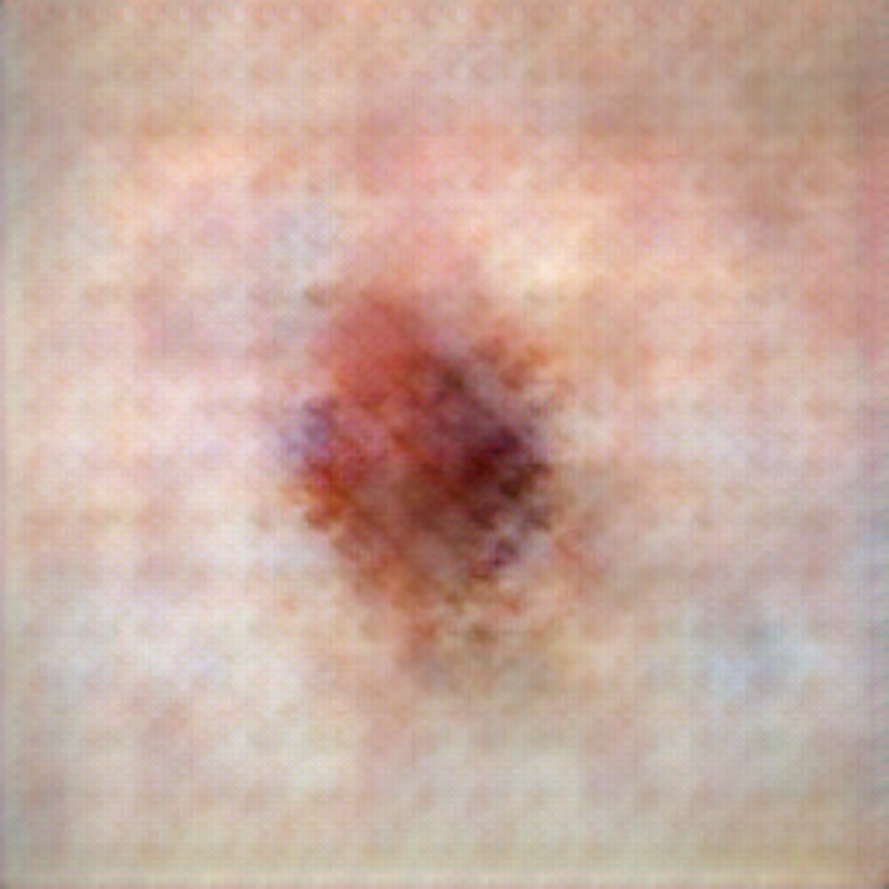}
        \includegraphics[width=0.19\textwidth]{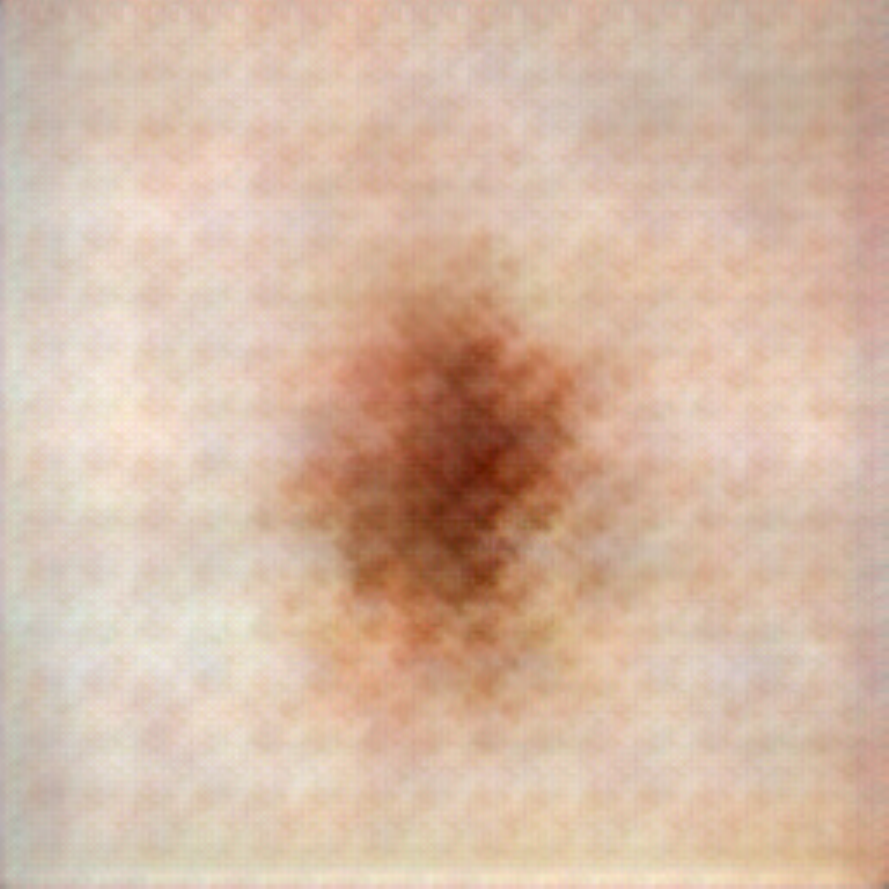}
        \includegraphics[width=0.19\textwidth]{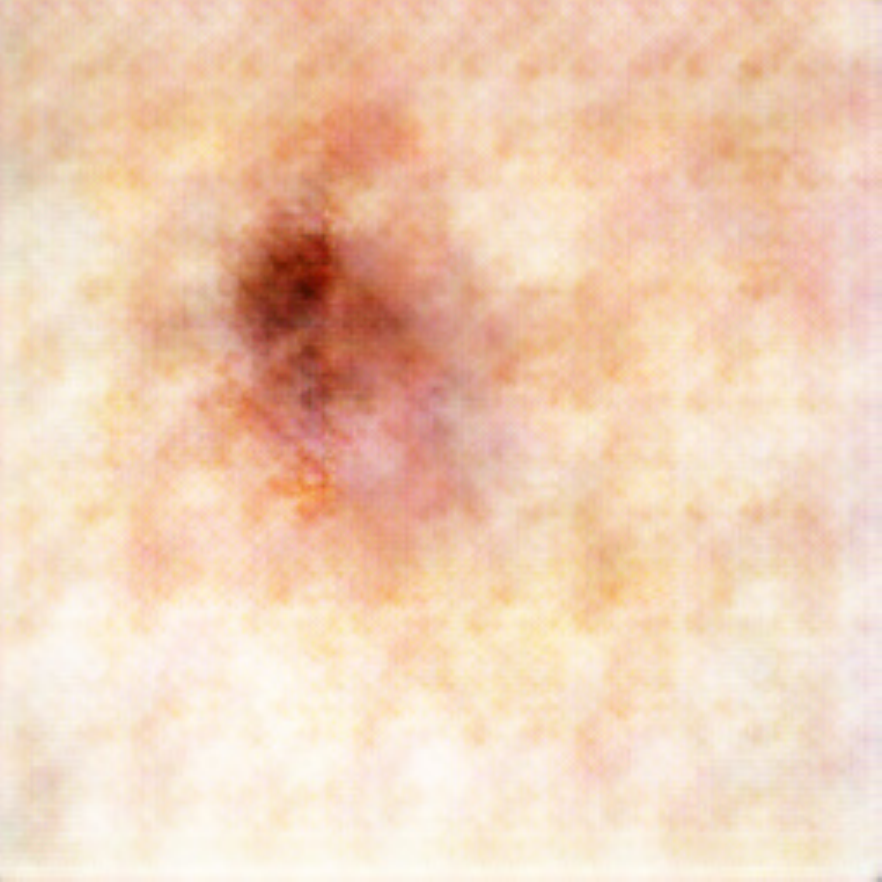}
        
        \includegraphics[width=0.19\textwidth]{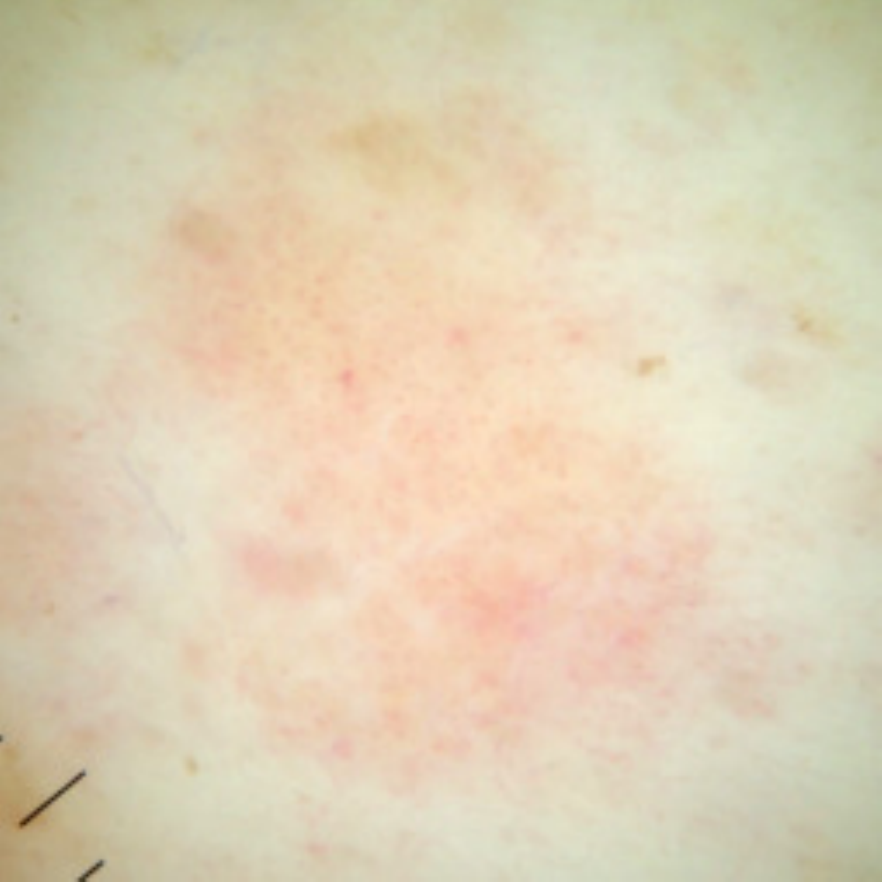}
        \includegraphics[width=0.19\textwidth]{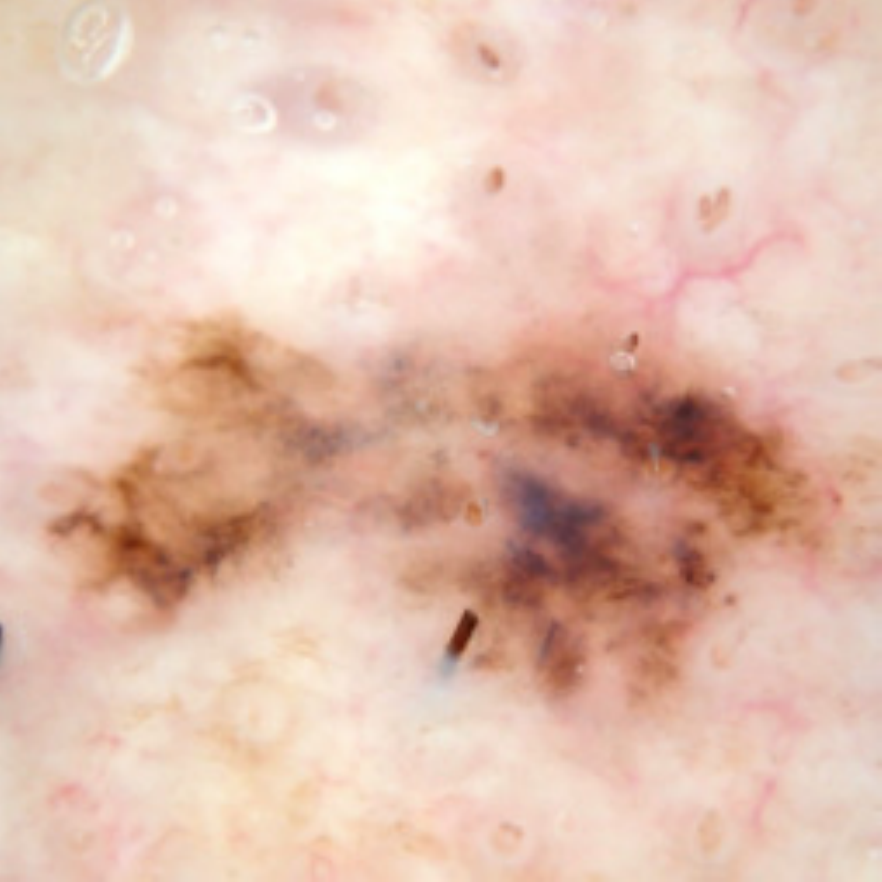}
        \includegraphics[width=0.19\textwidth]{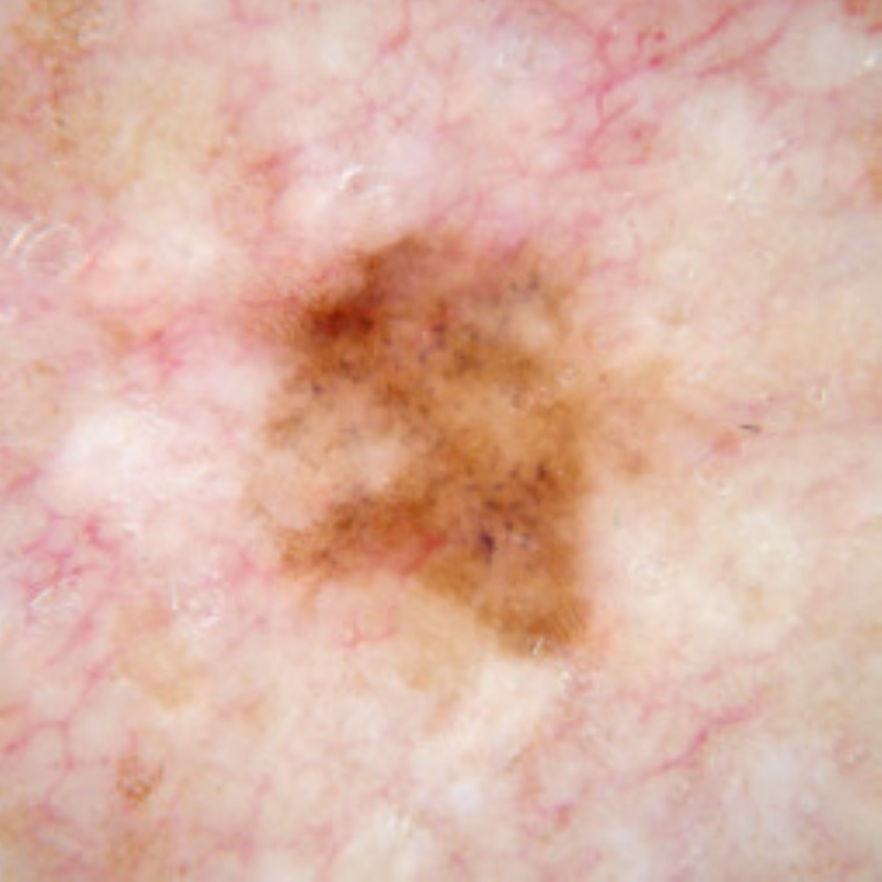}
        \includegraphics[width=0.19\textwidth]{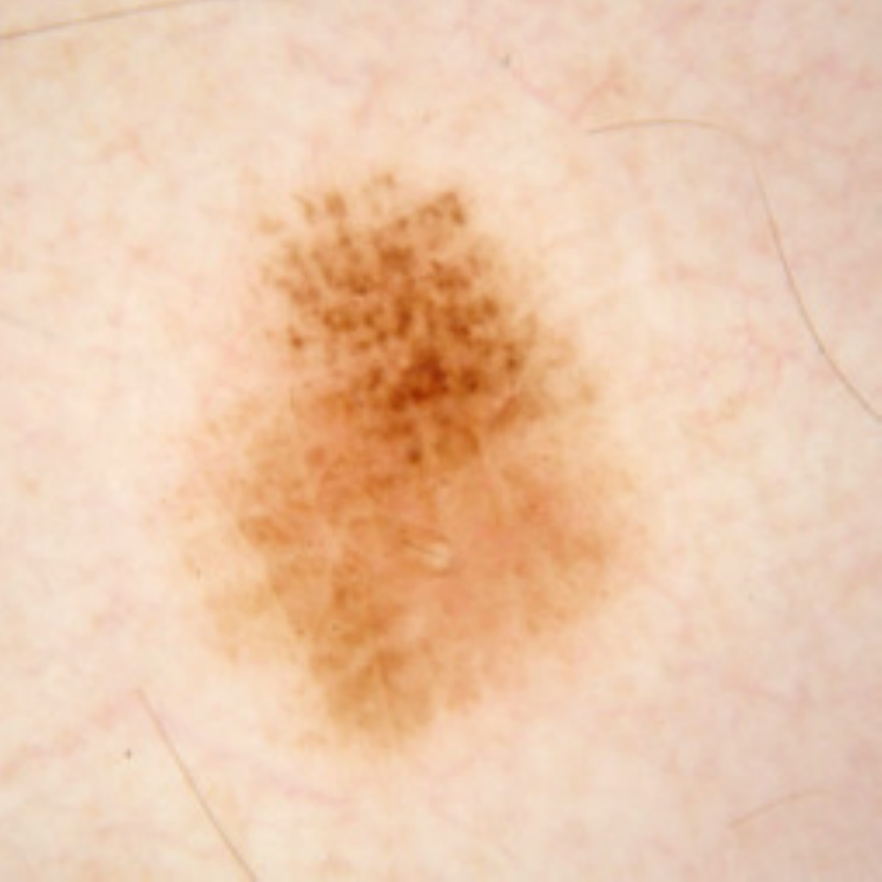}
        \includegraphics[trim={4pt 4pt 4pt 4pt},clip,height=0.19\textwidth, width=0.19\textwidth]{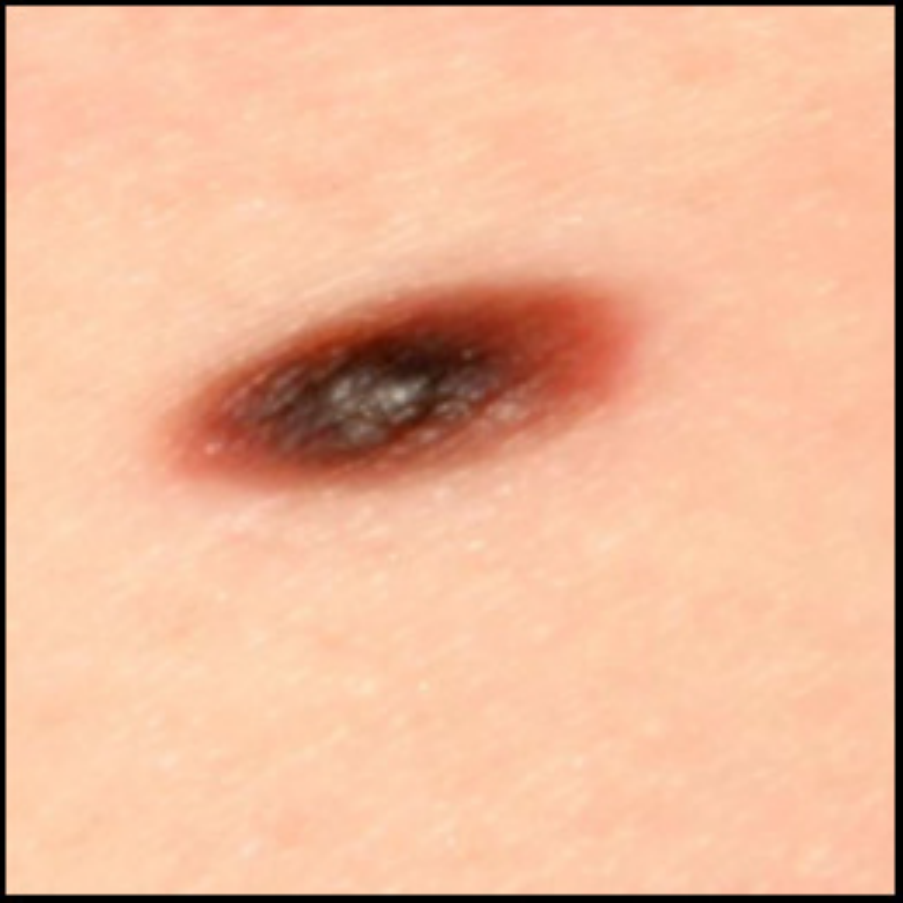}
    \end{figure}
    \end{minipage}\\
    (D)\begin{minipage}{0.45\textwidth}
     \begin{figure}[H]
        \includegraphics[width=0.19\textwidth]{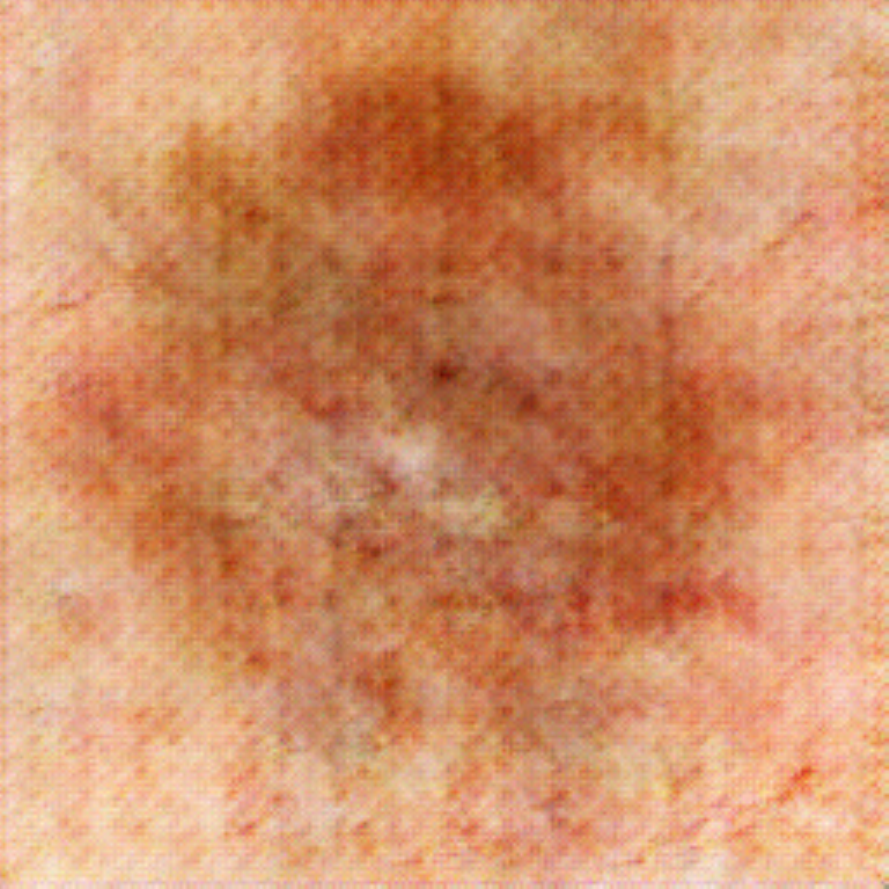}
        \includegraphics[width=0.19\textwidth]{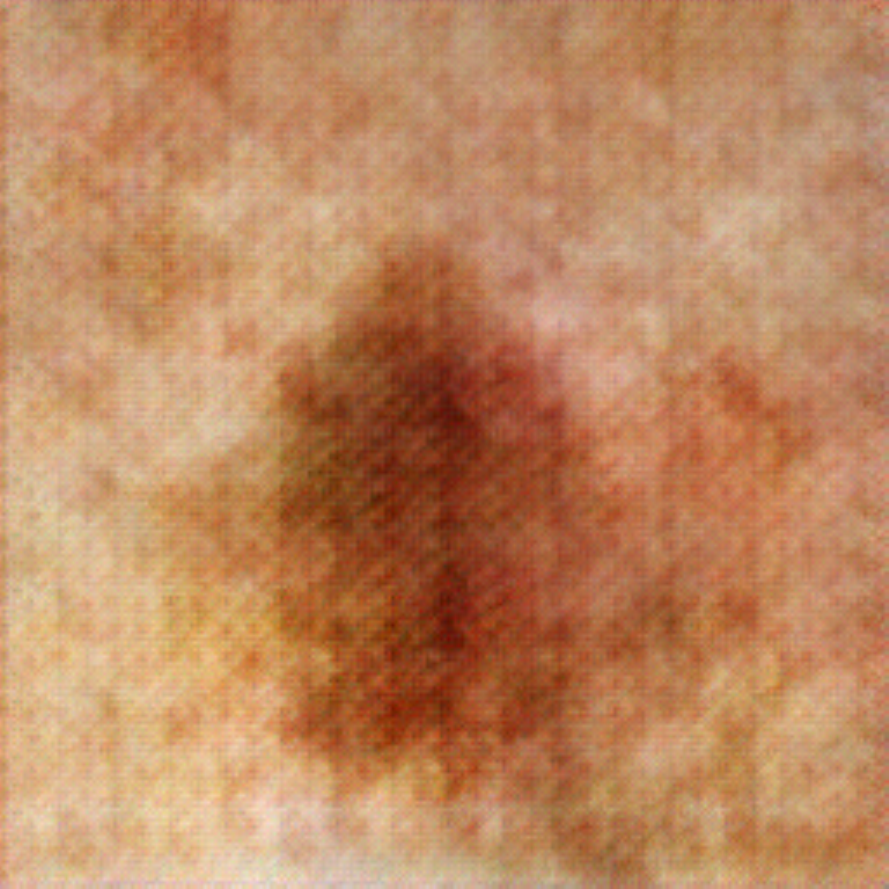}
        \includegraphics[width=0.19\textwidth]{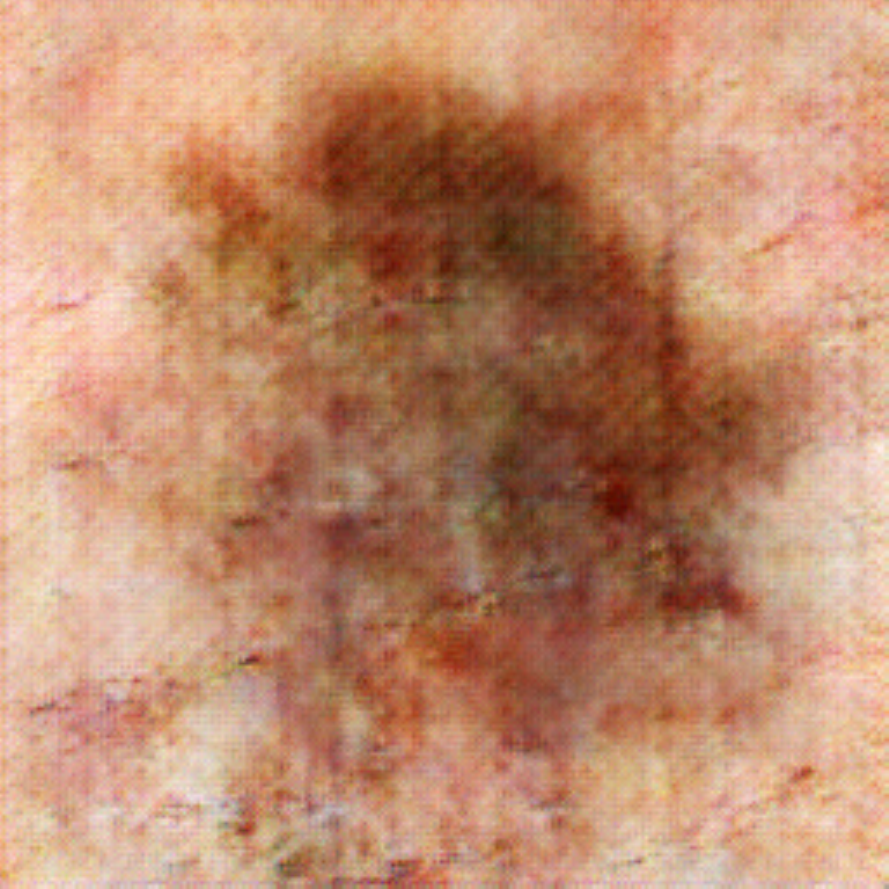}
        \includegraphics[width=0.19\textwidth]{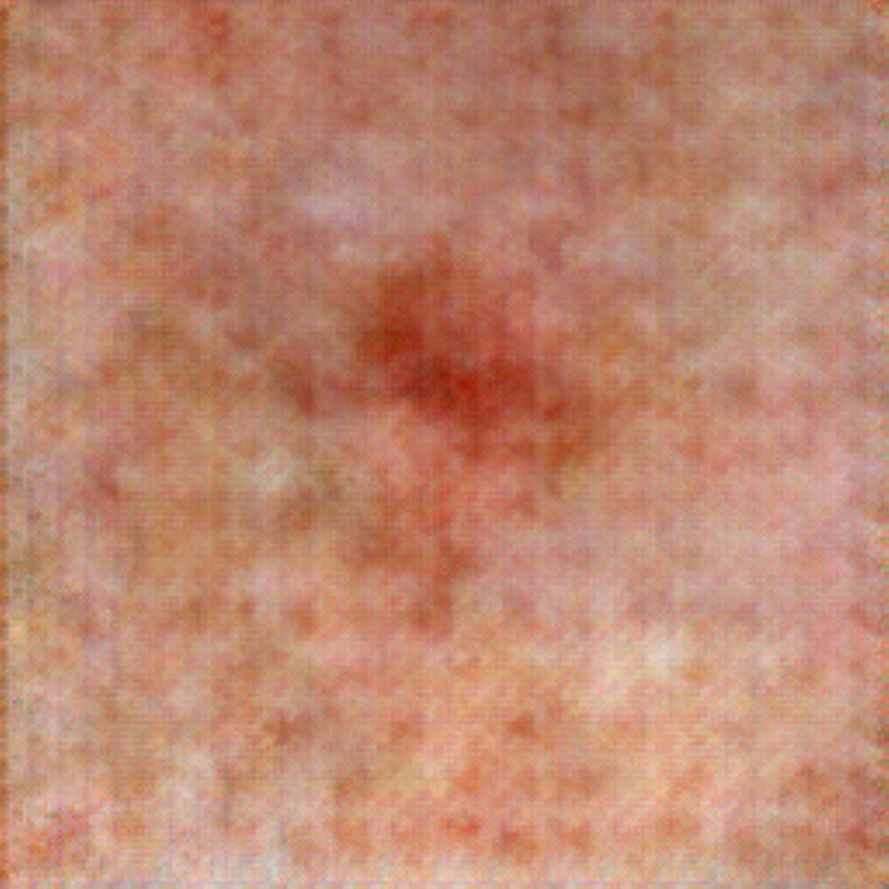}
        \includegraphics[width=0.19\textwidth]{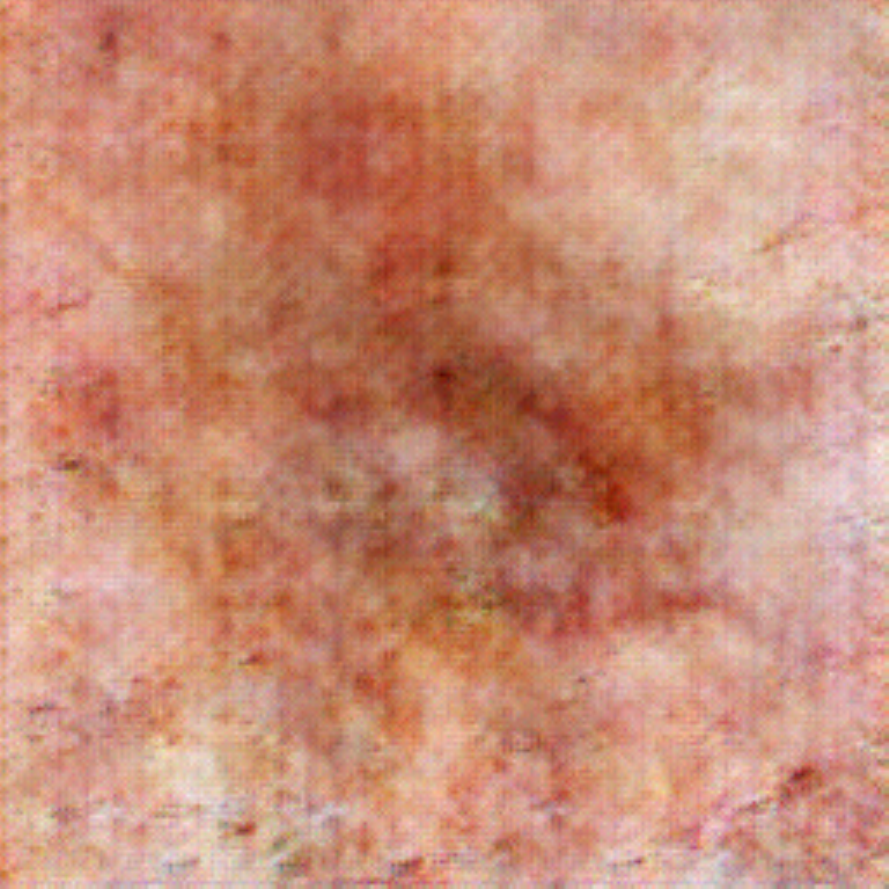}
        
        \includegraphics[width=0.19\textwidth]{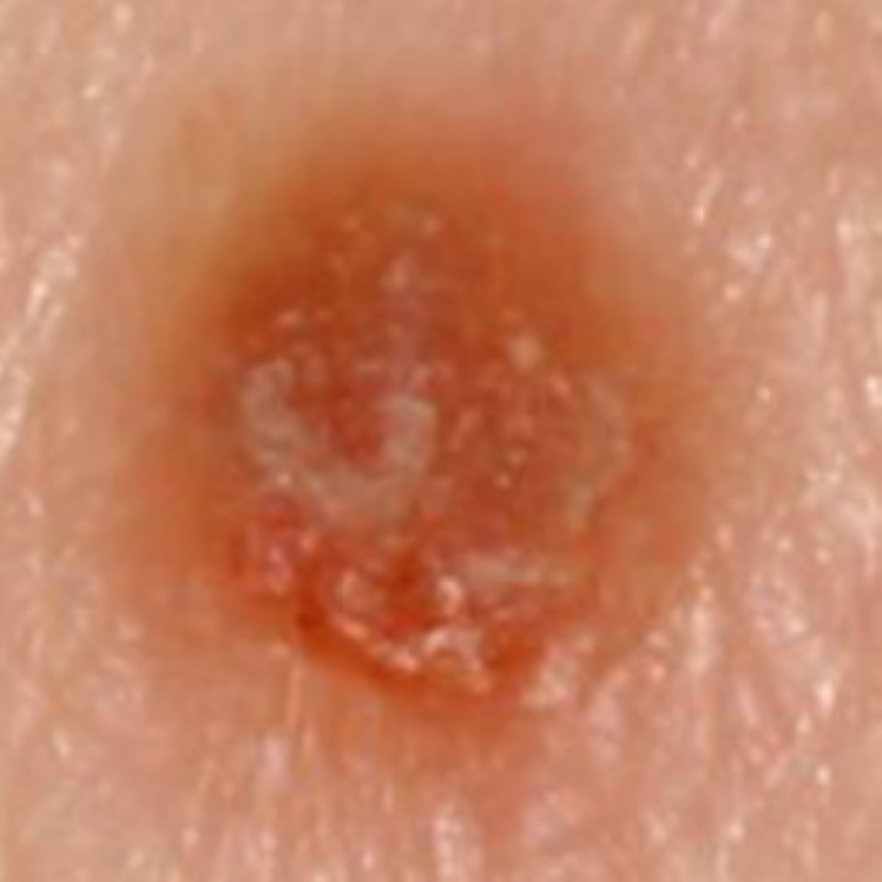}
        \includegraphics[width=0.19\textwidth]{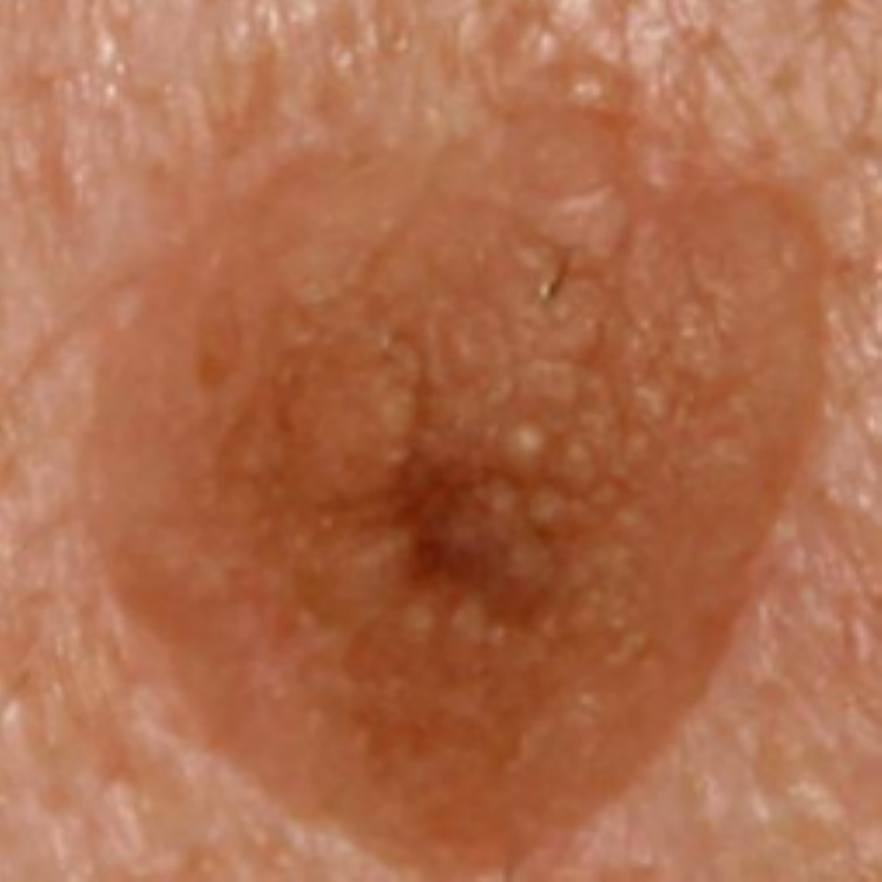}
        \includegraphics[width=0.19\textwidth]{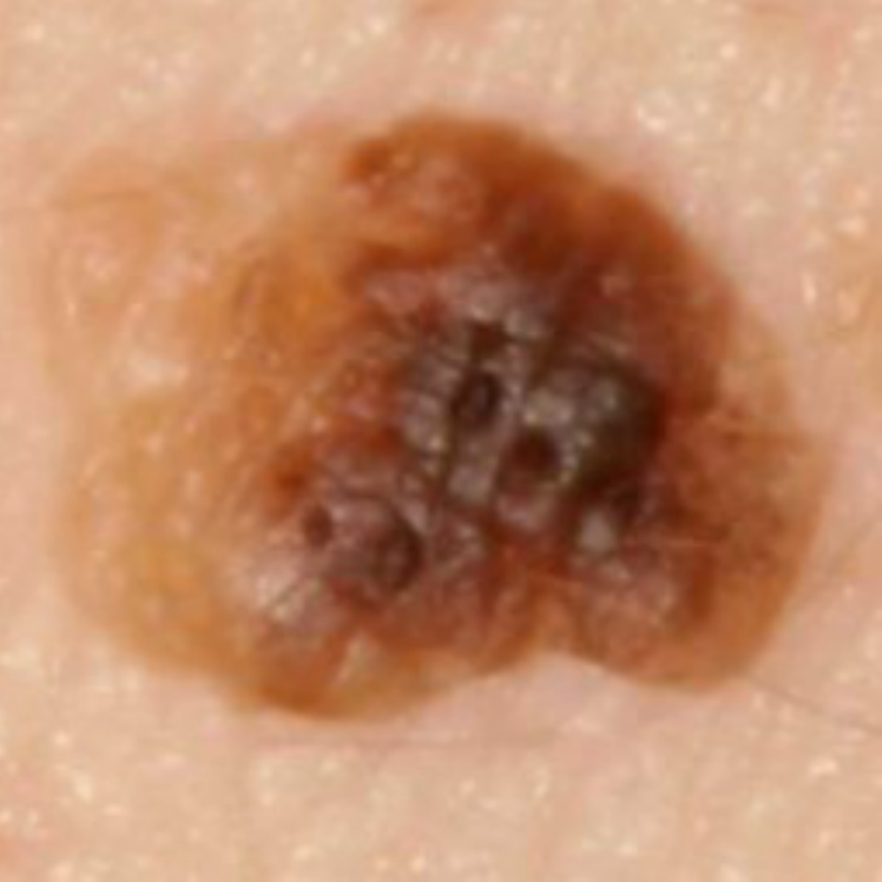}
        \includegraphics[width=0.19\textwidth]{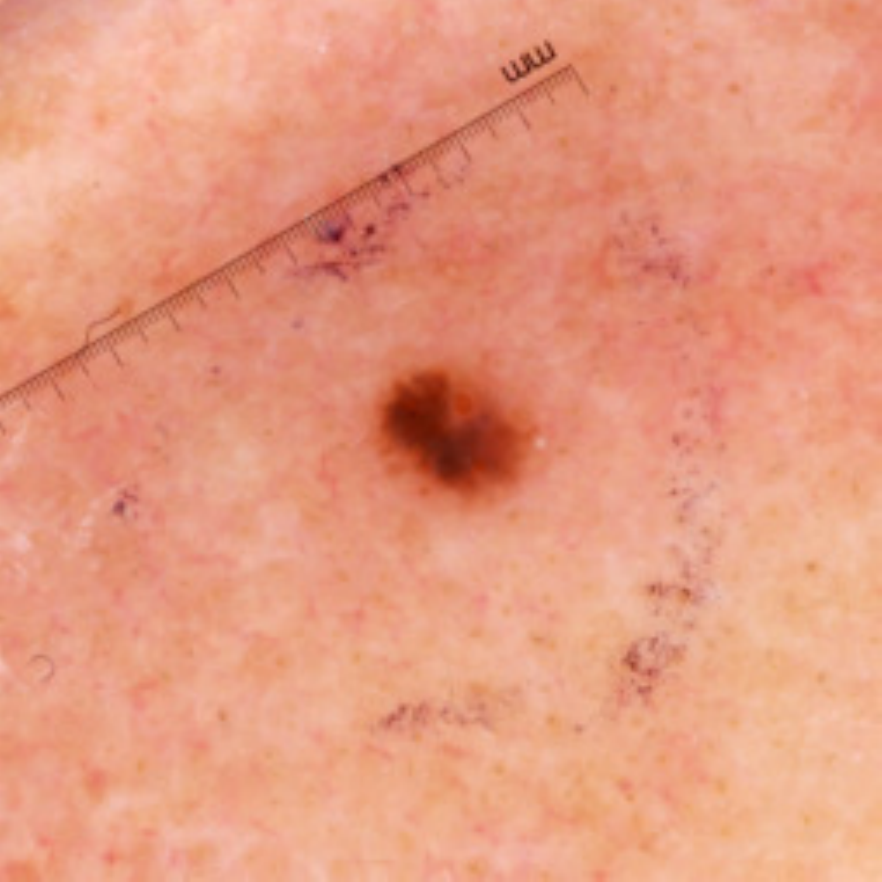}
        \includegraphics[width=0.19\textwidth]{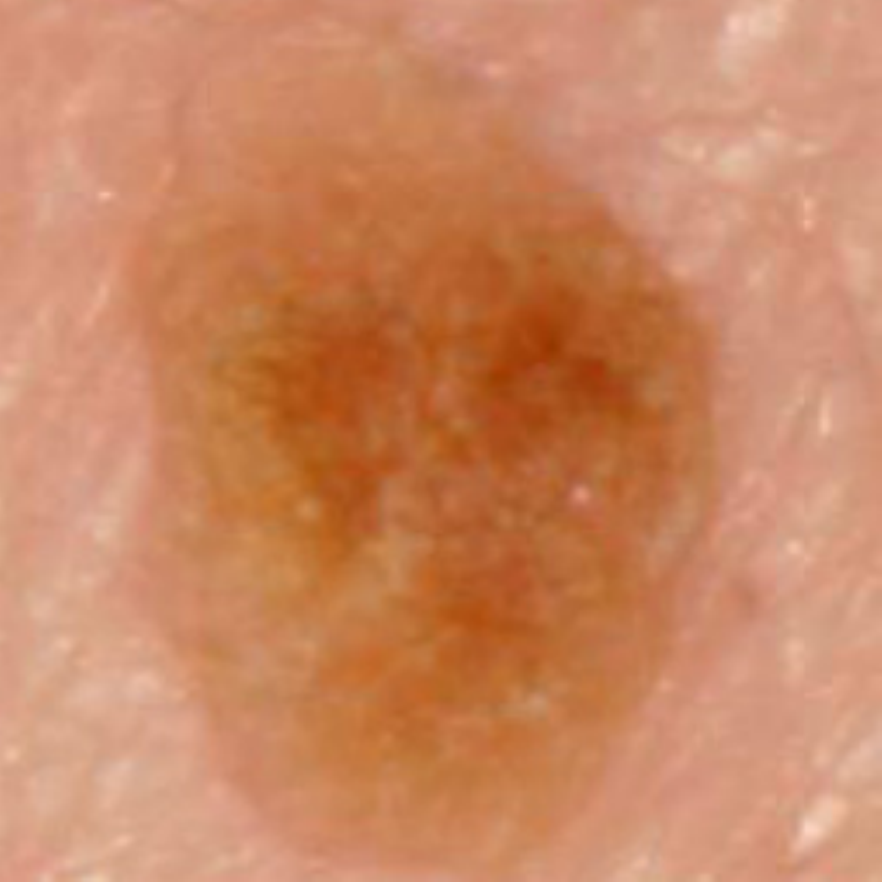}\\
    \end{figure}
    \end{minipage}
    \caption{ \textbf{(A)} Generated melanoma images \textbf{(top)} and the original images from the training set \textbf{(bottom)} for different values of MSE. \textbf{(B)} generated seborrheic keratosis images \textbf{(top)} and the original images from the training set \textbf{(bottom)} for different values of MSE. \textbf{(C)} generated melanoma images \textbf{(top)} and the original images from the training data set \textbf{(bottom)} for the smallest obtained values of MSE. \textbf{(D)} generated seborrheic keratosis images \textbf{(top)} and the original images from the training data set \textbf{(bottom)} for the smallest obtained values of MSE.}
\label{fig:examp}
\end{figure}

\begin{figure}[t]
        \centering
        \includegraphics[width=0.45\textwidth]{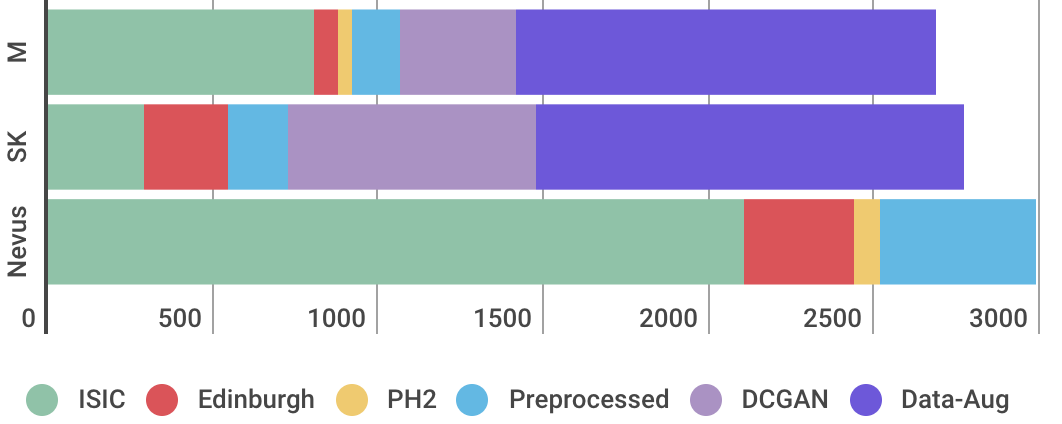}
        \caption{Illustration of the process of gradually increasing the size of the training data for the classification model through data pre-processing (hairs and rulers removal), data generation with DCGAN, and data augmentation (Data-Aug) using standard techniques.}
        \label{fig:il}
 \end{figure}

\begin{table*}[h]
    \centering
    \renewcommand{\arraystretch}{1.2} 
    \begin{tabular}{|c|p{5em}|p{6em}|l|l|l|l|}
        \hline
        & Layer & Output Size & Kernel & Stride  & Padding \\
        \hline
        \multirow{7}{7em}{Generator} 
        & TransConv & 256$\times$4$\times$4 & 4$\times$4  & 1 & 0  \\
        & TransConv & 128$\times$8$\times$8 & 4$\times$4 & 2 & 1   \\
        & TransConv & 64$\times$16$\times$16 & 4$\times$4 & 2 & 1  \\
        & TransConv & 32$\times$32$\times$32 & 4$\times$4 & 2 & 1  \\
        & TransConv & 16$\times$64$\times$64 & 4$\times$4 & 2 & 1  \\
        & TransConv & 8$\times$128$\times$128 & 4$\times$4 & 2 & 1 \\
        & TransConv & 3$\times$256$\times$256 & 4$\times$4 & 2 & 1 \\
        \hline
        \multirow{7}{7em}{Discriminator}
        & Conv & 16$\times$128$\times$128& 4$\times$4  & 2 & 1\\
        & Conv & 32$\times$64$\times$64 & 4$\times$4 & 2 & 1  \\
        & Conv & 64$\times$32$\times$32 & 4$\times$4 &2 & 1   \\
        & Conv & 128$\times$16$\times$16 & 4$\times$4 & 2 & 1 \\
        & Conv & 256$\times$8$\times$8 & 4$\times$4 & 2 & 1   \\
        & Conv & 512$\times$4$\times$4 & 4$\times$4 & 1 & 1   \\
        & Conv & 1$\times$1$\times$1 & 4$\times$4 & 1 & 0     \\
        \hline
    \end{tabular}
    \caption{Details of the architecture of the generative model. Let $n$ be the number of features maps, $h$ be the height and $w$ be the width. Size of the output feature maps is represented as $n\times h \times w$. Each convolution layer in the generator, except for the last one, is followed by a batch normalization and a ReLU nonlinearity. The last convolution layer is followed by a hyperbolic tangent. Similarly, each layer in the discriminator, except for the last convolution layer, is followed by a batch normalization and a leaky ReLU nonlinearity with the leakage coefficient of 0.2. The last convolution layer is followed by a sigmoid.}
    \label{tab:DCGAN}
\end{table*}
We propose the data generation method utilizing de-coupled DCGANs. We use two separate Deep Convolutional Generative Adversarial Networks (DCGANs) \cite{DCGAN} to generate $350$ images of melanoma and $750$ images of seborrheic keratosis, which were the two classes heavily under-represented in the ISIC 2017 data set compared to a much larger nevus class. Since we use separate networks for each class, we refer to this approach as ``de-coupled DCGANs''. ISIC-2018 dataset contains large number of classes thus, we couple the DCGANs \cite{DBLP:journals/corr/0001T16} making the initial layers share parameters. This technique is highlighted in the supplementary section of the paper. We extended the architecture of DCGAN to produce images of resolution $256 \times 256$. The model architecture is highlighted in the Figure~\ref{fig:histo} and Table \ref{tab:DCGAN}. GAN techniques rely on training a generator network to generate images which have similar distribution to the one followed by the training data. The discriminator provides a feedback how close the two distributions are.  In our experiments, the latent vector of length 10 that inputs the generator is obtained from standard Gaussian distribution with mean 0 and standard deviation 1. We modify DCGAN to enable the generation of images with the desired resolution by adding layers to both generator and discriminator. Binary cross entropy loss and Adam optimizer with learning rate of $2e^{-4}$ and beta values of $0.5$ and $0.999$ were used to train both discriminator and generator. To prevent generator from collapsing and perform stable discriminator-generator optimization we utilize stabilization techniques \cite{train_gan} and perform early stopping while training the network. Furthermore, we perform two rounds of additional generator training after every $10^{\text{th}}$ round of joint training of both discriminator and generator. The process of data generation needs to be done carefully. It is essential to make sure that the generated images differ from the ones contained in the training data to maximize data variety. In order to verify that, we calculate the mean squared error (MSE) between each generated image to all the images from the training data set and choose the training image that corresponds to the minimal value of the MSE. We then compared each generated image with its closest, i.e. MSE-minimizing, training image to make sure they are not duplicates. Figure \ref{fig:histo} shows the histograms of the mean squared error (MSE) for seborrheic keratosis and melanoma. The histograms indicate the wide variation in the images generated by the model. Figure \ref{fig:examp} highlights some exemplary images from both melanoma and seborrheic keratosis classes for different values of MSE. It also presents additional images with smallest obtained MSE i.e images obtained from the extreme left side of both histograms. 
We finally augment the data in the two considered classes by performing horizontal flipping of images such that the class sizes increase to $2685$ for melanoma and $2772$ for seborrheic keratosis. The process of augmenting the training data by adding purified images, GAN-based data generation, and mentioned traditional augmentation is captured in Figure \ref{fig:il}. We finally augment the entire data set using vertical flipping and random cropping to increase the data set further $6$ times. The final training data set was obtained and that will be used for the classification model is balanced and contains $16110$ melanoma cases, $17928$ nevus cases, and $16632$ seborrheic keratosis cases, among which a notable fraction, i.e $26\%$, constitute the artificially-generated data.

\section{Experiments}
\label{sec:E}

\subsection{Classification Network}

\begin{figure*}[t]
    \hspace{75pt}Nevus \hspace{120pt} Melanoma \hspace{95pt} Seborrheic Keratosis\\
    \includegraphics[width=0.075\textwidth]{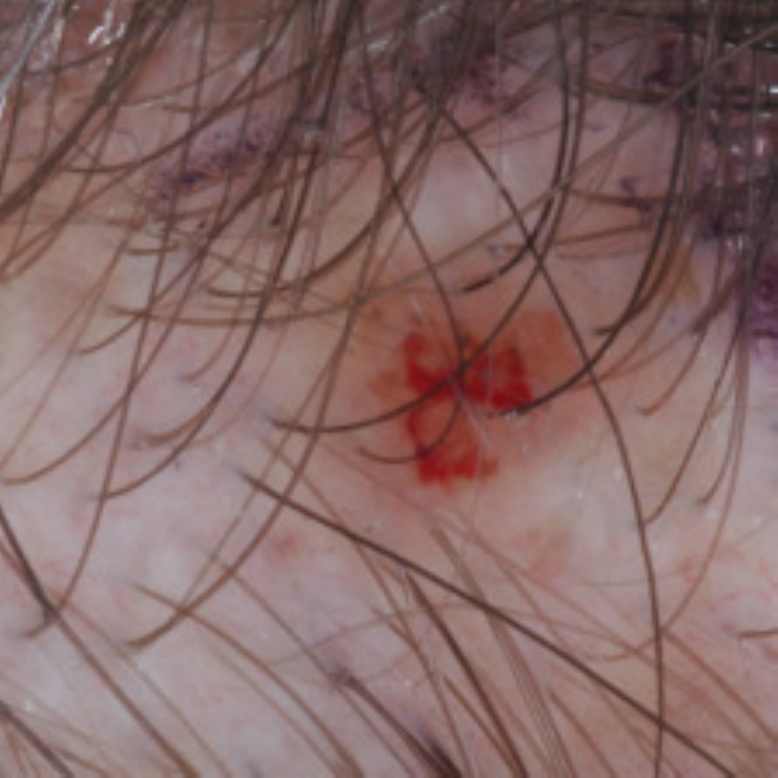}
    \includegraphics[width=0.075\textwidth]{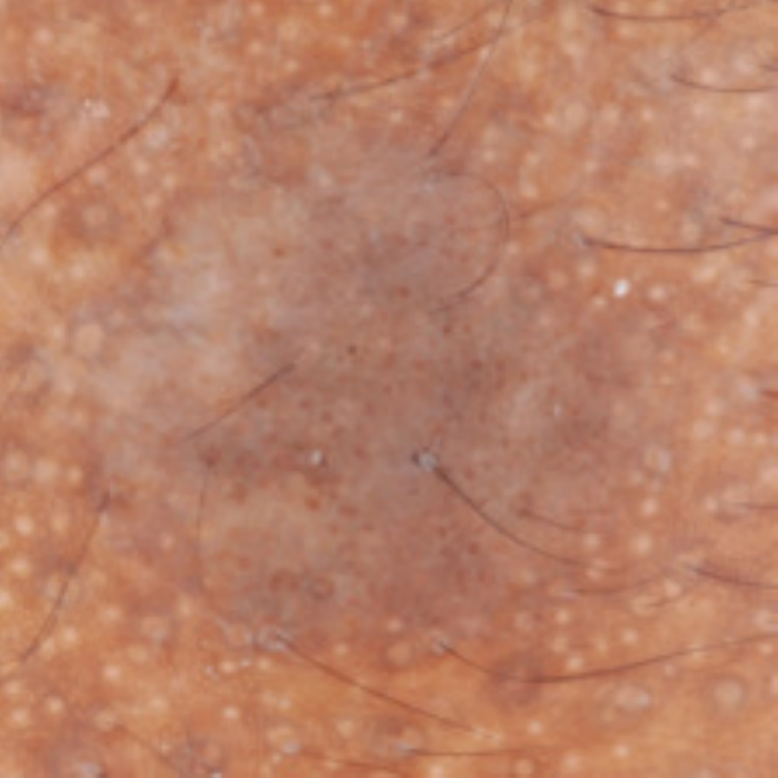}
    \includegraphics[width=0.075\textwidth]{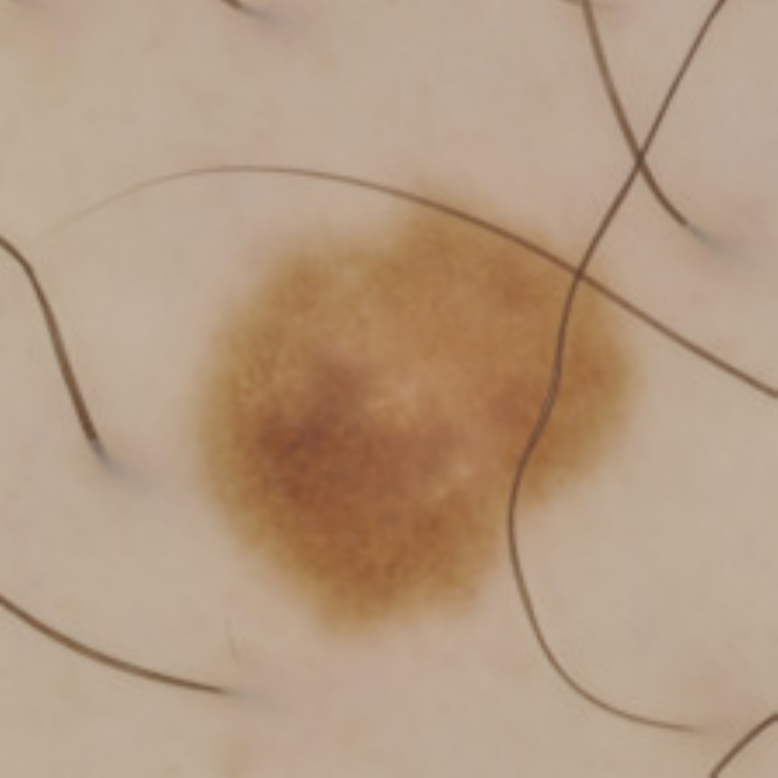}
    \includegraphics[width=0.075\textwidth]{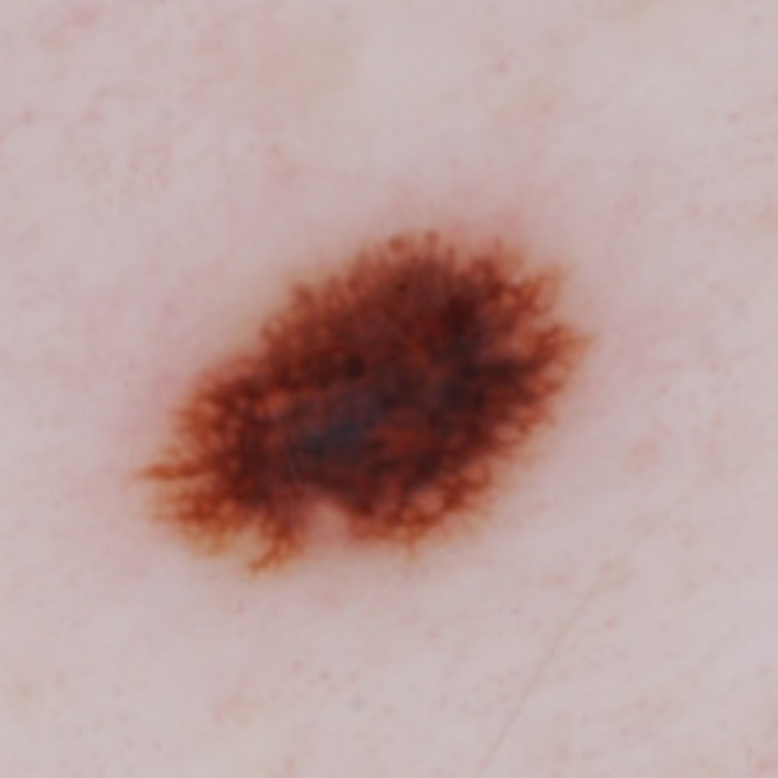}
    \hspace{5pt}
    \includegraphics[width=0.075\textwidth]{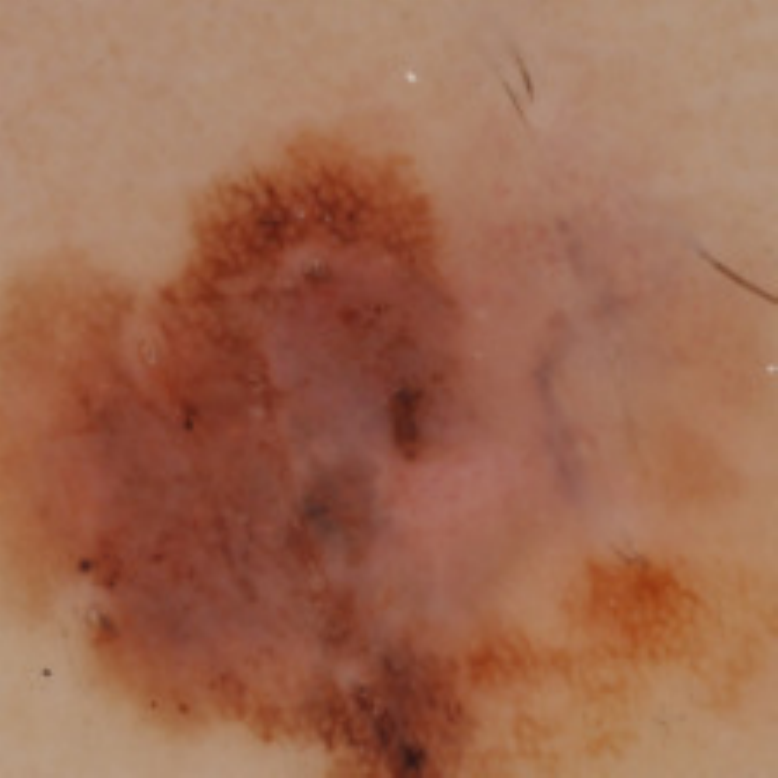}
    \includegraphics[width=0.075\textwidth]{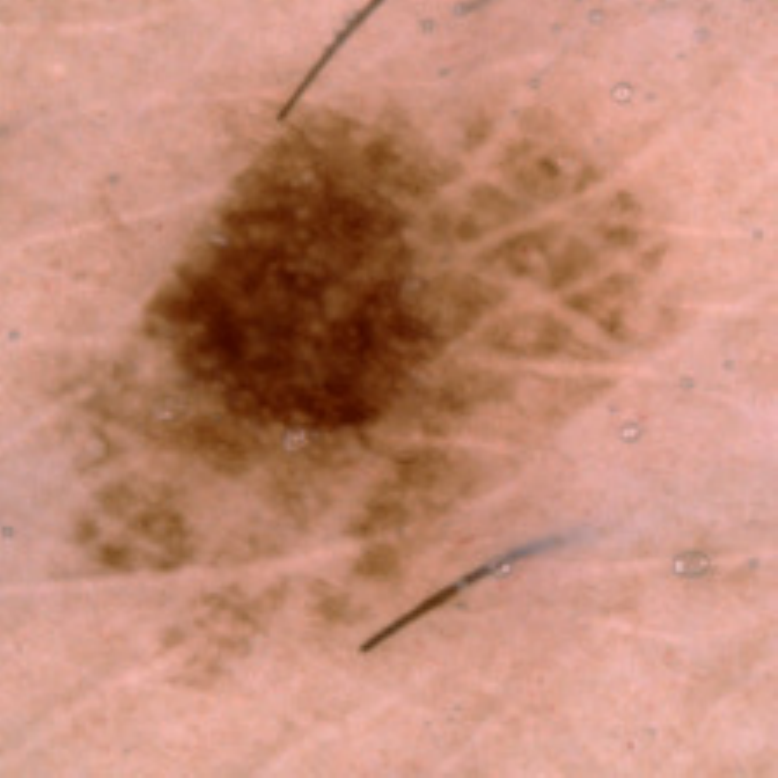}
    \includegraphics[width=0.075\textwidth]{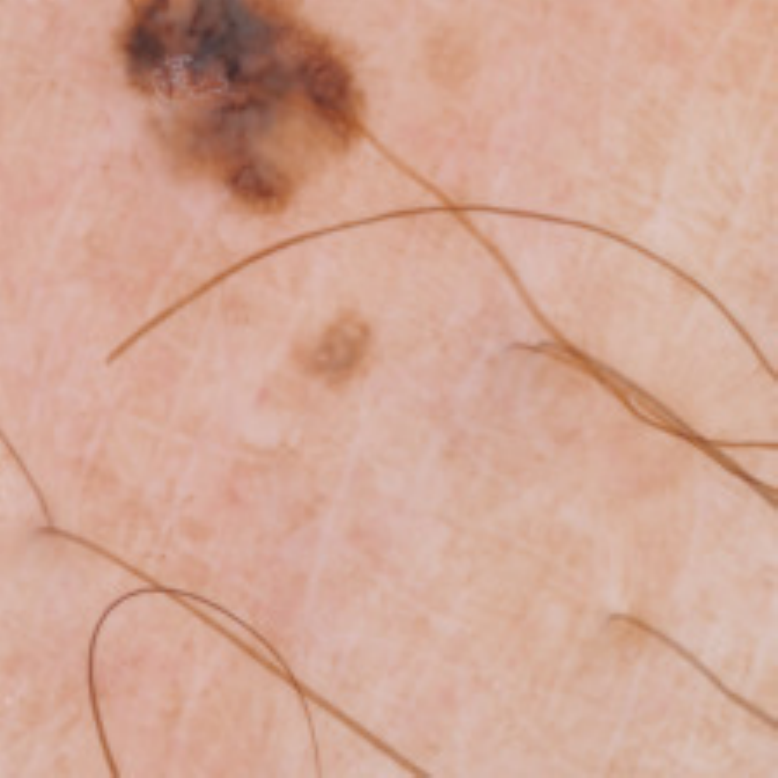}
    \includegraphics[width=0.075\textwidth]{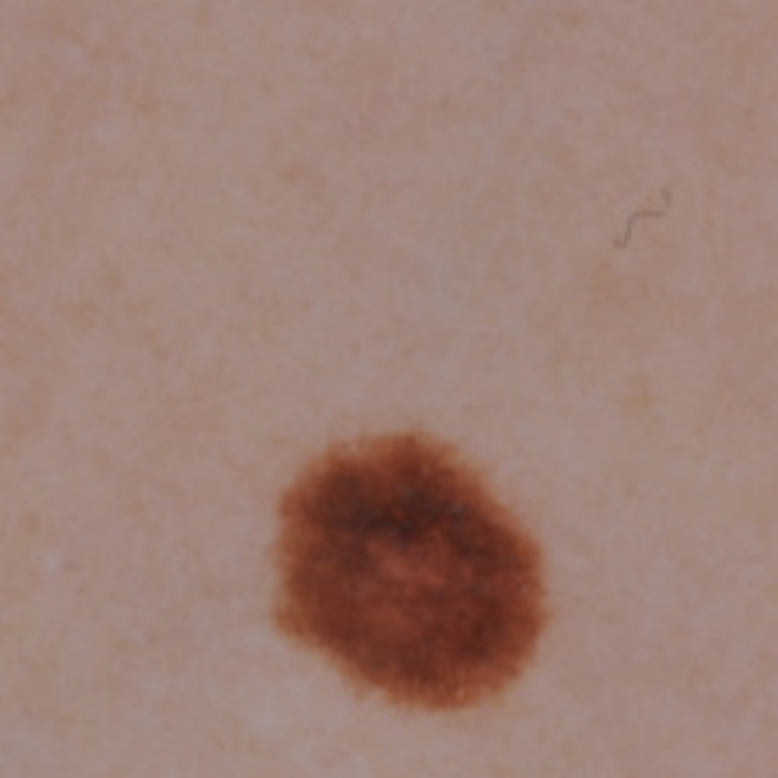}
    \hspace{5pt}
    \includegraphics[width=0.075\textwidth]{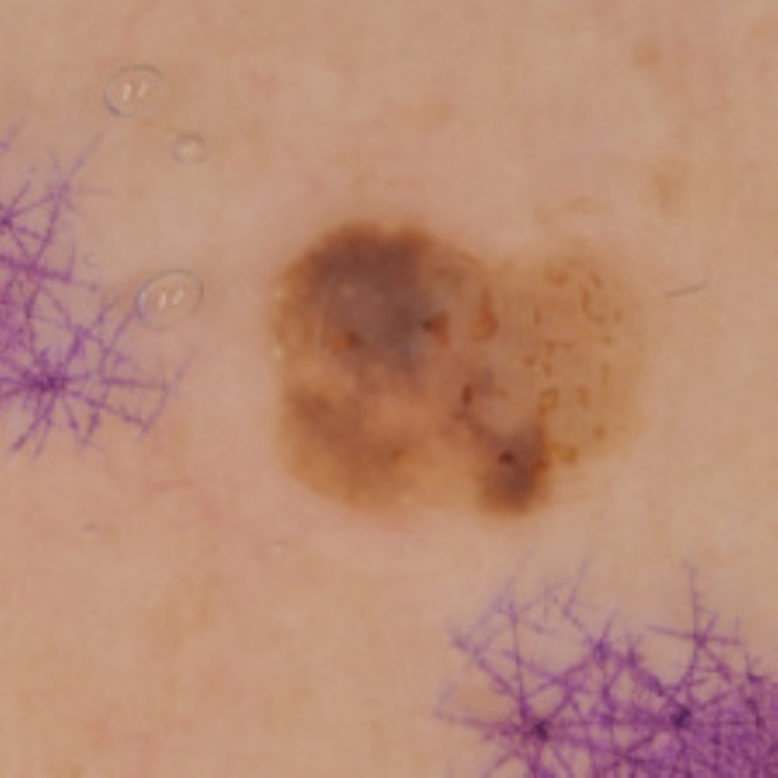}
    \includegraphics[width=0.075\textwidth]{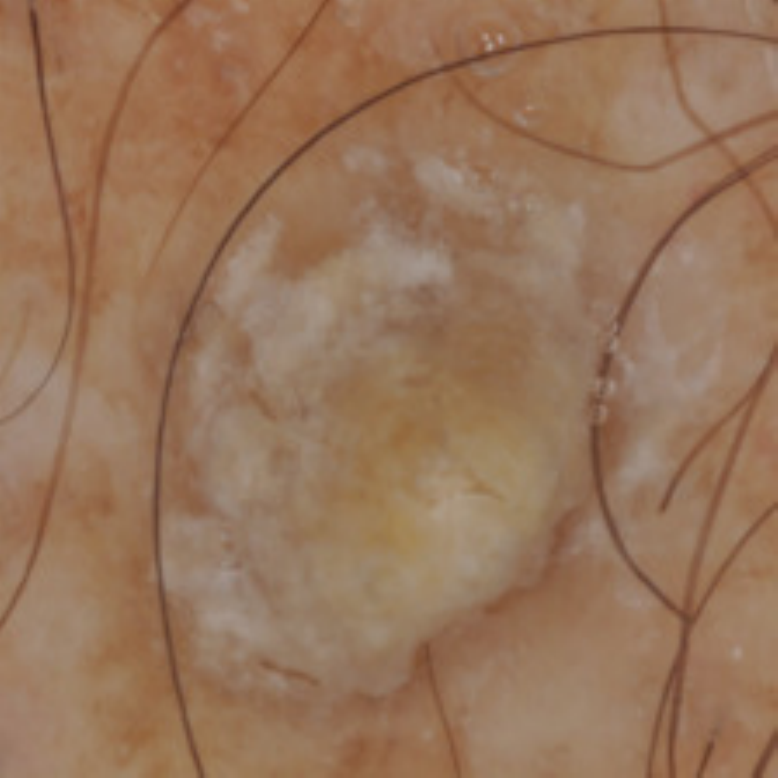}
    \includegraphics[width=0.075\textwidth]{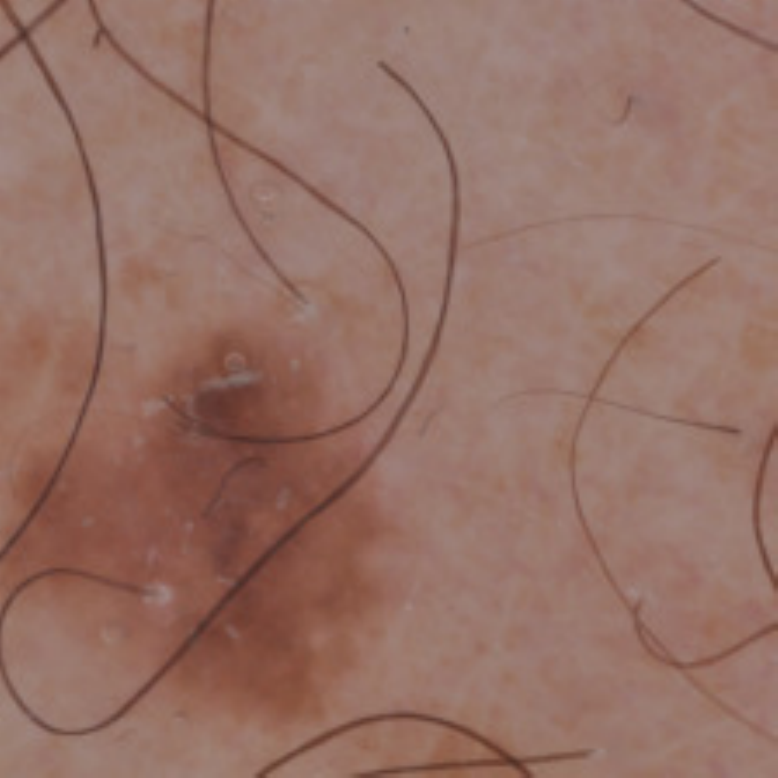}
    \includegraphics[width=0.075\textwidth]{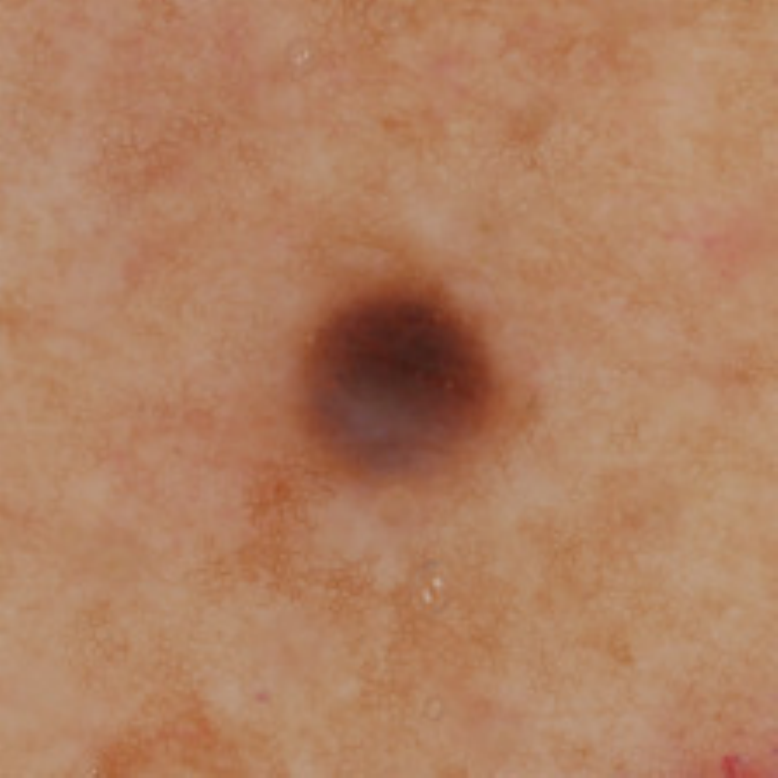}
    
    \stackunder[5pt]{\includegraphics[width=0.075\textwidth]{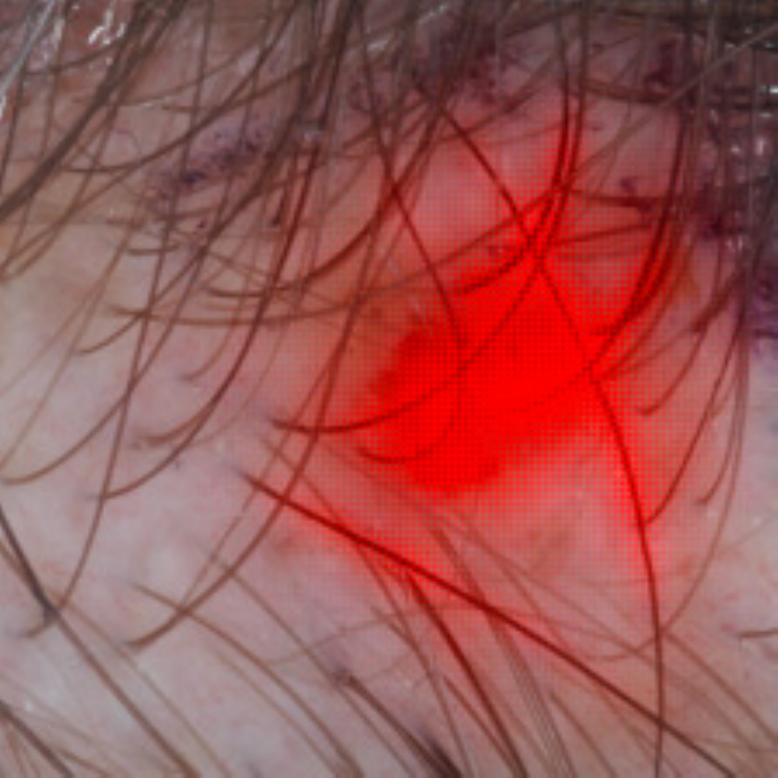}}{(TP)}
    \stackunder[5pt]{\includegraphics[width=0.075\textwidth]{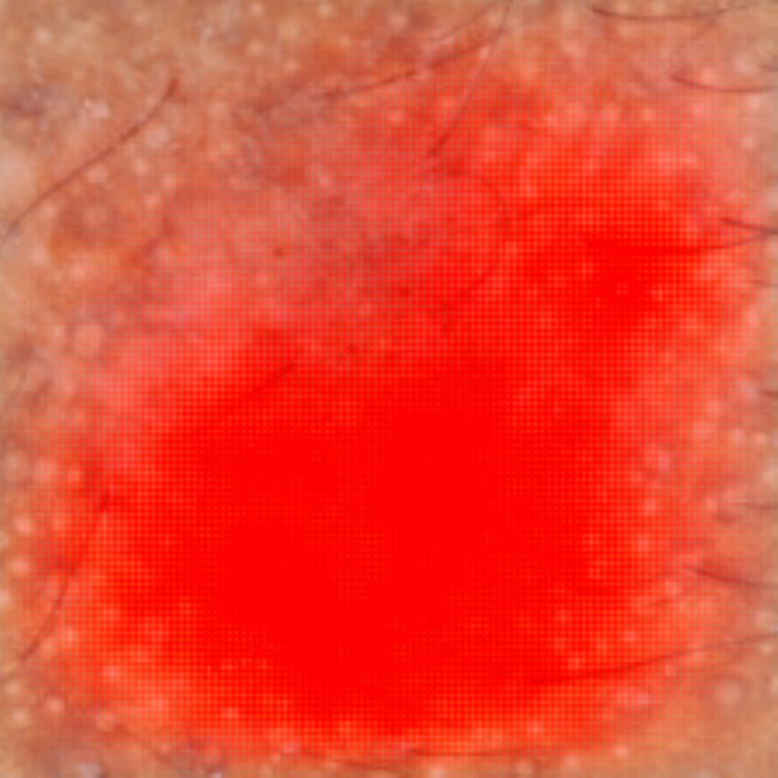}}{(FP)}
    \stackunder[5pt]{\includegraphics[width=0.075\textwidth]{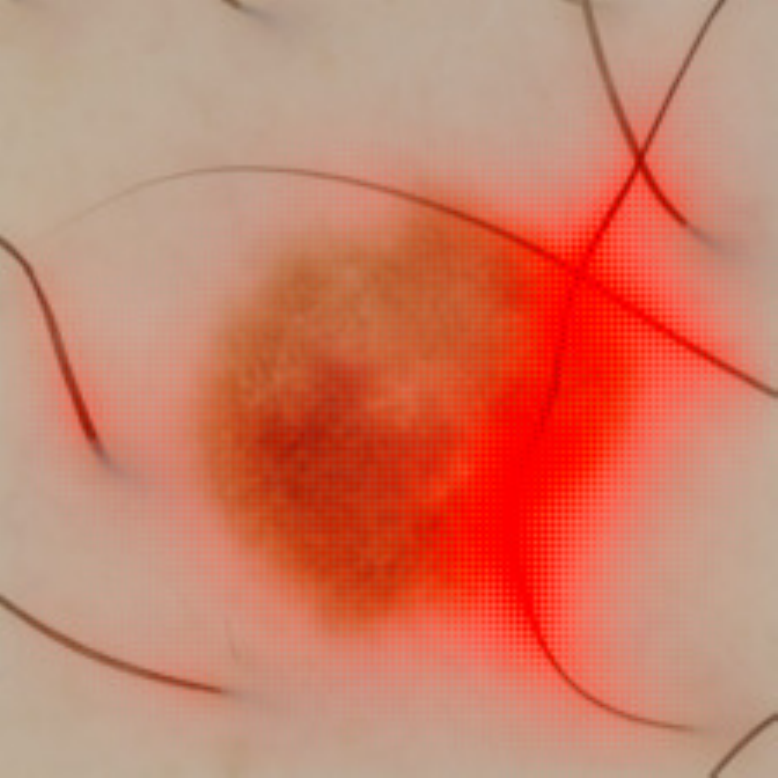}}{(FN)}
    \stackunder[5pt]{\includegraphics[width=0.075\textwidth]{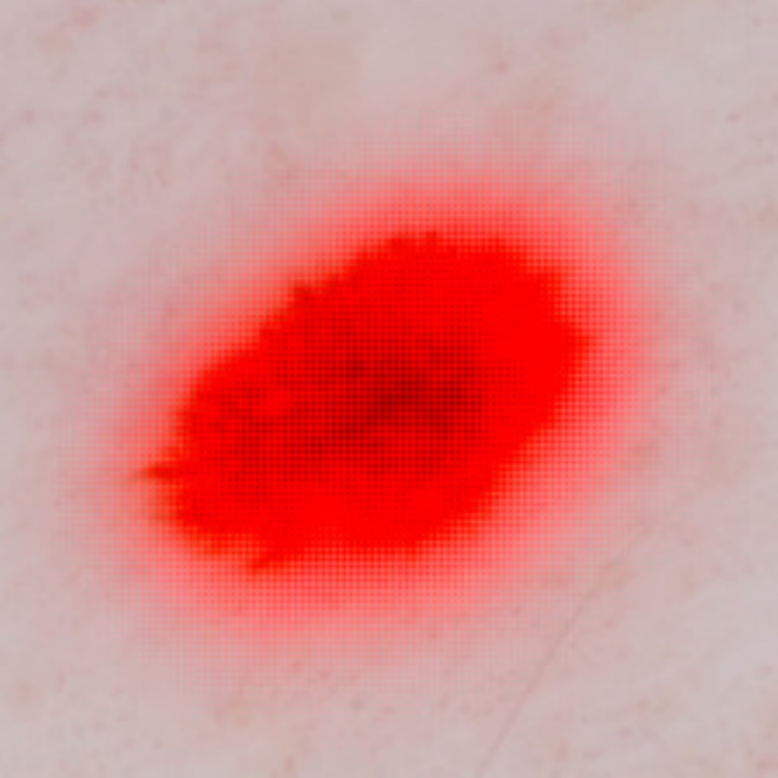}}{(TN)}
    \hspace{5pt}
    \stackunder[5pt]{\includegraphics[width=0.075\textwidth]{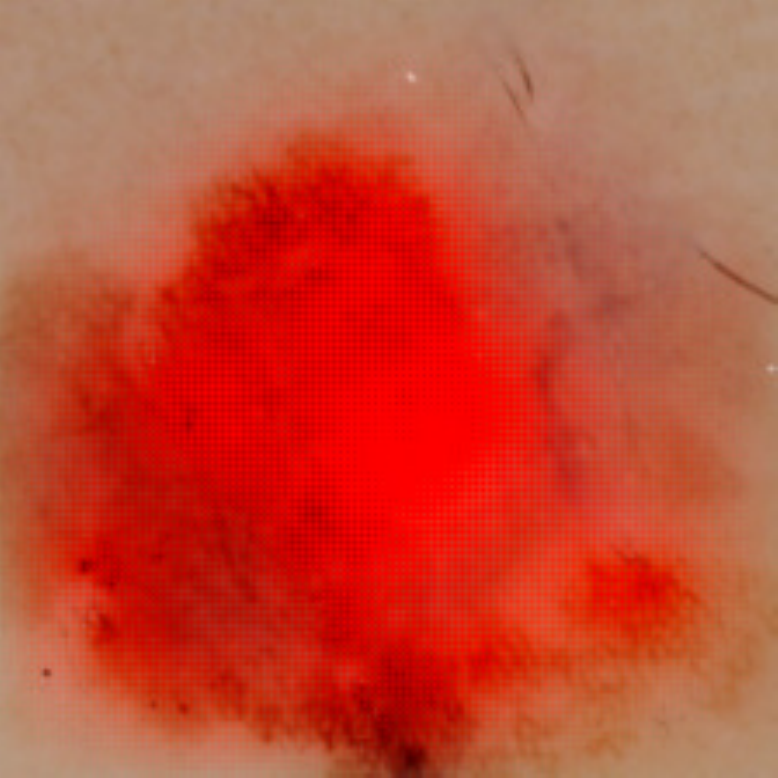}}{(TP)}
    \stackunder[5pt]{\includegraphics[width=0.075\textwidth]{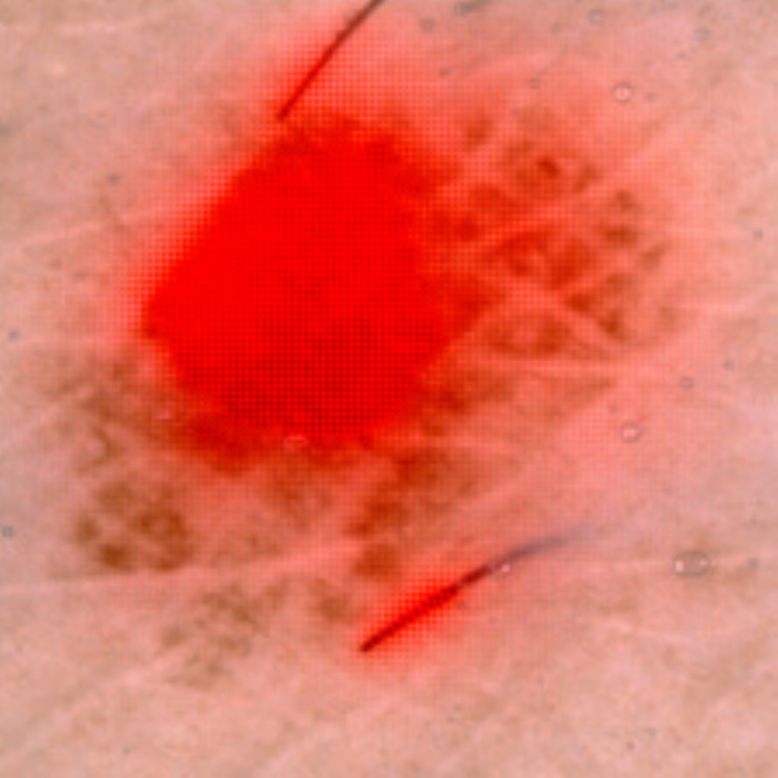}}{(FP)}
    \stackunder[5pt]{\includegraphics[width=0.075\textwidth]{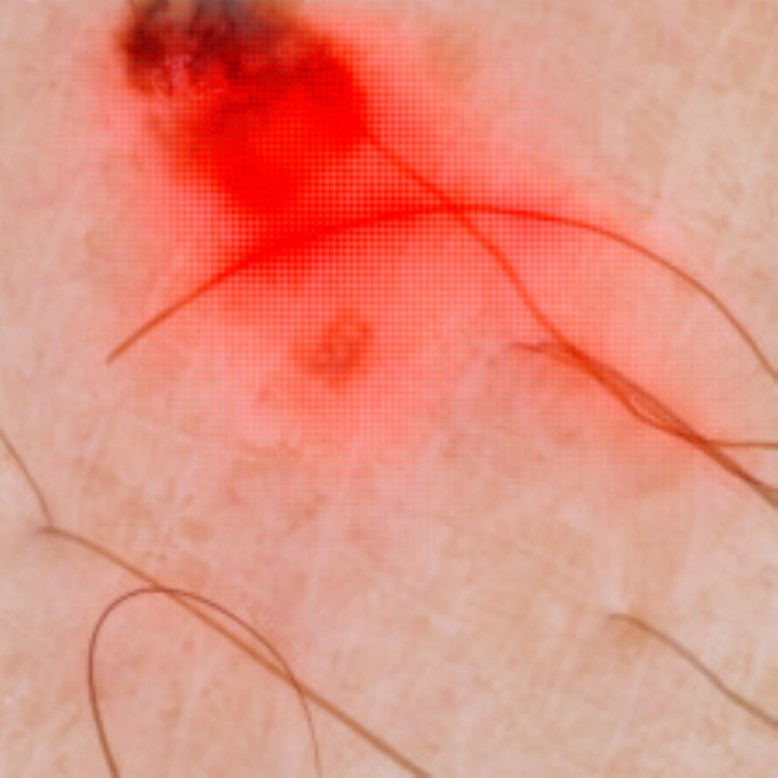}}{(FN)}
    \stackunder[5pt]{\includegraphics[width=0.075\textwidth]{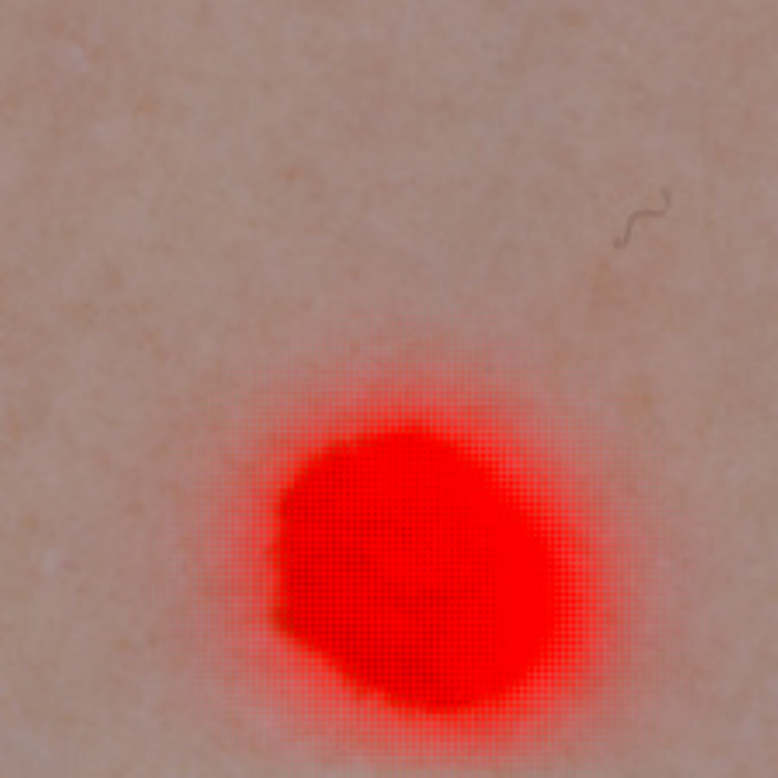}}{(TN)}    
    \hspace{5pt}
    \stackunder[5pt]{\includegraphics[width=0.075\textwidth]{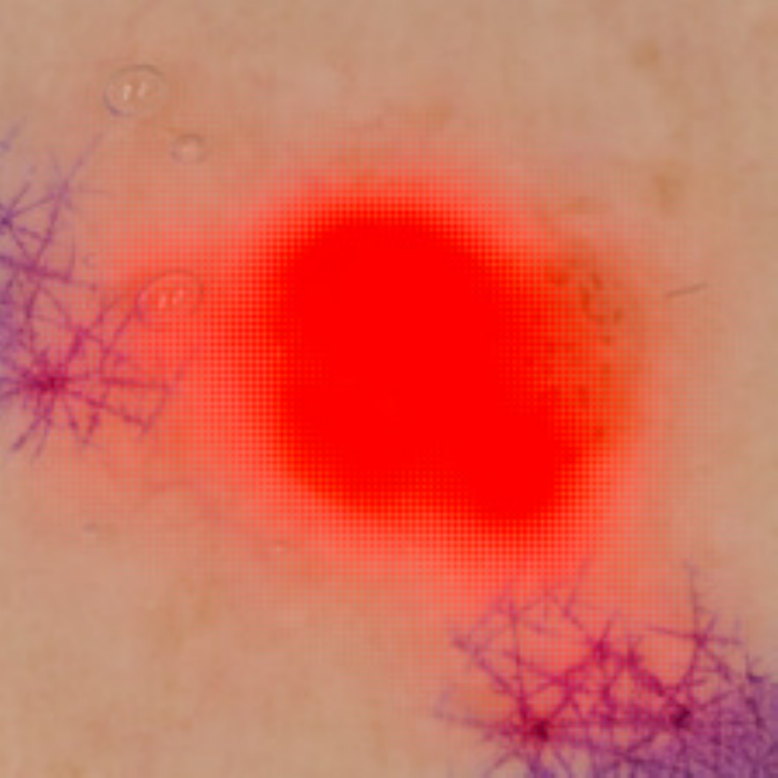}}{(TP)}
    \stackunder[5pt]{\includegraphics[width=0.075\textwidth]{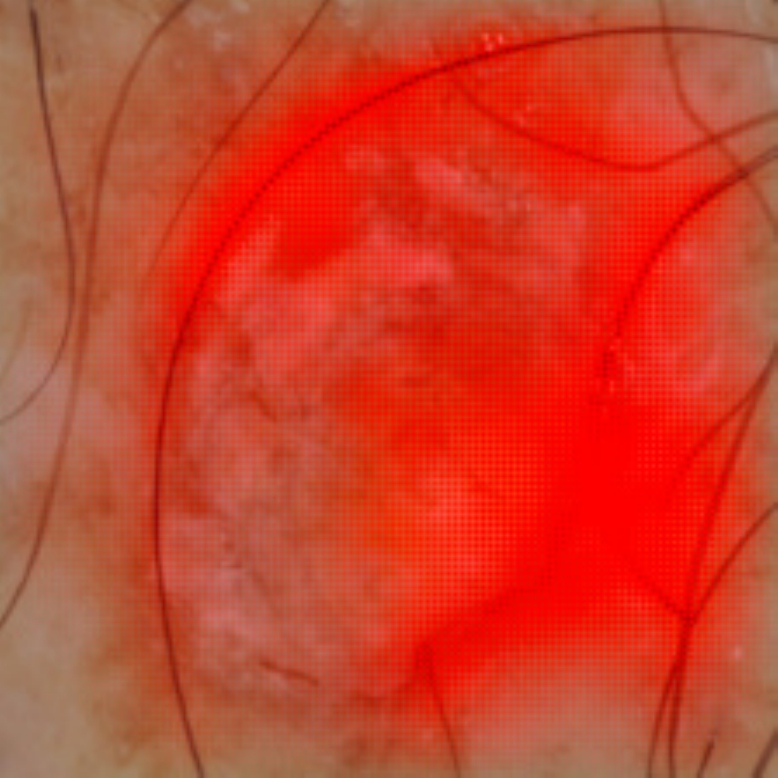}}{(FP)}
    \stackunder[5pt]{\includegraphics[width=0.075\textwidth]{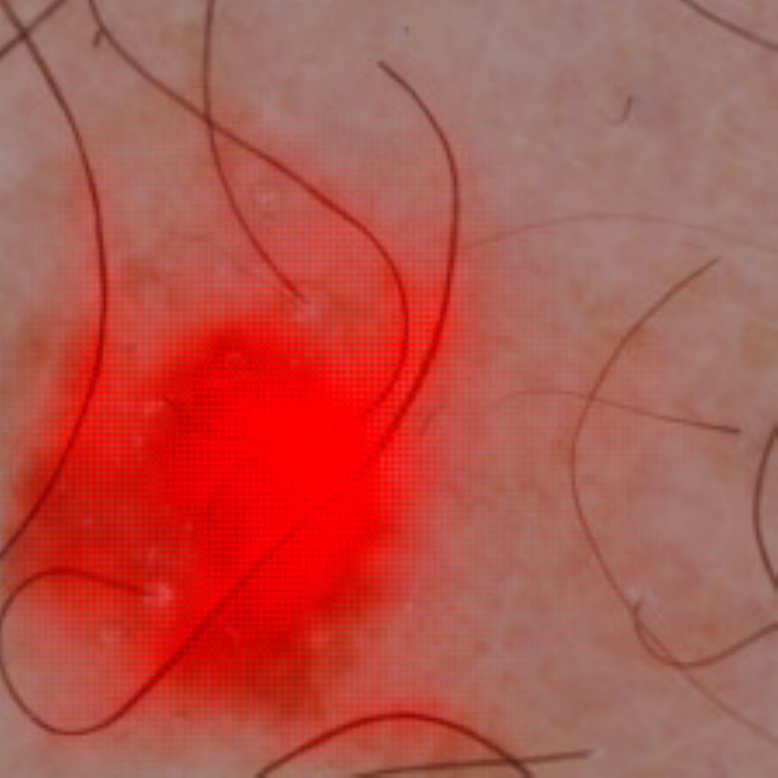}}{(FN)}
    \stackunder[5pt]{\includegraphics[width=0.075\textwidth]{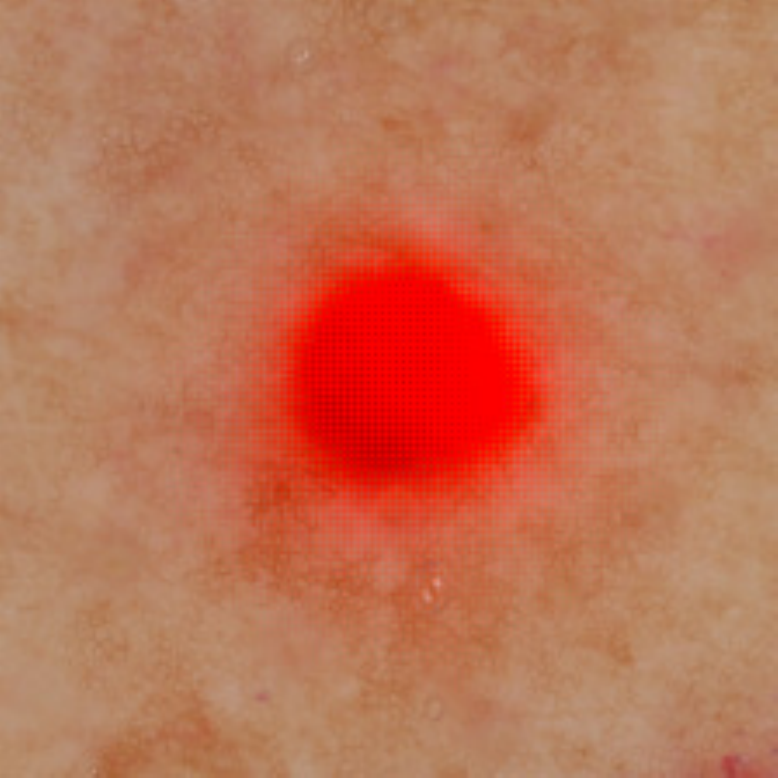}}{(TN)}
\caption{ \textbf{Top} Original image,  \textbf{Bottom} Visualization result for nevus, melanoma, and seborrheic keratosis for True Positive (TP), False Positive (FP), False Negative (FN), and True Negative (TN)}
\label{fig:class_visualization}
\end{figure*}
For the task of lesion classification, we utilized a ResNet-50 \cite{resnet} architecture pre-trained on ImageNet data set with final fully-connected layer modified to output probabilities of lesion being in each of the classes. We furthermore processed all the images in our training data to remove occlusions as described in the previous section. The pre-processed images were then added to the training data set of the lesion classification model to make it more robust to the presence of occlusions and prevent over-fitting. We then also augmented the data as described before. See Figure~\ref{fig:il} for the illustration of the process of obtaining training data for the classification network (this is discussed in details in previous sections - they also provide the numerical description of the changes of the data size that are incurred by purification and augmentation steps).

\subsection{Experimental results}
\begin{figure}[t]
    \centering
    \includegraphics[width=0.23\textwidth]{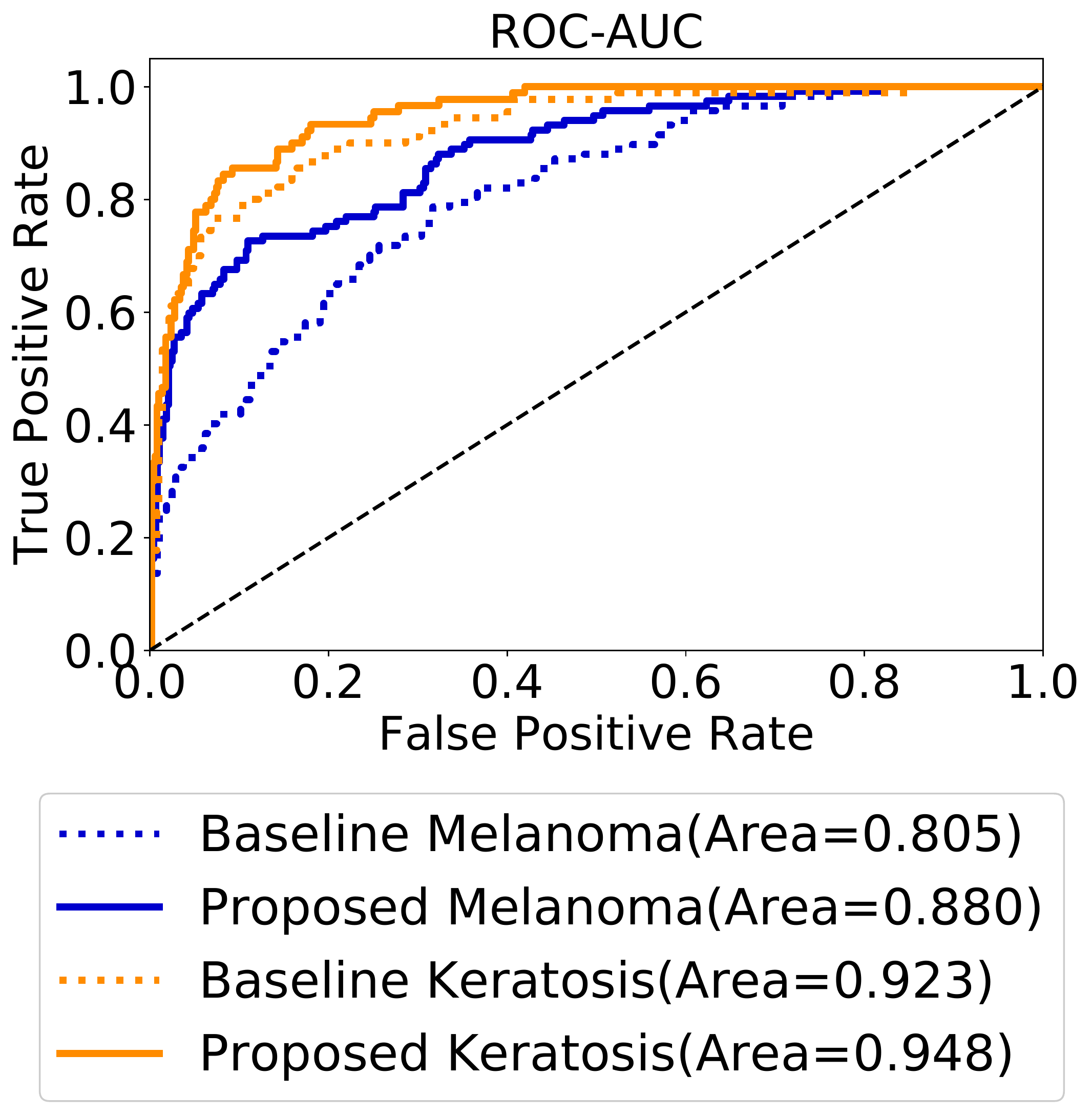}
    \includegraphics[width=0.23\textwidth]{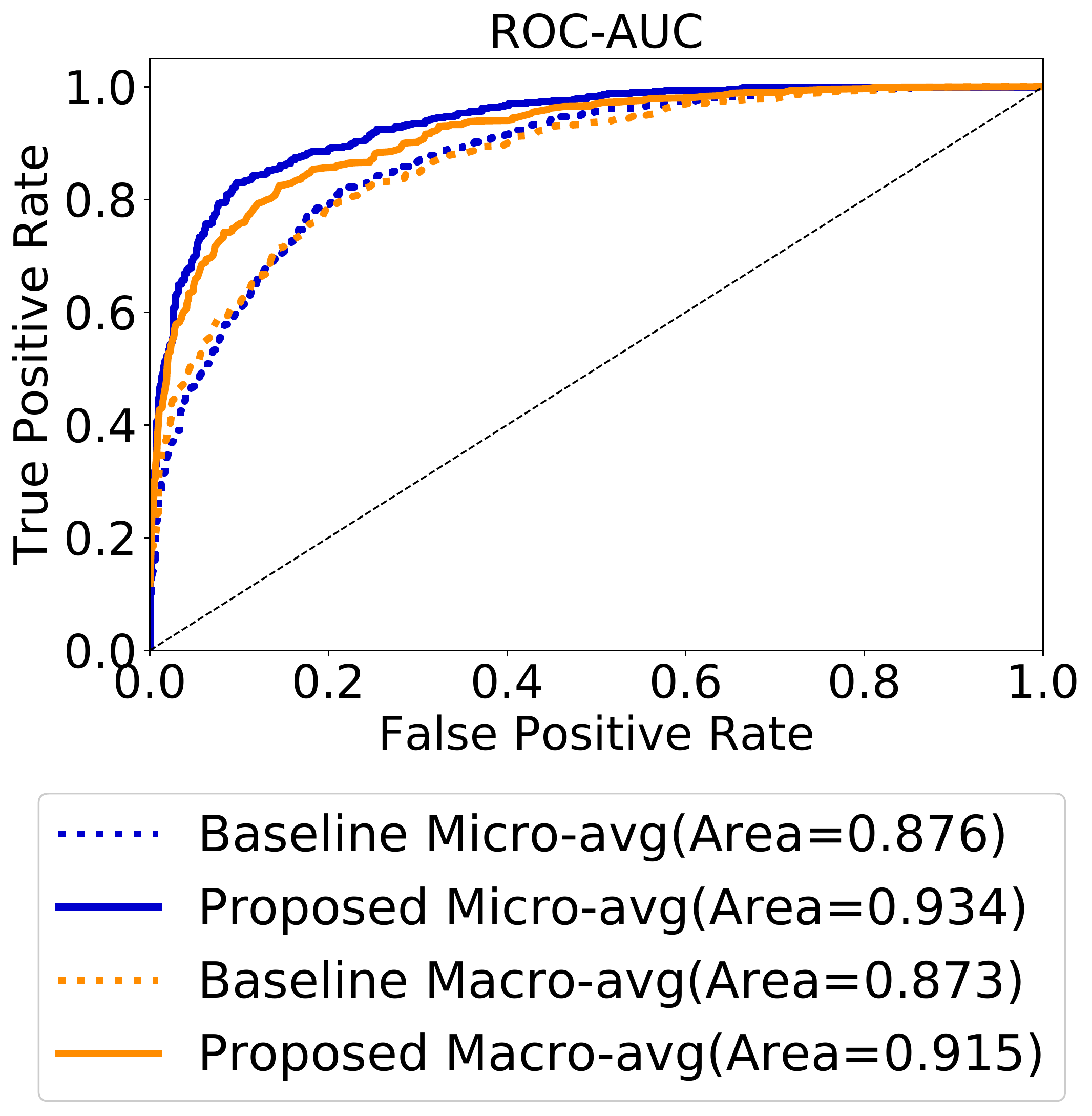}
    \includegraphics[width=0.23\textwidth]{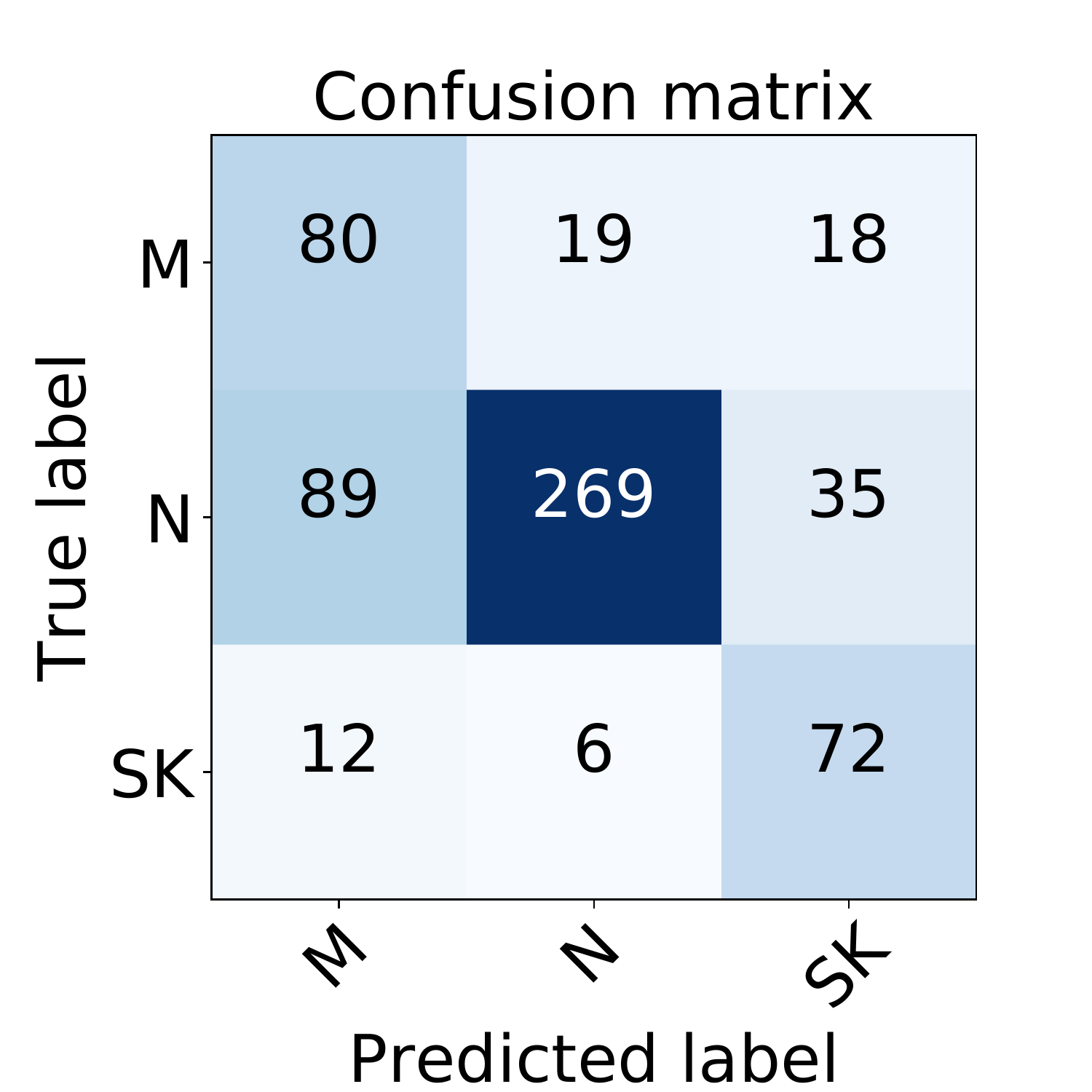}
    \includegraphics[width=0.23\textwidth]{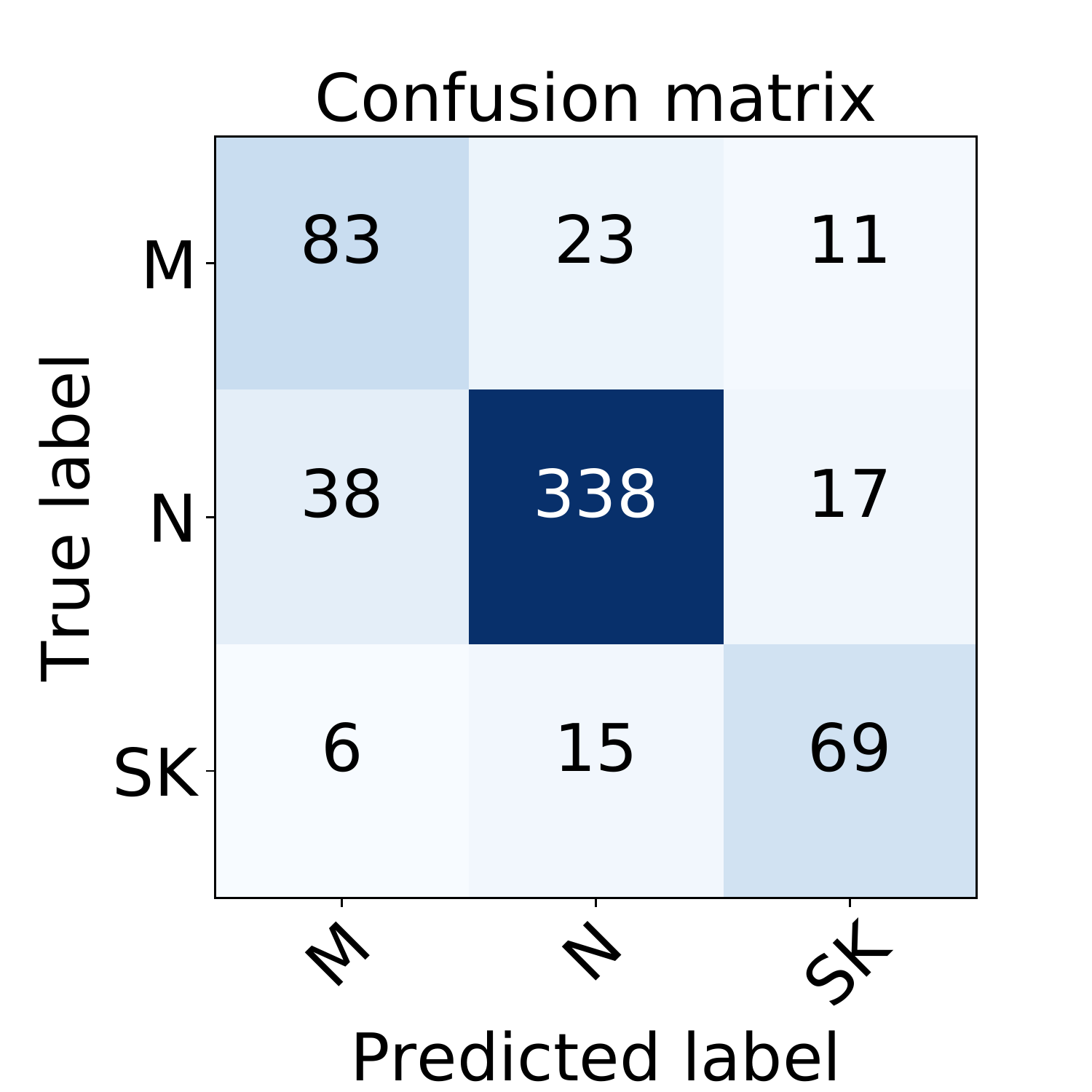}
    \caption{\textbf{(Top)} ROC curves obtained by traditional baseline (ResNet-50 pretrained on ImageNet and fine tuned on raw data without any purification/augmentation) \textbf{(Left)} and proposed classification model \textbf{(Right)} for ISIC 2017 test data-set. \textbf{(Bottom)} Confusion matrix obtained by traditional baseline (\textbf{left}) and proposed model (\textbf{right}). M - melanoma, N - nevus, SK - seborrheic keratosis.}
    \label{fig:curves}
\end{figure}
For empirical evaluation we used the images from the ISIC test data set. It contains $600$ images ($117$ melanoma images, $90$ seborrheic keratosis images, and $393$ nevus images), from which we did not remove the hairs and rulers. We used the main evaluation metrics defined in the ISIC 2017 challenge: area under the receiver operating characteristic curve (ROC AUC) for melanoma classification and ROC AUC for melanoma and seborrheic keratosis classifications combined (mean value). We compare our performance with the baseline model, i.e. ResNet-50 pretrained on ImageNet and fine tuned on raw data without performing any purification/augmentation, and the winning models of the ISIC 2017 challenge (note that these models were also using data sets from different digital lesion libraries for training their models, similarly to our approach). 

The ROC-AUC values obtained for different models are highlighted in Table \ref{tab:my_label}. We obtained ROC AUC of $0.88$ for melanoma classification and the mean performance of $0.915$ and outperform both the baseline, as shown in Figure~\ref{fig:curves}, as well as the winners of the challenge that obtained ROC AUC of $0.874$ for melanoma and average ROC AUC of $0.911$. Thus, leading to performance improvement of order $\approx 4\%$ We also report there the resulting specificity values for different values of sensitivity for melanoma classification. Note that in a separate study by Esteva et al.~\cite{Esteva2017}, dermatologists were asked to classify dermoscopic images into three categories: benign, malignant (melanoma) and non-neoplastic. Two dermatologists attained $65.56\%$ and $66.0\%$ accuracy on a small subset of the data corpus owned and maintained by the research group and not open-sourced to public. We obtain $81.6\%$ accuracy on the task of classifying images into three categories (this number was computed based on the confusion matrix in Figure~\ref{fig:curves}): benign (nevus), malignant (melanoma), and seborrheic keratosis which, in the prism of this study, is much higher than the average accuracy achieved by dermatologists. 

\begin{table}[t]
    \centering
    \begin{tabular}{|c|c|c|}
        \hline
        & \multicolumn{2}{c|}{ROC-AUC} \\
        Model & Melanoma & Avg  \\
        \hline 
        Traditional Baseline & 0.805 & 0.873\\
        \hline 
        Proposed Model & \textbf{0.880} & \textbf{0.915} \\
        \hline  
        ISIC Challenge winners \cite{DBLP:journals/corr/MatsunagaHMK17,DBLP:journals/corr/MenegolaTFLAV17} & 0.874 & 0.911 \\
        \hline
    \end{tabular}
    \caption{ROC AUC values for traditional baseline (ResNet-50 pretrained on ImageNet and fine-tuned of raw data without any purification/augmentation) and winners for ISIC challenge.}
    \label{tab:my_label}
\end{table}
\begin{table}[htp!]
\begin{tabular}{|c|c|c|c|}
        \hline
        Method & $82\%$ & $89\%$ & $95\%$ \\
        \hline
        Top AVG \cite{DBLP:journals/corr/MatsunagaHMK17} & 0.729 & 0.588 & 0.366 \\
        \hline
        Top SK \cite{DBLP:journals/corr/Gonzalez-Diaz17} & 0.727 & 0.555 & 0.404 \\
        \hline
        Top M \cite{DBLP:journals/corr/MenegolaTFLAV17} & \textbf{0.747} & 0.590 & 0.395 \\
        \hline
        Our Classification Model & 0.697 & \textbf{0.648} & \textbf{0.492}  \\
        \hline
    \end{tabular}
    \caption{Specificity values at sensitivity levels of $82\%/89\%/95\%$ for melanoma classification. Top AVG, Top SK, and Top M denote the winning approaches of the ISIC 2017 challenge.}
    \label{tab:values1}
\end{table}

In Figure~\ref{fig:curves} we report the confusion matrix for predicting melanoma, seborrheic keratosis, and nevus. In Table \ref{tab:values1} we also report the resulting specificity values for different values of sensitivity for melanoma classification. It is clearly observed that we show superior performance over other methods for different levels of sensitivity.

The confusion matrix in Figure \ref{fig:curves} shows that the classification model occasionally confuses nevus and melanoma images.  We employ visualization technique called VisualBackProp to discover the reasons that led to these false positives and false negatives. The false positive case in (Figure \ref{fig:class_visualization}, Nevus) indicates that the lesion area is not well-identified by the deep learning model (most likely because it is blended with the skin surface). The false negative cases show that the occlusions still sometimes affect the prediction, though the pre-processing step has eliminated many such similar mis-classification cases, \eg.: see true positive cases for nevus (the model handles well hair occlusions) and seborrheic keratosis (the model leaves out the ink and focuses on the lesion). 

\section{Conclusion}
\label{sec:CC}
The techniques that aim at automating the visual examination of skin lesions, traditionally done by dermatologists, are nowadays dominated by the deep-learning-based methods. These methods are the most accurate and scalable, but they require large training data sets and thus their applicability in dermatology is compromised by the size of the publicly available dermatological data sets, which are often small and contain occlusions. We present a solution for this problem that relies on careful data purification that removes common occlusions from dermoscopic images and augmentation that uses the modern technique of deep-learning-based data generation to improve data balancedness. We demonstrate the effectiveness of our system on the lesion classification task.

{\small
\bibliographystyle{ieee_fullname}
\bibliography{ms}
}
\clearpage







\section{Supplementary Section}
In this section we present our early approach for the ISIC-2018 data set that rely on methods proposed in the main text. Section \ref{supp:data_puri} describes deep-learning-based approach for data purification, Section \ref{supp:imba} and Section \ref{supp:imba2} address the problem of data imbalancedness for ISIC-2018 and propose a solution that rely on coupled DCGAN model proposed in the main text, Section \ref{supp:class} incorporates both algorithms into a classification network. 

\subsection{Data Purification}
\label{supp:data_puri}
\begin{figure}[t!]
    \includegraphics[width=0.09\textwidth]{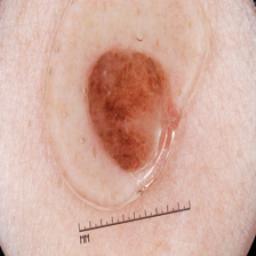}
    \includegraphics[width=0.09\textwidth]{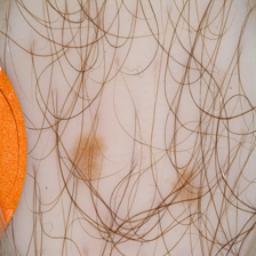}
    \includegraphics[width=0.09\textwidth]{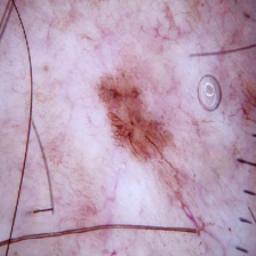}
    \includegraphics[width=0.09\textwidth]{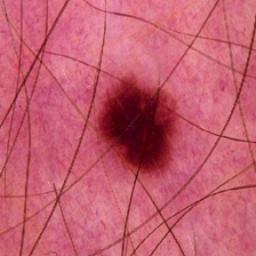}
    \includegraphics[width=0.09\textwidth]{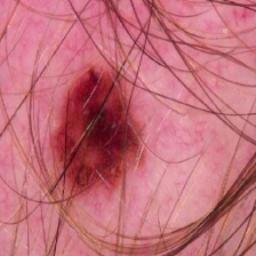}

    \includegraphics[width=0.09\textwidth]{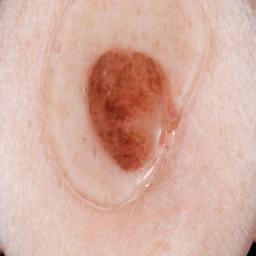}
    \includegraphics[width=0.09\textwidth]{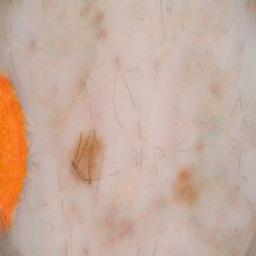}
    \includegraphics[width=0.09\textwidth]{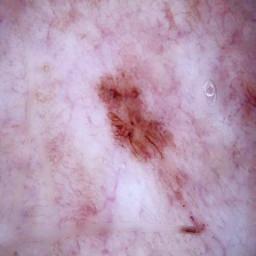}
    \includegraphics[width=0.09\textwidth]{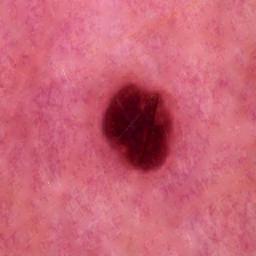}
    \includegraphics[width=0.09\textwidth]{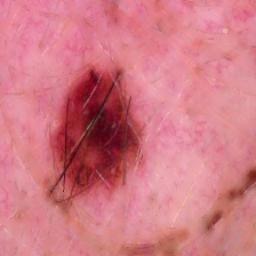}
    \caption{\textbf{(Top)}: Original images. \textbf{(Bottom)}: Images after removing occlusions, i.e. hairs and rulers, using Data Purification Network.}
    \label{fig:hair_removal}
\end{figure}

Hair removal algorithm described in main section is not scalable to massive datasets as it requires manual tuning of parameters. These parameters vary with type of images (dermoscopic or clinical), size of the image, color of the hair (black or blonde) and the amount of hair present in the image. Thus, a deep learning solution was desired to fully incorporate it into our  classification pipeline for ISIC 2018 task. We utilized a Unet based encoder-decoder model architecture with convolution operations replaced by partial convolutions~\cite{DBLP:journals/corr/abs-1804-07723} to improve the performance of the model (Partial convolution is often referred as segmentation-aware convolution. The intuition behind using partial convolutions arises from the fact that given an input image and the corresponding binary mask, the output of the convolution operation should only depend on the regions of the image that are not zeroed-out by the mask and should not take into account regions where pixels values are zero.). We refer to this network as data purification network. The input of the network is the original image and the output is the same image after occlusion removal. Since the training data set for this model requires coupled pairs of images before and after occlusion removal and those, to the best of our knowledge, are not available in any public data sets, we created such training data using the algorithm described in the main text. The obtained training data-set consisted of 16, 270 images and was augmented through random masking. The data purification network  was trained using the loss proposed for training networks with partial convolutions \cite{DBLP:journals/corr/abs-1804-07723} and that targets both per-pixel reconstruction accuracy ans well as composition i.e how smoothly the predicted hole values transition into their surrounding context. The data purification network was trained using an Adam optimizer with beta values set to $(0.5, 0.99)$ and constant learning rate set to $2e^{-4}$ in the beginning of training and SGD optimizer with momentum $0.9$ and learning rate $1e^{-4}$ in the later stages of learning. The model was trained over a week on $4$ GTX $1080$ $12$Gb GPU cards. Figure~\ref{fig:hair_removal} show the results of data purification obtained by our model. Figure \ref{fig:hair_removal} show the results of data purification obtained by our model. We processed all the images in ISIC-2018 data-set to remove occluded objects. The preprocessed images were then added to the training data set of the lesion classification model to make it more robust to the presence of occlusions and prevent overfitting. 

\subsection{Data Imbalancedness}
\label{supp:imba}
\begin{figure}[htp!]
    \centering
    \includegraphics[width=0.35\textwidth]{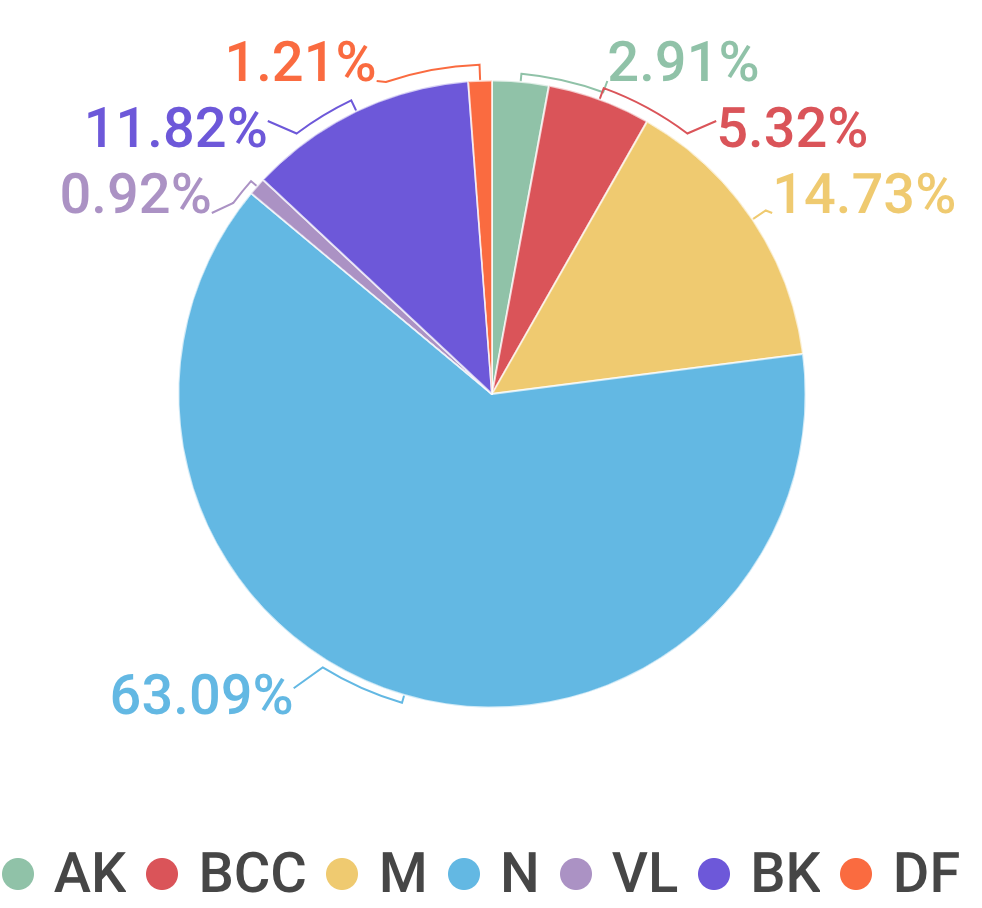}
\caption{Sizes of the training data sets for the ISIC 2018 task. Data set is heavily imbalanced.}
\label{fig:il2}
\end{figure}

Figure \ref{fig:il2} highlights the class imbalancedness problem for ISIC-2018 challenge dataset. The dataset contains $452$ cases of acitinic keratosis (AK), $825$ cases of basal cell carcinoma (BCC), $1833$ samples of benign keratosis (BK), $187$ cases of dermatofibroma (DF), $2285$ cases of melanoma (M), $9786$ cases of nevus (N), and $142$ cases of vascular lesion (VL)). It is clearly observed Dermatofibroma constitute less than 2\% of the entire dataset.

\subsection{Data Generation with coupled DCGANs}
\begin{table*}[t!]
    \centering
    \renewcommand{\arraystretch}{1.2} 
    \begin{tabular}{|c|p{5em}|p{6em}|l|l|l|l|}
        \hline
        & Layer & Output Size & Kernel & Stride  & Padding \\
        \hline
        \multirow{4}{7em}{Generator - Weight sharing block} 
        & TransConv & 256$\times$4$\times$4 & 4$\times$4  & 1 & 0  \\
        & TransConv & 128$\times$8$\times$8 & 4$\times$4 & 2 & 1   \\
        & TransConv & 64$\times$16$\times$16 & 4$\times$4 & 2 & 1  \\
        & TransConv & 32$\times$32$\times$32 & 4$\times$4 & 2 & 1  \\
        \hline
        \multirow{3}{7em}{Generator- Class specific block} 
        & TransConv & 16$\times$64$\times$64 & 4$\times$4 & 2 & 1  \\
        & TransConv & 8$\times$128$\times$128 & 4$\times$4 & 2 & 1 \\
        & TransConv & 3$\times$256$\times$256 & 4$\times$4 & 2 & 1 \\
        \hline
        \multirow{3}{7em}{Discriminator - Class specific bloc}
        & Conv & 16$\times$128$\times$128& 4$\times$4  & 2 & 1\\
        & Conv & 32$\times$64$\times$64 & 4$\times$4 & 2 & 1  \\
        & Conv & 64$\times$32$\times$32 & 4$\times$4 &2 & 1   \\
        \hline
        \multirow{4}{7em}{Discriminator - Weight sharing block}
        & Conv & 128$\times$16$\times$16 & 4$\times$4 & 2 & 1 \\
        & Conv & 256$\times$8$\times$8 & 4$\times$4 & 2 & 1   \\
        & Conv & 512$\times$4$\times$4 & 4$\times$4 & 1 & 1   \\
        & Conv & 1$\times$1$\times$1 & 4$\times$4 & 1 & 0     \\
        \hline
    \end{tabular}
    \caption{Details of the architecture of the generative model. Let $n$ be the number of features maps, $h$ be the height and $w$ be the width. Size of the output feature maps is represented as $n\times h \times w$. Each convolution layer in the generator, except for the last one, is followed by a batch normalization and a ReLU nonlinearity. The last convolution layer is followed by a hyperbolic tangent. Similarly, each layer in the discriminator, except for the last convolution layer, is followed by a batch normalization and a leaky ReLU nonlinearity with the leakage coefficient of 0.2. The last convolution layer is followed by a sigmoid. The class specific blocks are repeated for $7$ different classes.}
    \label{tab:DCGAN2}
\end{table*}
\label{supp:imba2}
\begin{figure}[h]
    \centering
    (A)\begin{minipage}{0.45\textwidth}
    \includegraphics[width=\textwidth]{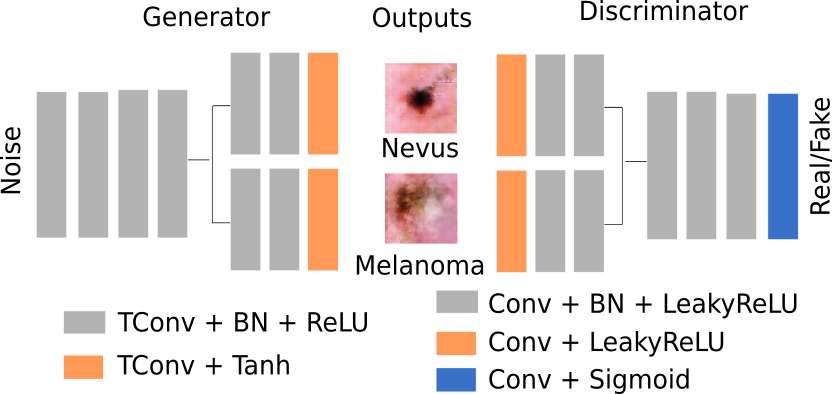}
    \end{minipage}\\
    (B)\begin{minipage}{0.46\textwidth}
    \includegraphics[width=0.13\textwidth]{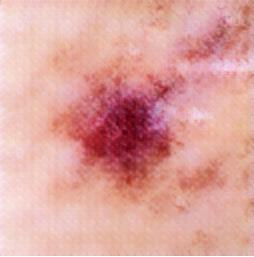}
    \includegraphics[width=0.13\textwidth]{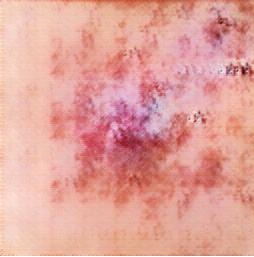}
    \includegraphics[width=0.13\textwidth]{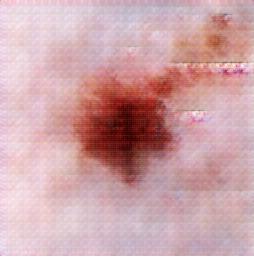}
    \includegraphics[width=0.13\textwidth]{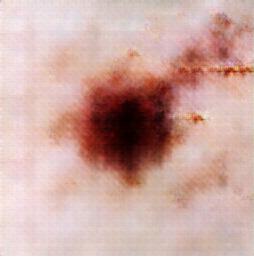}
    \includegraphics[width=0.13\textwidth]{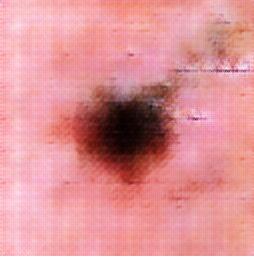}
    \includegraphics[width=0.13\textwidth]{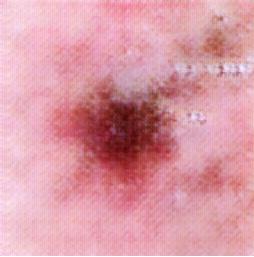}
    \includegraphics[width=0.13\textwidth]{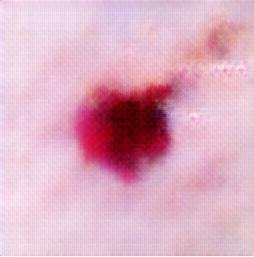}
    
    \includegraphics[width=0.13\textwidth]{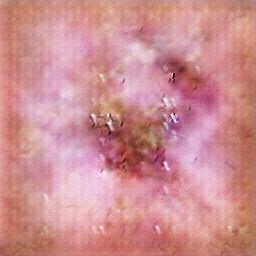}
    \includegraphics[width=0.13\textwidth]{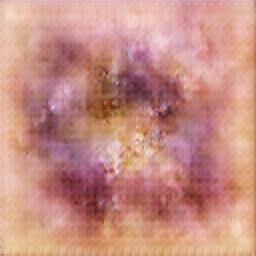}
    \includegraphics[width=0.13\textwidth]{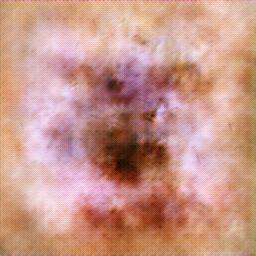}
    \includegraphics[width=0.13\textwidth]{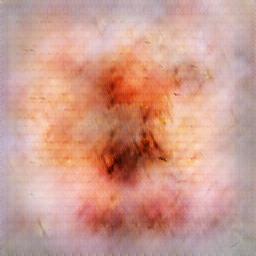}
    \includegraphics[width=0.13\textwidth]{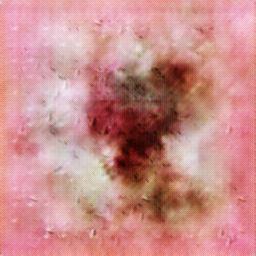}
    \includegraphics[width=0.13\textwidth]{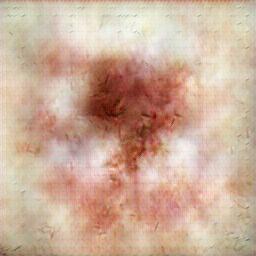}
    \includegraphics[width=0.13\textwidth]{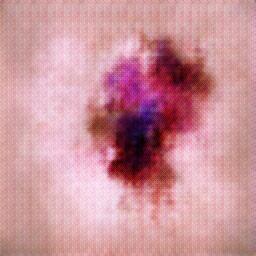}

    \includegraphics[width=0.13\textwidth]{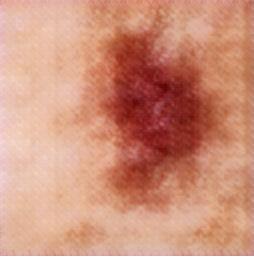}
    \includegraphics[width=0.13\textwidth]{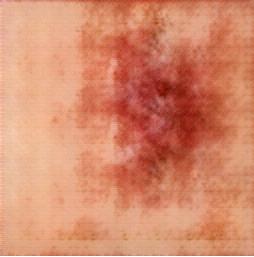}
    \includegraphics[width=0.13\textwidth]{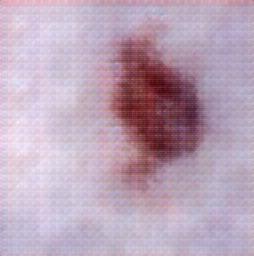}
    \includegraphics[width=0.13\textwidth]{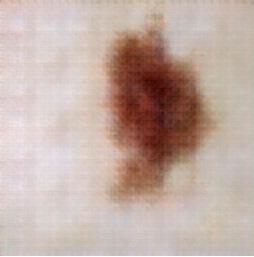}
    \includegraphics[width=0.13\textwidth]{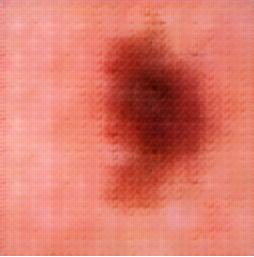}
    \includegraphics[width=0.13\textwidth]{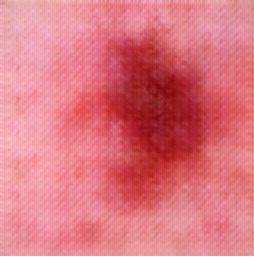}
    \includegraphics[width=0.13\textwidth]{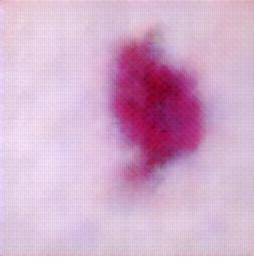}

    \stackunder{\includegraphics[width=0.13\textwidth]{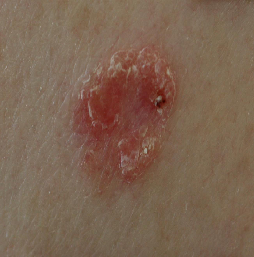}}{(P)}
    \stackunder{\includegraphics[width=0.13\textwidth]{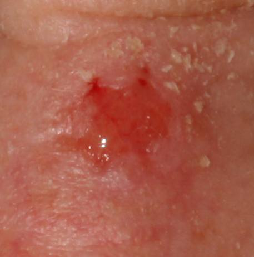}}{(Q)}
    \stackunder{\includegraphics[width=0.13\textwidth]{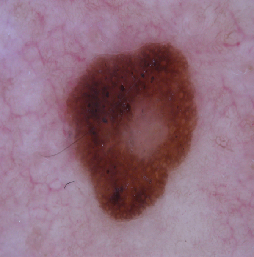}}{(R)}
    \stackunder{\includegraphics[width=0.13\textwidth]{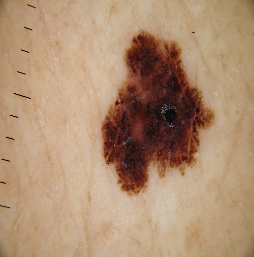}}{(S)}
    \stackunder{\includegraphics[width=0.13\textwidth]{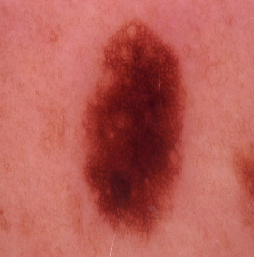}}{(T)}
    \stackunder{\includegraphics[width=0.13\textwidth]{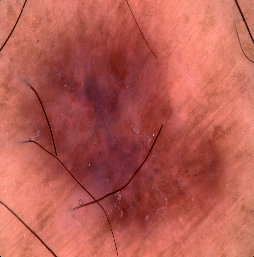}}{(U)}
    \stackunder{\includegraphics[width=0.13\textwidth]{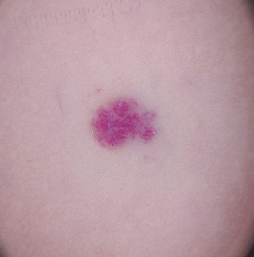}}{(V)}
    \end{minipage}
    
    \caption{\textbf{(A)} Coupled DCGAN architecture. BN refers to batch normalization, Conv and TConv refer to convolution and transposed convolution, respectively. \textbf{(B) First three rows:} Generated image of size $256 \times 256$ (we show $3$ exemplary images per class; images in the same row are generated from the same random latent vector) for \textbf{(P)}: Actinic Keratosis \textbf{(Q)} Basal Cell Carcinoma \textbf{(R)} Benign Keratosis \textbf{(S)} Melanoma \textbf{(T)} Nevus \textbf{(U)} Dermatofibroma \textbf{(V)} Vascular Lesion. \textbf{Fourth row:} Images of real lesion similar (in terms of the MSE) to the generated ones from the third row.}
	\label{fig:gan_out}
\end{figure}
As the number of classes in the ISIC 2018 task is larger than in case of ISIC 2017, using multiple separate DCGANs for the former becomes inefficient. Instead, for the ISIC 2018 we coupled seven DCGAN architectures. They share parametrization of their initial $4$ layers with each other and the final $3$ layers are class-specific. Figure~\ref{fig:gan_out} and Table~\ref{tab:DCGAN2} shows the idea behind the coupled DCGAN models. The same figure shows exemplary images generated using this approach. The coupled DCGAN models were trained using Adam optimizer with learning rate of $2e^{-4}$ and beta values of $0.5$ and $0.999$. The latent vector of length $100$ that inputs the generator is obtained from standard Gaussian distribution with mean $0$ and standard deviation $1$. Binary cross entropy loss was used to train both discriminator and generator. This time we balanced the data online i.e the data was augmented at each mini-batch. Thus the model processes a balanced mini-batch before updating the model parameters. We additionally used standard online data augmentation techniques.

\subsection{Classification}
\label{supp:class}
In Table~\ref{tab:values} we demonstrate the advantage of using data purification using ISIC 2018. We report the results when at testing we either perform or not perform data purification. Note that ISIC 2018 does not publish labels for the test data and utilizes black box system to output relevant metrics.
\begin{table}[t]
    \setlength{\tabcolsep}{1pt}
    \begin{tabular}{|c|c|c|c|}
        \hline
        Method &  Accuracy & Sensitivity &Specificity \\
        \hline
        Our Classification Model  & 0.675 & 0.561	& \textbf{0.954}\\
        but without performing &&&\\
        data purification at testing &&&\\
        Our Classification Model & \textbf{0.717} & \textbf{0.754}& 0.837\\
        \hline
    \end{tabular} 
    \caption{The effect of inducing data purification at testing on the performance of Classification Network. }
    \label{tab:values}
\end{table}

\end{document}